\documentclass{article}

\usepackage{spconf,amsmath,graphicx}
\usepackage{graphicx}
\usepackage{epstopdf}
\usepackage{amsmath,amsthm,amssymb}
\usepackage{psfrag}
\usepackage{tabularx}
\usepackage{eso-pic}
\usepackage{xspace}
\usepackage{color}
\usepackage{longtable}
\usepackage{rotating}
\usepackage{subfigure}
\usepackage{epsfig}
\usepackage{fancyhdr}
\usepackage{url}
\usepackage{setspace}
\usepackage{colortbl}

\usepackage{algorithm}
\usepackage[noend]{algpseudocode}
\makeatletter
\def\BState{\State\hskip-\ALG@thistlm}
\makeatother

\newcommand{\eq}[1]{Eq.~\ref{#1}}
\newcommand{\fig}[1]{Fig.~\ref{#1}}
\newcommand{\tbl}[1]{Table~\ref{#1}}
\newcommand{\ssec}[1]{Section~\ref{#1}}
\newcommand{\alg}[1]{Algorithm~\ref{#1}}

\newcommand{\pushcode}[1][1]{\hskip\dimexpr#1\algorithmicindent\relax}

\title{UNSUPERVISED CONVOLUTIONAL NEURAL NETWORKS FOR MOTION ESTIMATION}
\name{Aria Ahmadi, Ioannis Patras}
\address{School of Electronic Engineering and Computer Science\\
	Queen Mary University of London\\
	Mile End road, E1 4NS, London, UK}
\begin{document}
%
\maketitle
\begin{abstract} 

Traditional methods for motion estimation estimate the motion field $F$ between a pair of images as the one that minimizes a predesigned cost function. In this paper, we propose a direct method and train a Convolutional Neural Network (CNN) that when, at test time, is given a pair of images as input it produces a dense motion field $F$ at its output layer. In the absence of large datasets with ground truth motion that would allow classical supervised training, we propose to train the network in an unsupervised manner. The proposed cost function that is optimized during training, is based on the classical optical flow constraint. The latter is differentiable with respect to the motion field and, therefore, allows backpropagation of the error to previous layers of the network. Our method is tested on both synthetic and real image sequences and performs similarly to the state-of-the-art methods.

\end{abstract}
\section{Introduction}
\label{sec:intro}


Motion fields, that is fields that describe how pixels move from a reference to a target frame, are rich source of information for the analysis of image sequences and beneficial for several applications such as video coding \cite{dufaux2000efficient, li2001mpeg}, medical image processing \cite{keeling2006medical}, segmentation \cite{long_shelhamer_fcn} and human action recognition \cite{jain2013better,wang2013action}. Traditionally, motion fields are estimated using the variational model proposed by Horn and Schunck \cite{horn1981determining} and its variants such as \cite{brox2004high, brox2011large}. Very recently, inspired by the great success of Deep Neural Networks in several Computer Vision problems \cite{krizhevsky2012imagenet}, a CNN has been proposed Fischer et al. in \cite{DFIB15} for motion estimation. The method showed performance that was close to the state-of-the-art in a number of synthetically generated image sequences.




A major problem with the method proposed in \cite{DFIB15} is that the proposed CNN needed to be trained in a supervised manner, that is, it required for training synthetic image sequences where ground truth motion fields were available. Furthermore, in order to generalize well to an unseen dataset, it needed fine tuning, also requiring ground truth data on samples from that dataset. Ground truth motion estimation are not easily available though. For this reason, the method proposed in \cite{DFIB15} was applied only on synthetic image sequences.

In this paper, we propose training a CNN for motion estimation in an unsupervised manner. We do so by designing a cost function that is differentiable with respect to the unknown motion field and, therefore, allows the backpropagation of the error and the end to end training of the CNN. The cost function builds on the widely used optical flow constraint - our major difference to Horn-Schunk based methods is that the cost function is used only during training and without regularization. Once trained, given a pair of frames as input the CNN gives at its output layer an estimate of the motion field. In order to deal with motions large in magnitude, we embed the proposed network in a classical iterative scheme, in which at the end of each iteration the reference image is warped towards the target image and in a classical coarse-to-fine multiscale framework. We train our CNN using randomly chosen pairs of consecutive frames from UCF101 dataset and test it on both the UCF101 where it performs similarly to the state-of-the-art methods and on the synthetic MPI-Sintel dataset where it outperforms them. 

The remainder of the paper is organized as follows. In \ssec{sec:method} we describe our method. In \ssec{sec:experiment} we present our experimental results. Finally, in \ssec{sec:conclusion} we give some conclusion.

\section{Method}
\label{sec:method}

At the heart of all motion estimation methods is the minimization of the difference between features extracted at a certain location $x$,$y$ at the reference frame $t$ and its correspondence in the frame $t+dt$. The now classical Horn and Schunck \cite{horn1981determining} method penalizes the deviation from the assumption of constant intensity, that states that the intensity at a pixel in the reference frame at time $t$ and the intensity at its correspondence at time $t+dt$ are the same. Formally, the goal is the minimization of the motion compensated intensity differences, that is, \cite{goldstein2012global}
\begin{equation}
\begin{split}
\label{Eq:hornshunck}
 \sum_{x,y = 1}^M{|I(x+u(x,y),y+v(x,y),t+\Delta{t})-I(x,y,t)|^2},
\end{split}
 \end{equation}
where $I(x,y,t)$ is the intensity at pixel $(x,y)$ at frame $t$, and $F(x,y) \triangleq \left[ \begin{array}{c} u(x,y) \\ v(x,y) \end{array} \right]$ is the unknown motion vector at pixel $(x,y)$. Clearly, $u(x,y)$ and $v(x,y)$ are respectively the horizontal and vertical displacement of the pixel with coordinates $(x,y)$. To arrive at a computationally tractable method, Horn and Schunck \cite{horn1981determining} introduced a regularization term that penalized discontinuities in the motion field and linearized the cost by taking the first order Taylor expansion with respect to the horizontal and vertical displacements. By doing the latter, they arrived at the optical flow equation $uI_x + vI_y + I_t = 0$, where $I_x$, $I_y$ and $I_t$ are the horizontal, vertical and temporal intensity derivatives respectively, and penalized deviations from it. In the equation above, we omit the pixel coordinates for notation simplicity. 

In this work, we build on the constant intensity assumption. However, instead of using it at test time to estimate the motion field as the one that minimizes the deviations from it, we use it at training time only in order to train a CNN that takes as input a pair of images and outputs at its last layer a dense motion field $F$ between them. Clearly, the motion field $F$ is a function of the weights $w$ of the CNN and the images at its input. In order to reduce the influence of outliers we use a robust error norm, that is the Charbonnier penalty $\rho(x) = \sqrt{x^2+\epsilon}$ \cite{bruhn2005lucas}, a differentiable variant of the $L1$ norm, the most robust convex function. Formally, during training we learn the CNN by optimizing the sum of costs that for a pair of images $I(t)$ and $I(t+dt)$ are defined as follows: 

\begin{equation}
\label{Eq:TaylorExpansion}
  E(F) = \sum_{x,y = 1}^M \sqrt{(uI_x + vI_y + I_t)^2 + \epsilon}, 
\end{equation}
where the image coordinates $x$,$y$ are omitted for notation simplicity.

Clearly, the motion field $F$ is a function of the weights $w$ of the CNN and the images at its input. Therefore, the cost function in \eq{Eq:TaylorExpansion} is a function of the CNN weights. More importantly, our cost function allows us to calculate the derivatives of it with respect to the network weights. Specifically, using the chain rule,
\begin{equation}
\label{Eq:DrvCostWrtWeights}
  \frac{\partial E}{\partial w} = \frac{\partial E}{\partial F} \frac{\partial F}{\partial w}.
\end{equation}

The second part, that is $\frac{\partial F}{\partial w}$, are the partial derivatives of the output $F$ of the CNN with respect to its weights $w$. This can be calculated in a classical manner using the standard form of the backpropagation algorithm. The first term, that is $\frac{\partial E}{\partial F}$, can be calculated in closed form as 
\begin{equation}
\label{Eq:ClosedForm}
  \frac{\partial E}{\partial F} = \begin{bmatrix}
    \frac{\partial E}{\partial u} \\[0.3em]
    \frac{\partial E}{\partial v} \\[0.3em]       
  \end{bmatrix}= \begin{bmatrix}
    \sum_{x,y = 1}^M \frac{I_x(uI_x + vI_y + I_t)}{\sqrt{(uI_x + vI_y + I_t)^2 + \epsilon}} \\[0.3em]
    \sum_{x,y = 1}^M \frac{I_y(uI_x + vI_y + I_t)}{\sqrt{(uI_x + vI_y + I_t)^2 + \epsilon}} \\[0.3em]       
  \end{bmatrix}.
\end{equation}

The cost function that is used for training relies on the optical flow constraint. This is known that it does not hold when the motions are large in magnitude. Following the dominant paradigm in the field, we embed our method in a coarse-to-fine multiscale iterative scheme.  

At test time, that is once trained, given a pair of frames as input the CNN gives as output an update on the motion field. The updated motion field is then used to warp the second frame towards the first one, and the new pair of images are given as input to the CNN to calculate another update on the motion field. Several iterations are performed at each scale of the muti-scale framework. After each update, and similarly to other methods in the literature \cite{sun2014quantitative}, the calculated motion field in each iteration is filtered (by a median filter in our case). The proposed algorithm at test time is summarized in \alg{alg:mainapproach}.

\begin{algorithm}
\caption{The algorithm of our proposed framework during the test time.}
\label{alg:mainapproach}
\begin{algorithmic}[1]
\Procedure{}{}
\State $I_1$,$I_2$: Two input frames
\State $ F_{tot} $: The desired motion field
\State $I_1^n$,$I_2^n$: The downsampled versions of $I_1$ and $I_2$ by \\ \pushcode[1] a factor of $ n = 2 ^ k$
\State $ F_{tot} = 0$
\State $I_{2_w}^n \leftarrow I_2^n$ 
\While{$n \geq 1$}
  \While{Number of iterations is less than $4$  }
    \State $\Delta{F} \leftarrow CNN(I_1^n,I_{2_w}^n)$ : Calculate the \\ \pushcode[1] update on the motion field
    \State $\Delta{F} \leftarrow MedFilt(\Delta{F})$ : Median filter the \\ \pushcode[1] motion field
    \State $F_{tot} \leftarrow F_{tot} + \Delta{F}$ : Update $F_{tot}$ using the \\ \pushcode[1] motion field
    \State $I_{2_w}^n \leftarrow warp(I_2^n,F_{tot})$ : Warp $I_2^n$ towards $I_1^n$ \\ \pushcode[1] using the motion field
  \EndWhile
\State \textbf{end while}
\State Up-sample $F_{tot}$ by a factor of 2
\State $n = \frac{n}{2}$
\EndWhile
\State \textbf{end while}
\State \textbf{Return} $F_{tot}$
\EndProcedure
\end{algorithmic}
\end{algorithm}

\begin{figure*}[!t]
  \centering
  \begin{tabular}{  c  c  c  c  c  c }
    Images & DeepFlow & EpicFlow & HAOF & LDOF & USCNN \\ 
    \begin{minipage}{.1\textwidth}
      \vspace*{0.05\textwidth}
      \includegraphics[width=\linewidth, height=10mm]{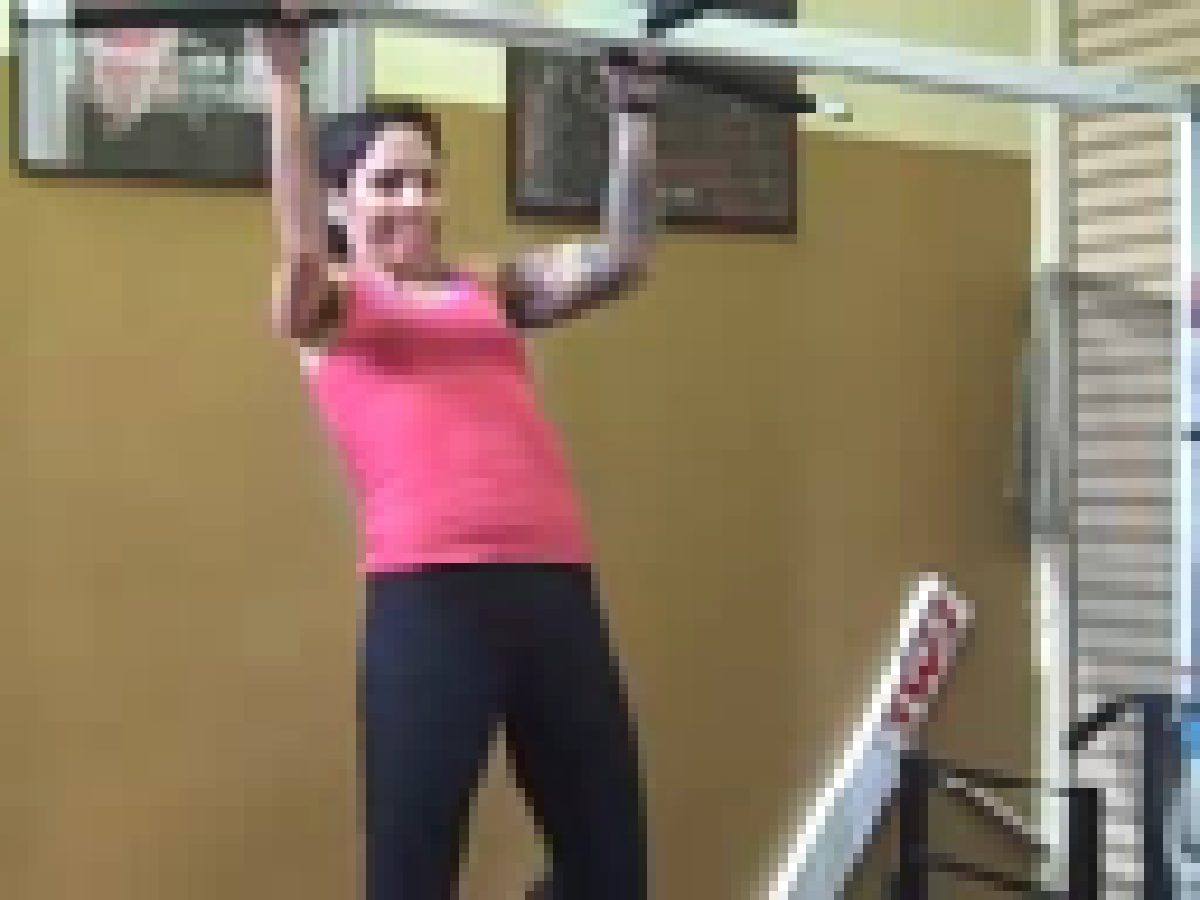}
    \end{minipage}
    &
    \begin{minipage}{.1\textwidth}
      \vspace*{0.05\textwidth}
      \includegraphics[width=\linewidth, height=10mm]{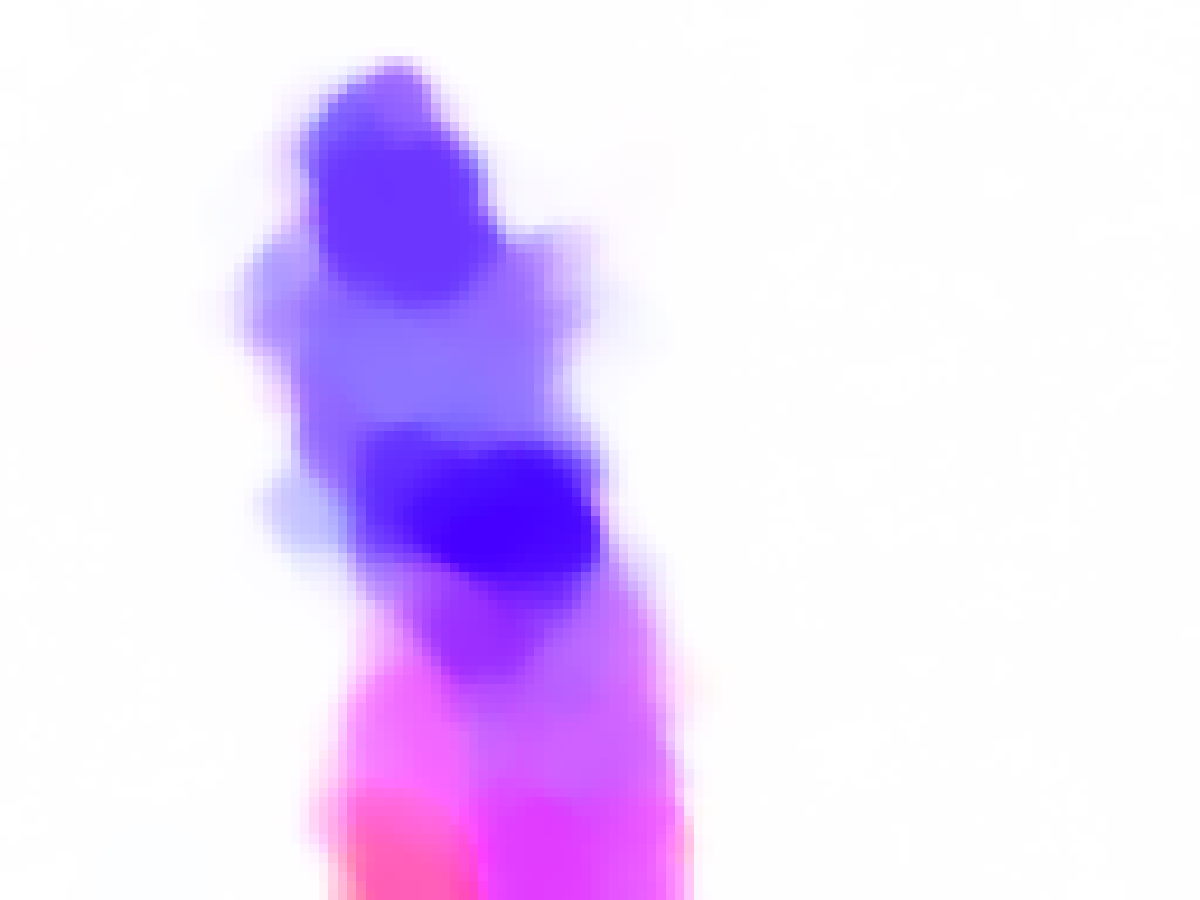}
    \end{minipage}
    & 
    \begin{minipage}{.1\textwidth}
      \vspace*{0.05\textwidth}
      \includegraphics[width=\linewidth, height=10mm]{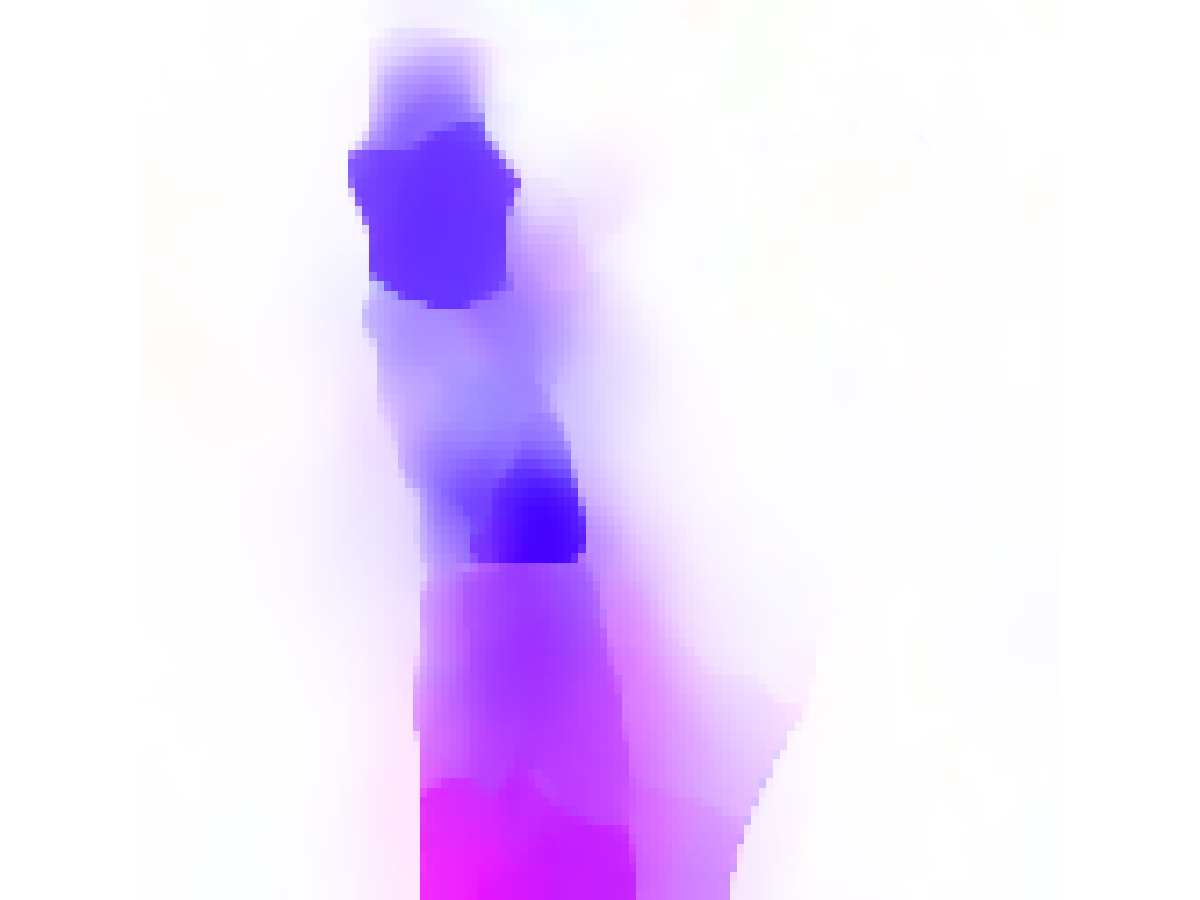}
    \end{minipage}
    &
    \begin{minipage}{.1\textwidth}
      \vspace*{0.05\textwidth}
      \includegraphics[width=\linewidth, height=10mm]{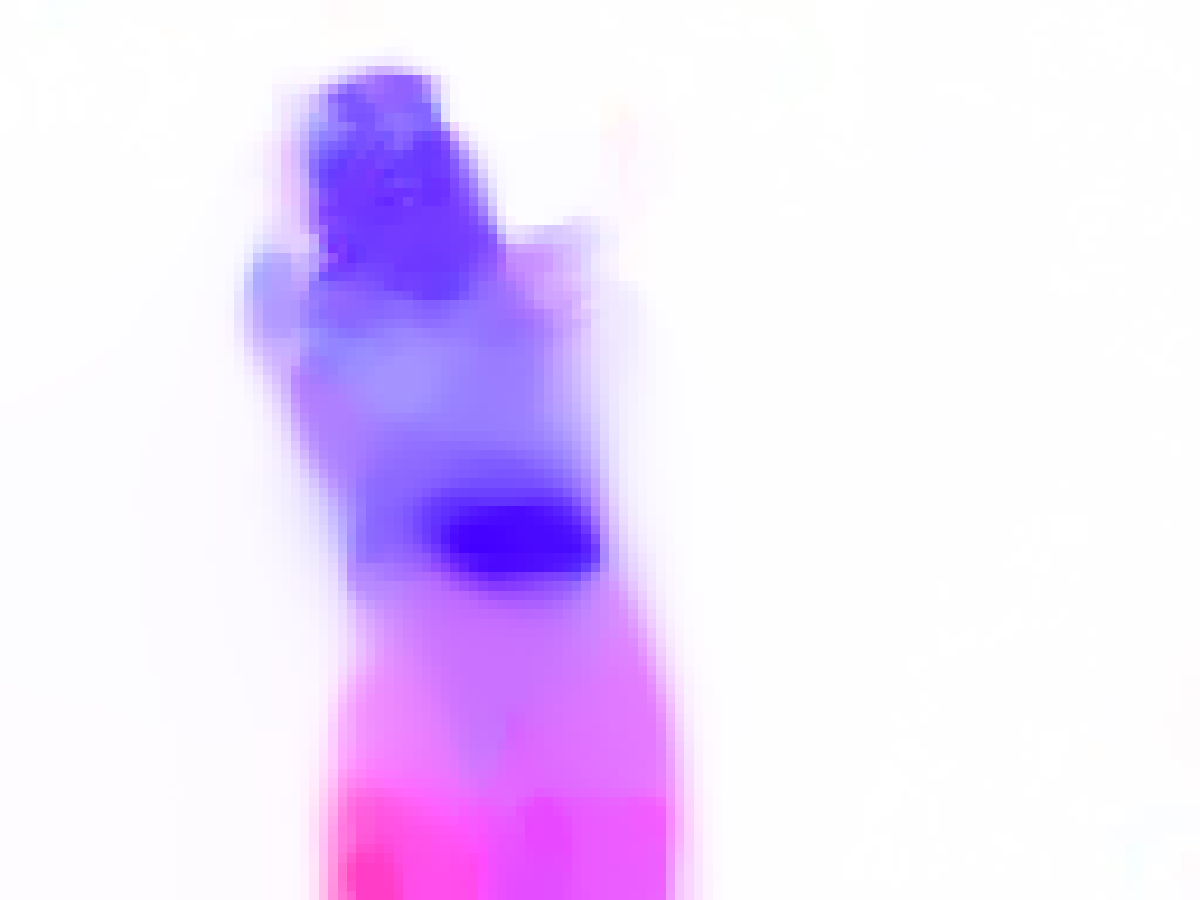}
    \end{minipage}
    &
    \begin{minipage}{.1\textwidth}
      \vspace*{0.05\textwidth}
      \includegraphics[width=\linewidth, height=10mm]{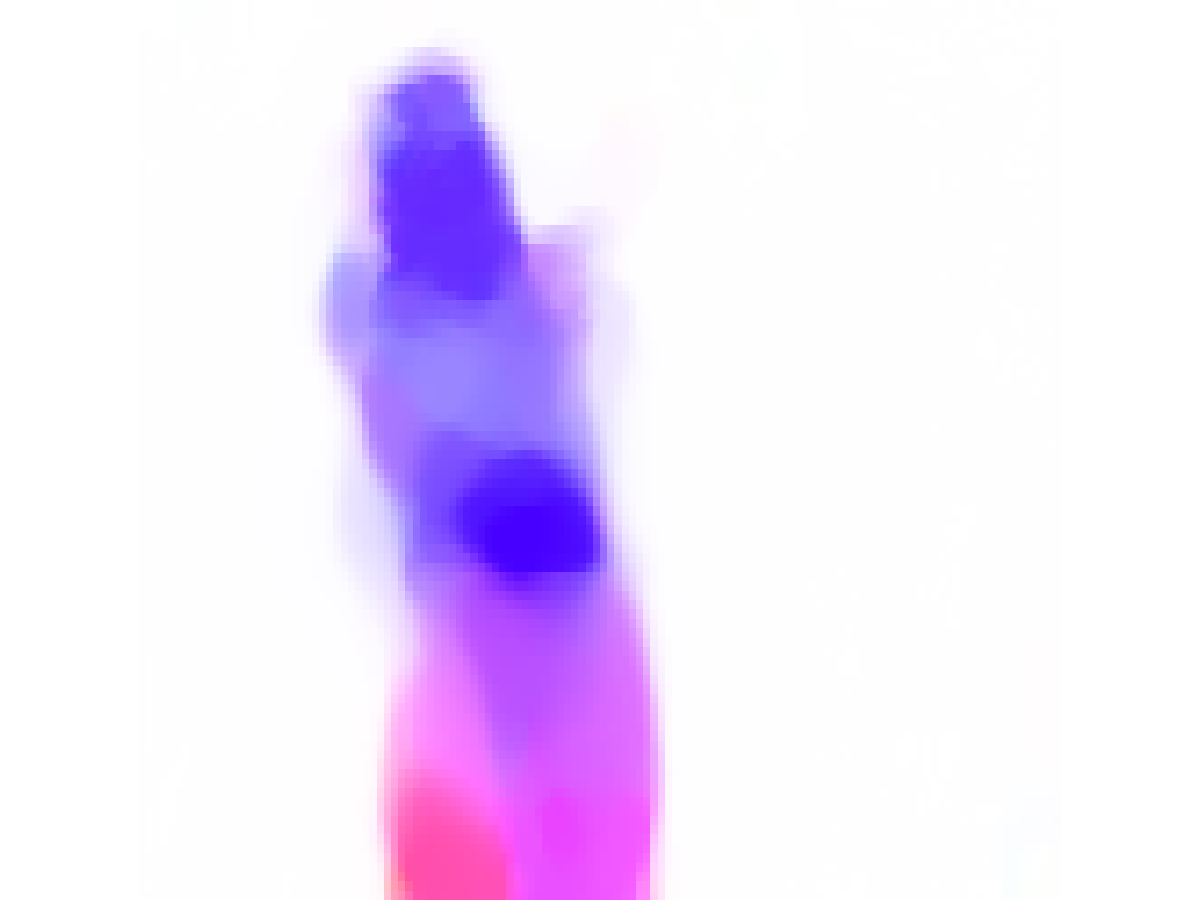}
    \end{minipage}
    &
    \begin{minipage}{.1\textwidth}
      \vspace*{0.05\textwidth}
      \includegraphics[width=\linewidth, height=10mm]{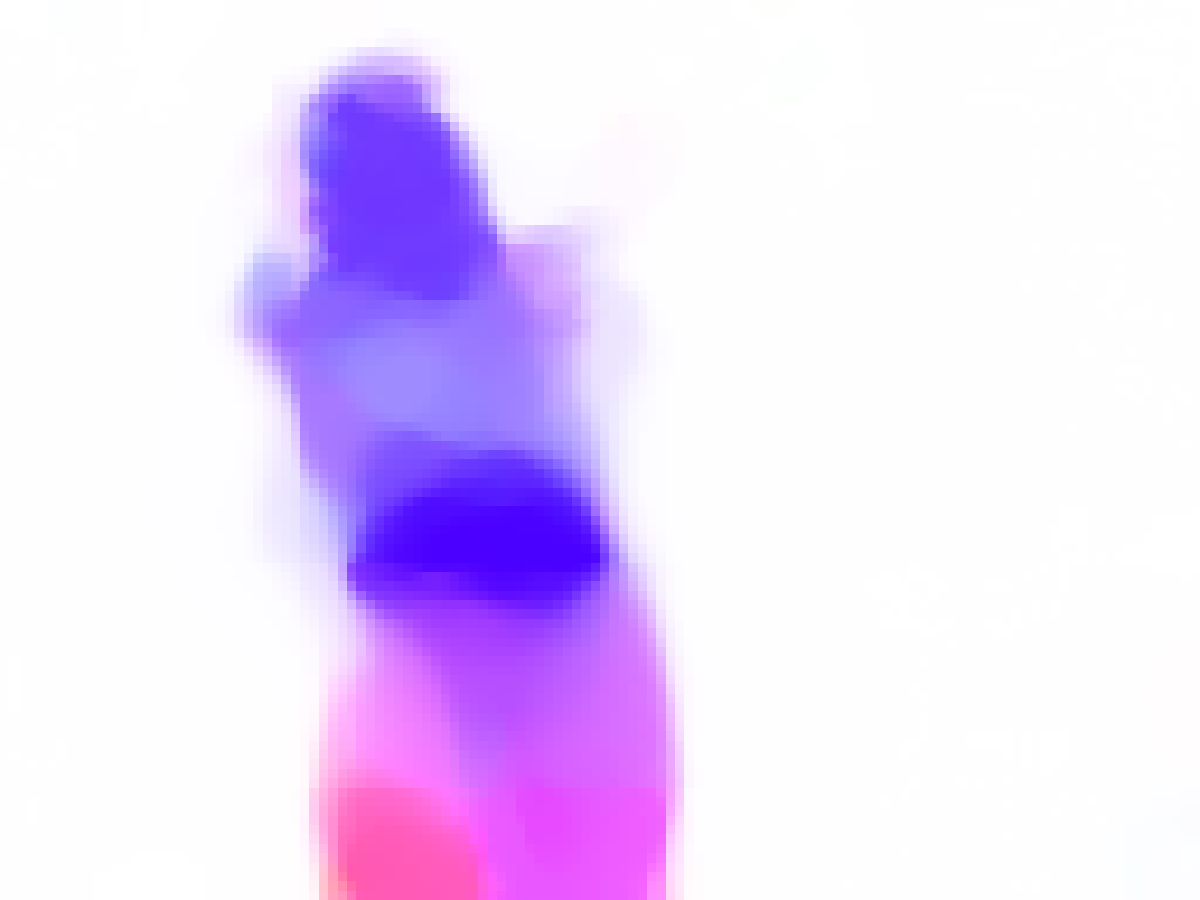}
    \end{minipage}
    
    \\
    
    \begin{minipage}{.1\textwidth}
      \vspace*{0.05\textwidth}
      \includegraphics[width=\linewidth, height=10mm]{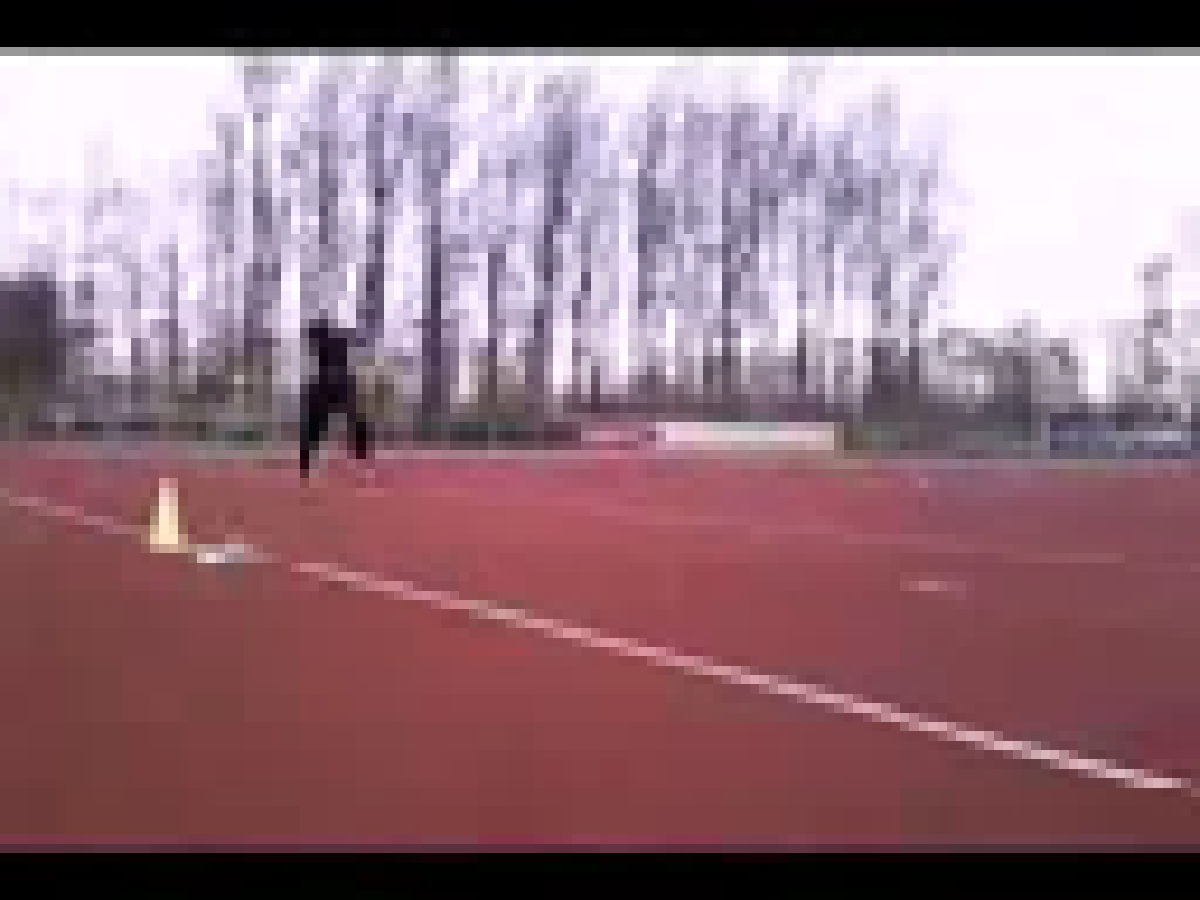}
    \end{minipage}
    &
    \begin{minipage}{.1\textwidth}
      \vspace*{0.05\textwidth}
      \includegraphics[width=\linewidth, height=10mm]{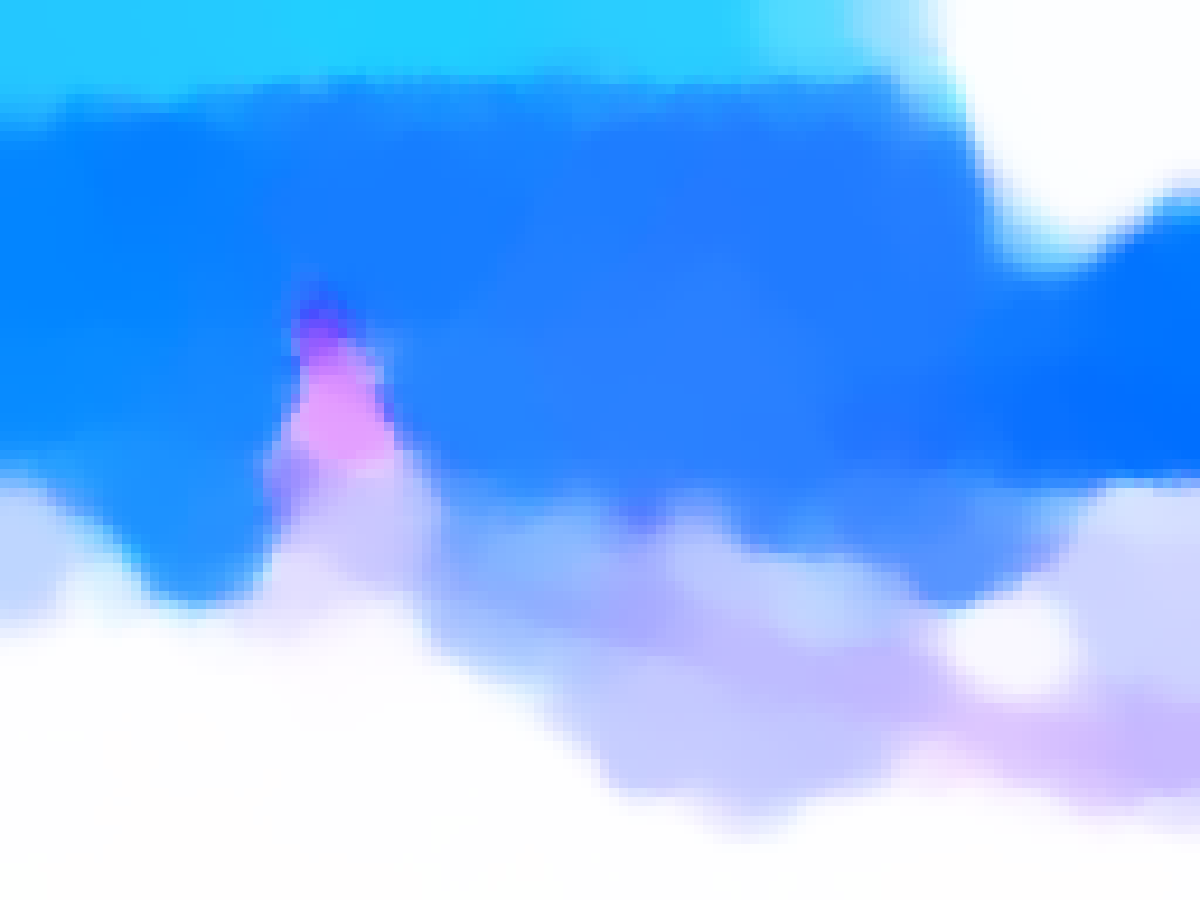}
    \end{minipage}
    & 
    \begin{minipage}{.1\textwidth}
      \vspace*{0.05\textwidth}
      \includegraphics[width=\linewidth, height=10mm]{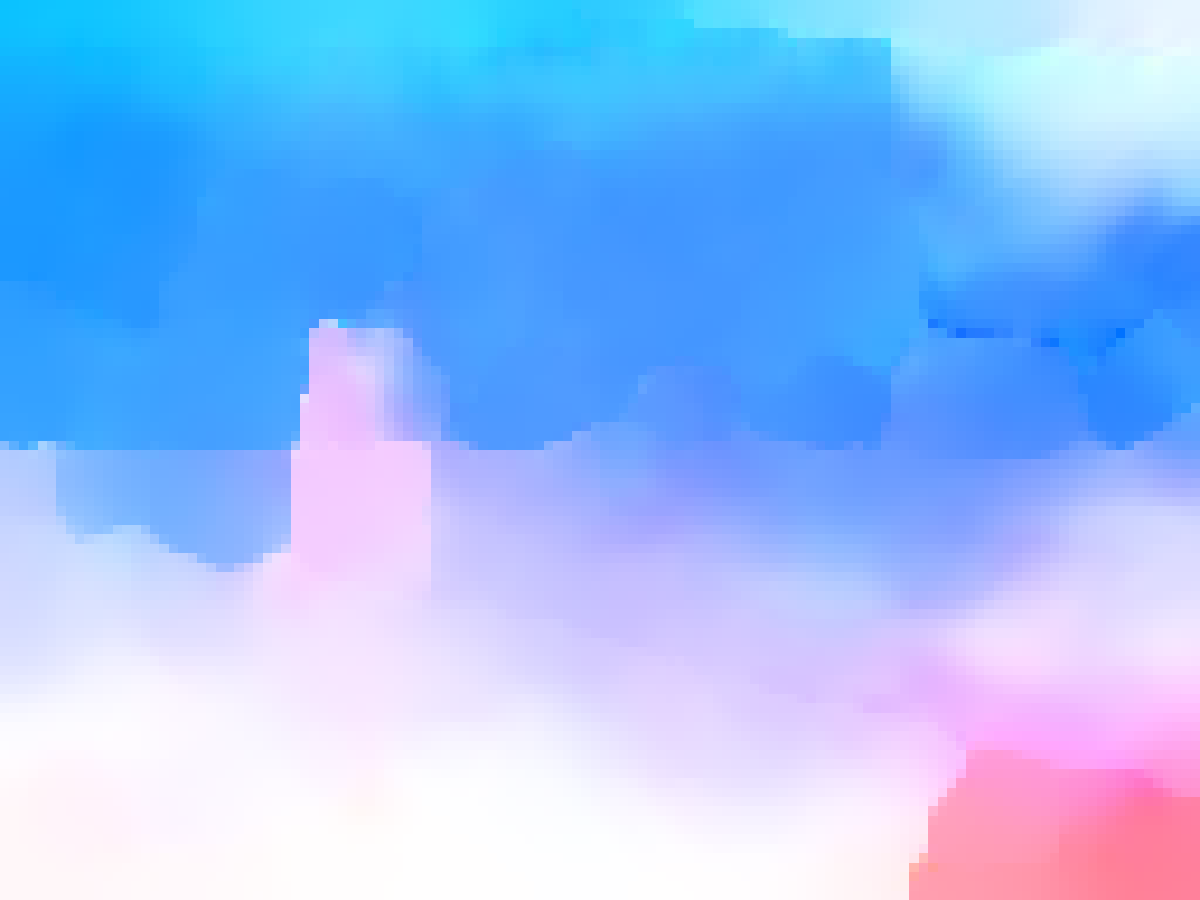}
    \end{minipage}
    &
    \begin{minipage}{.1\textwidth}
      \vspace*{0.05\textwidth}
      \includegraphics[width=\linewidth, height=10mm]{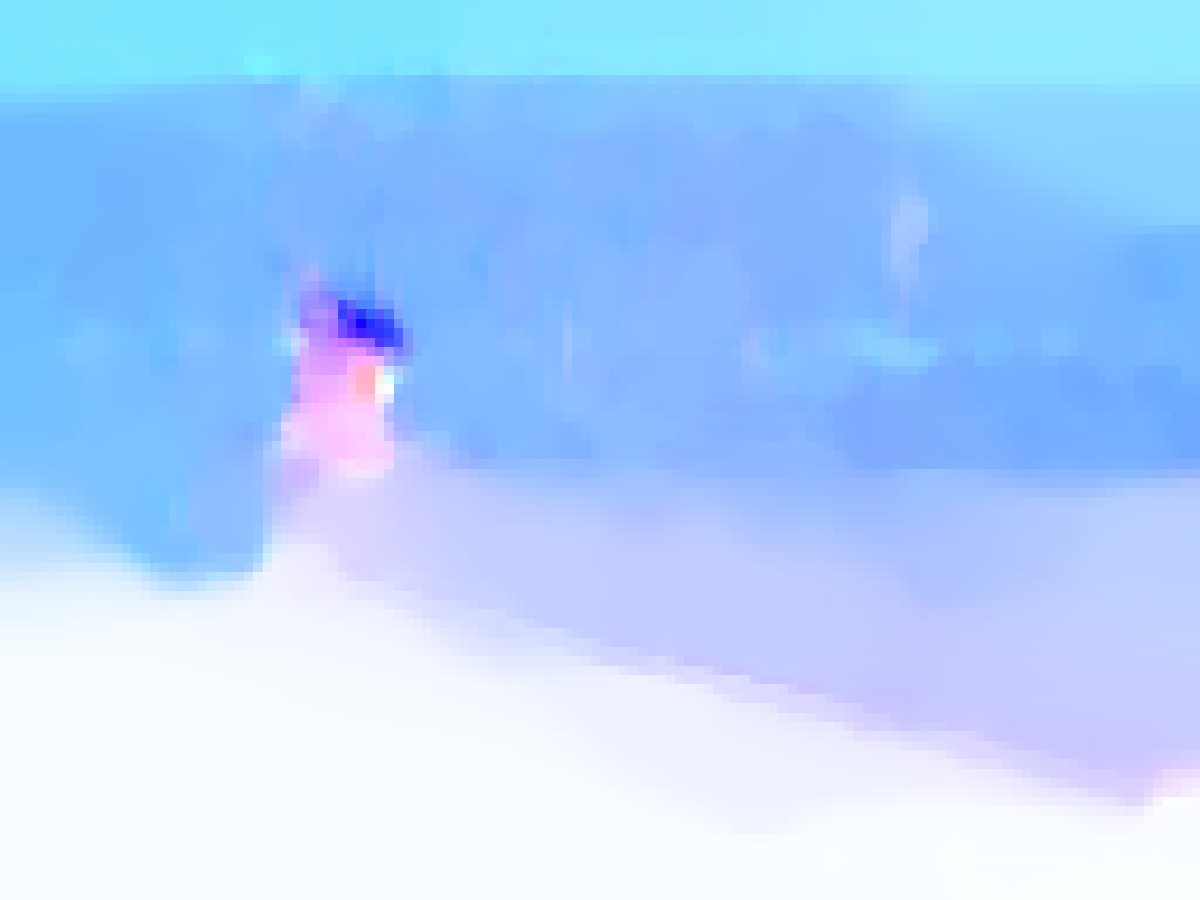}
    \end{minipage}
    &
    \begin{minipage}{.1\textwidth}
      \vspace*{0.05\textwidth}
      \includegraphics[width=\linewidth, height=10mm]{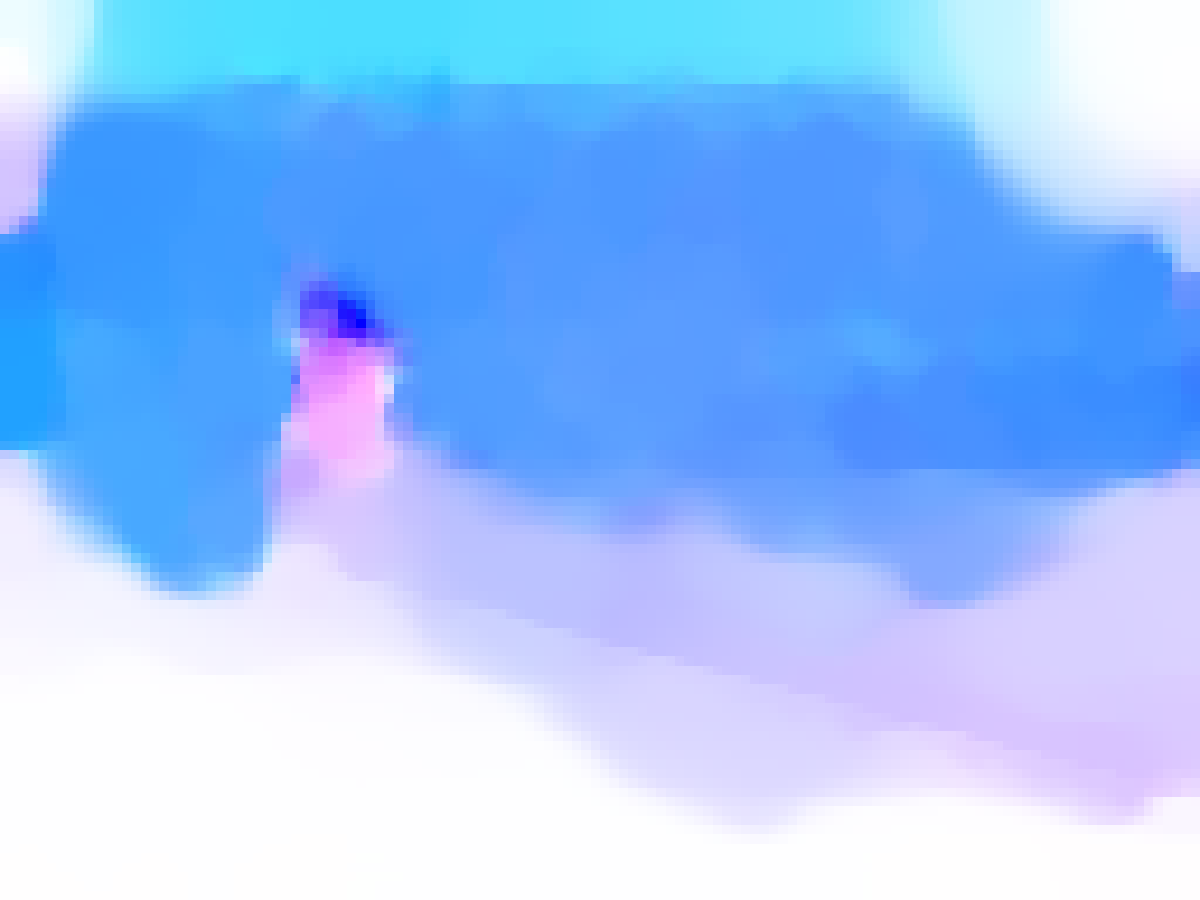}
    \end{minipage}
    &
    \begin{minipage}{.1\textwidth}
      \vspace*{0.05\textwidth}
      \includegraphics[width=\linewidth, height=10mm]{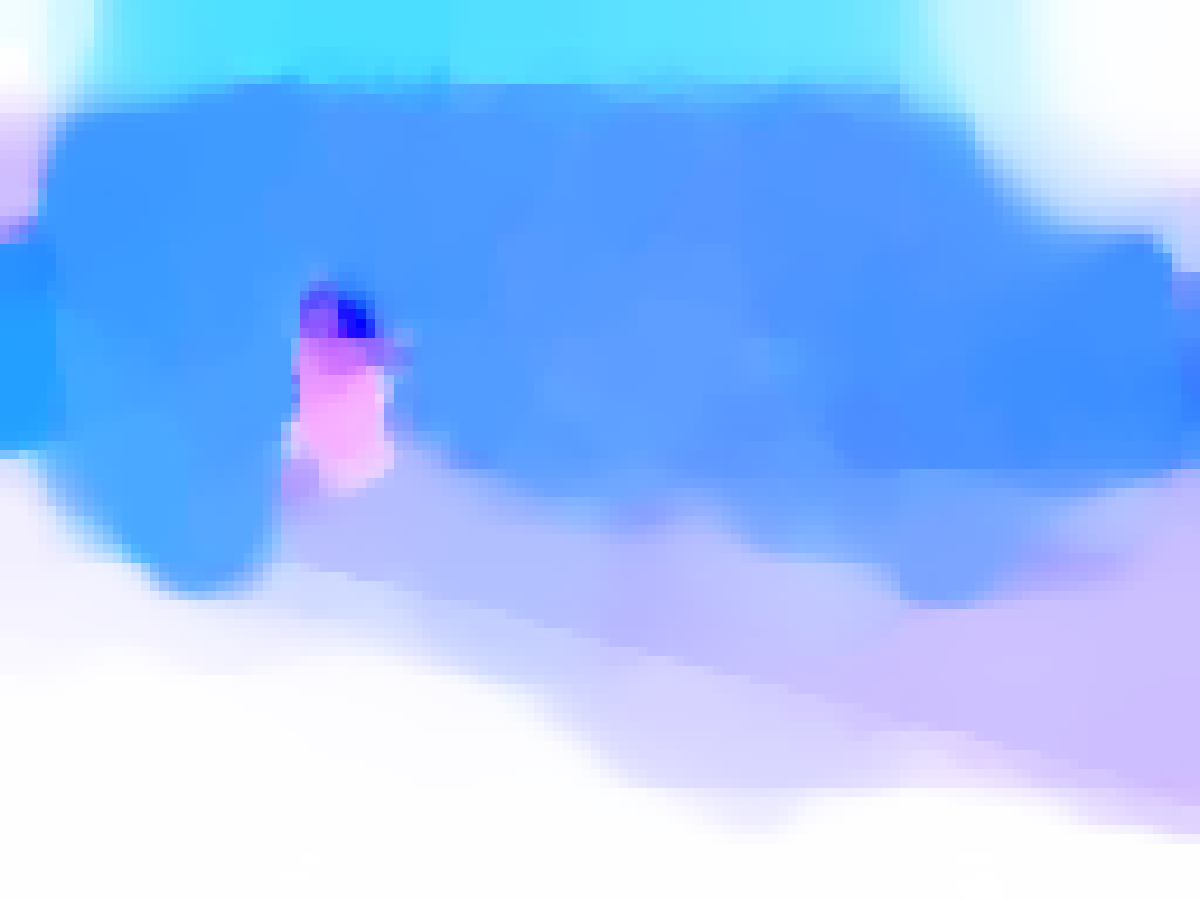}
    \end{minipage}
    \\
    
    \begin{minipage}{.1\textwidth}
      \vspace*{0.05\textwidth}
      \includegraphics[width=\linewidth, height=10mm]{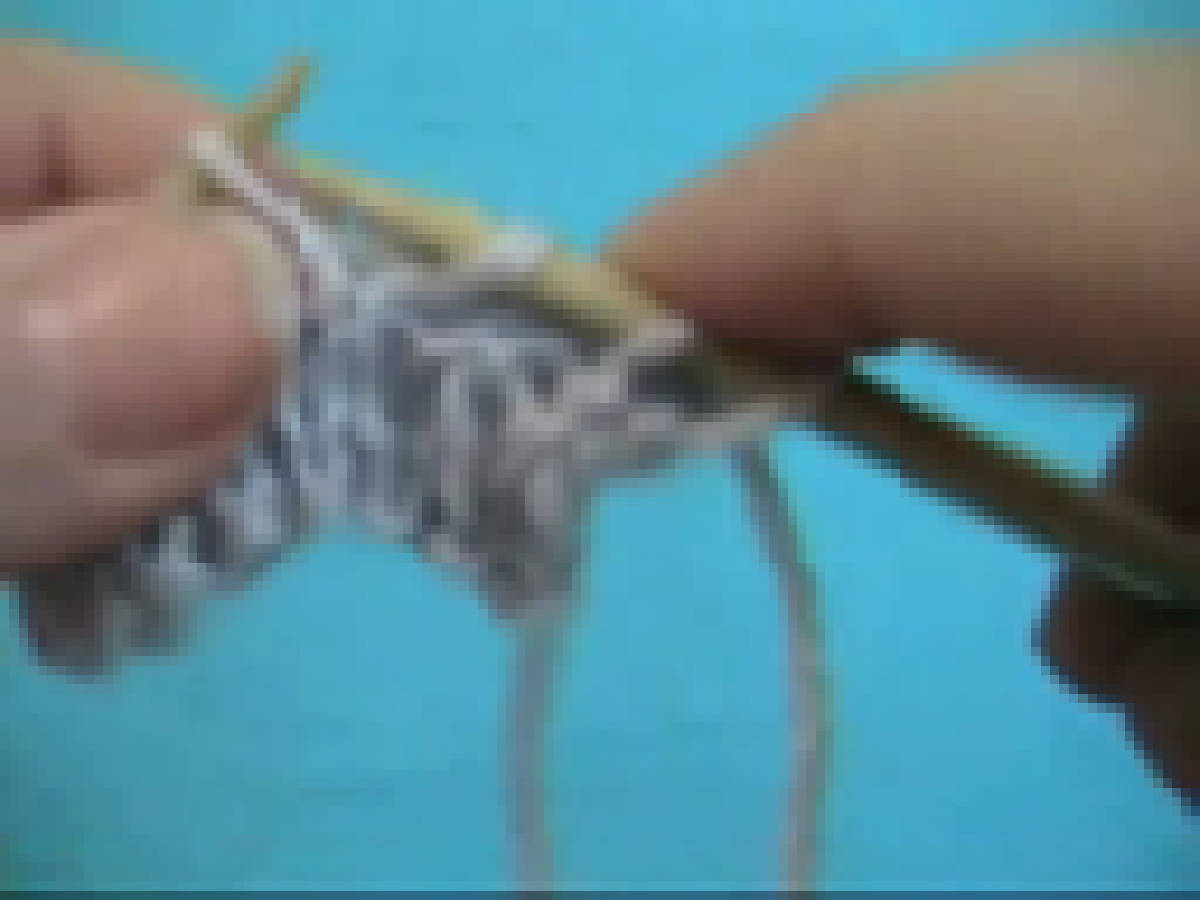}
    \end{minipage}
    &
    \begin{minipage}{.1\textwidth}
      \vspace*{0.05\textwidth}
      \includegraphics[width=\linewidth, height=10mm]{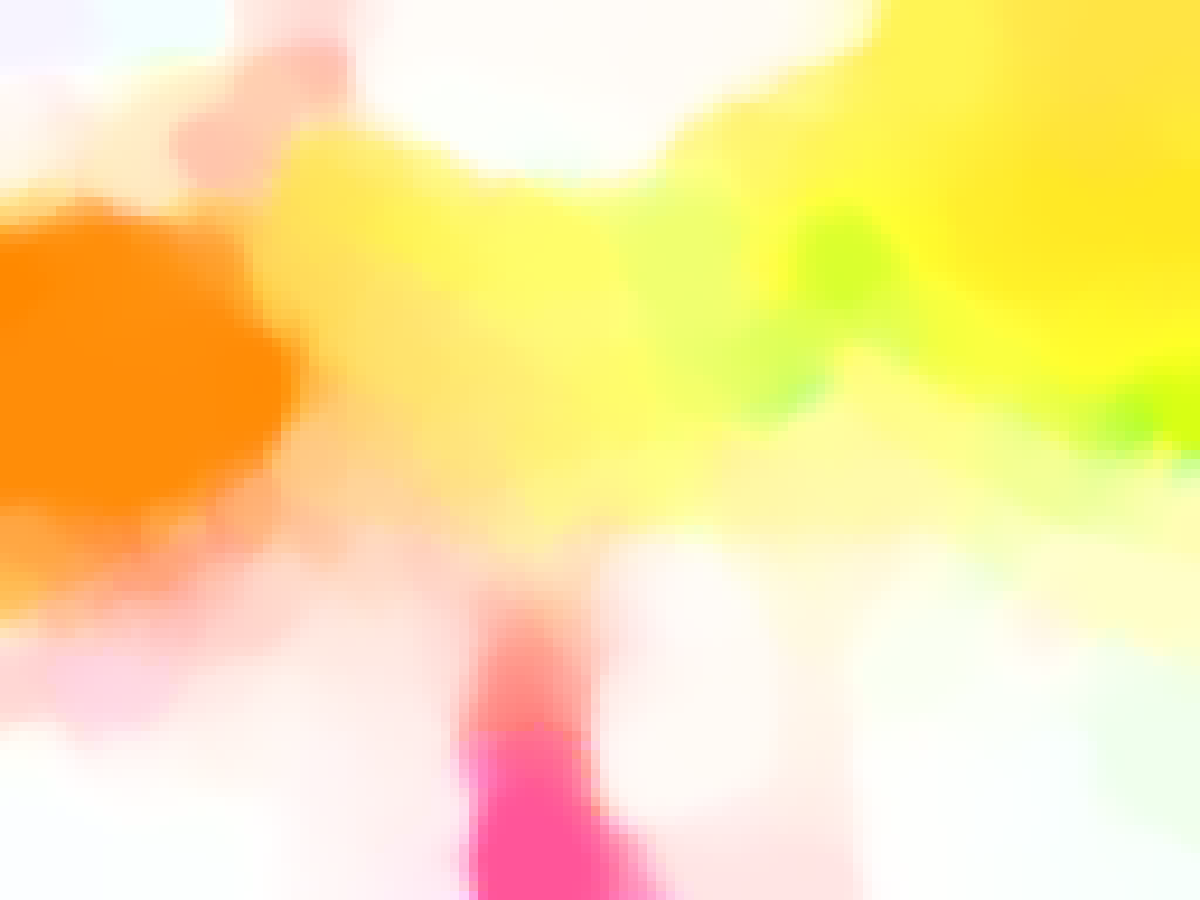}
    \end{minipage}
    & 
    \begin{minipage}{.1\textwidth}
      \vspace*{0.05\textwidth}
      \includegraphics[width=\linewidth, height=10mm]{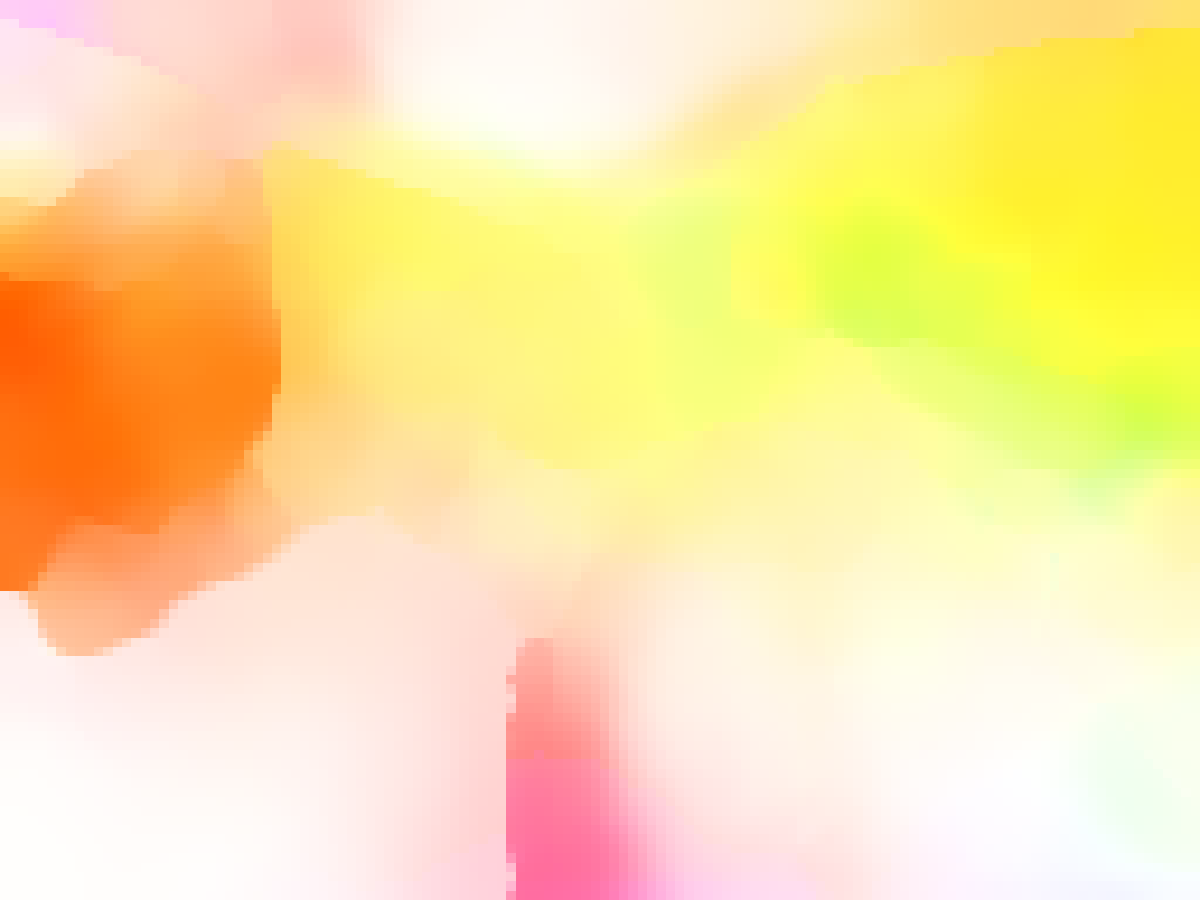}
    \end{minipage}
    &
    \begin{minipage}{.1\textwidth}
      \vspace*{0.05\textwidth}
      \includegraphics[width=\linewidth, height=10mm]{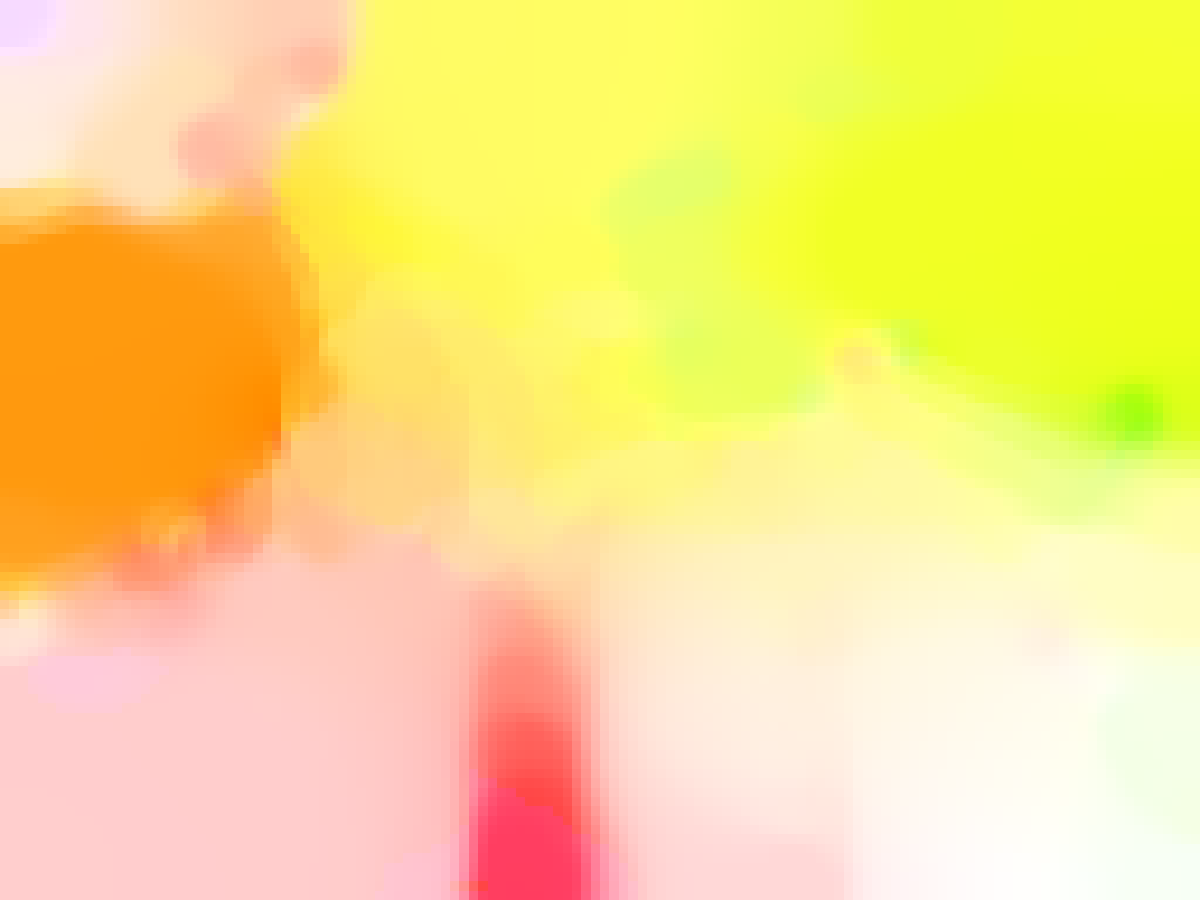}
    \end{minipage}
    &
    \begin{minipage}{.1\textwidth}
      \vspace*{0.05\textwidth}
      \includegraphics[width=\linewidth, height=10mm]{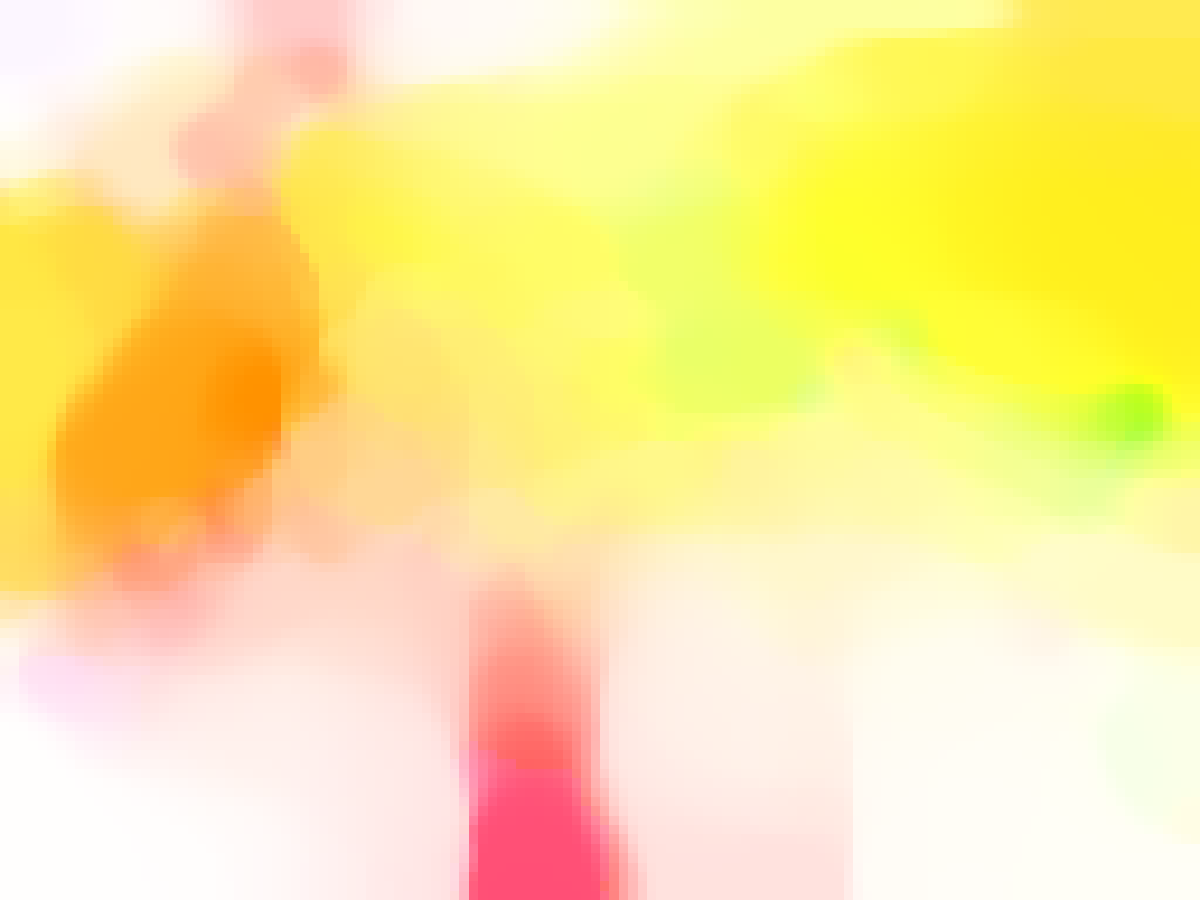}
    \end{minipage}
    &
    \begin{minipage}{.1\textwidth}
      \vspace*{0.05\textwidth}
      \includegraphics[width=\linewidth, height=10mm]{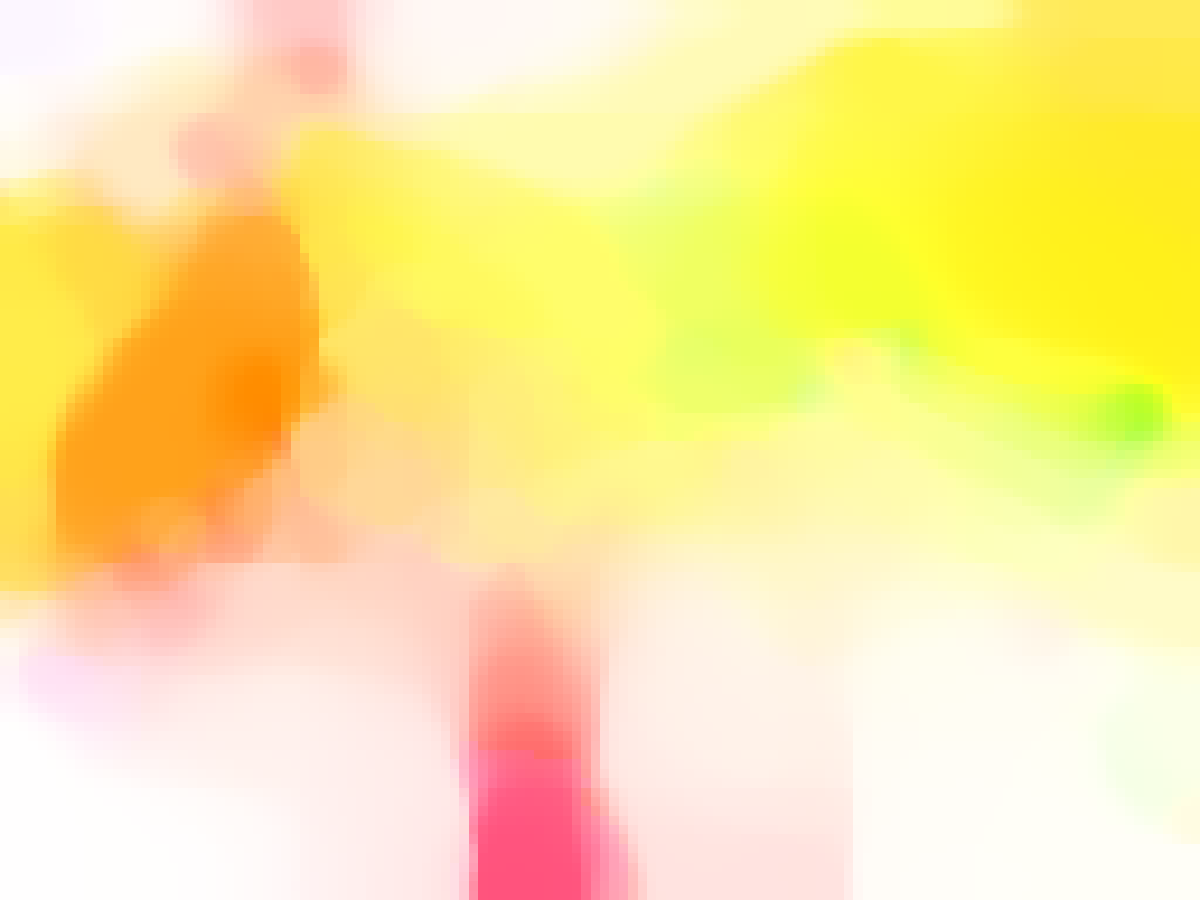}
    \end{minipage}
    \\
    
    \begin{minipage}{.1\textwidth}
      \vspace*{0.05\textwidth}
      \includegraphics[width=\linewidth, height=10mm]{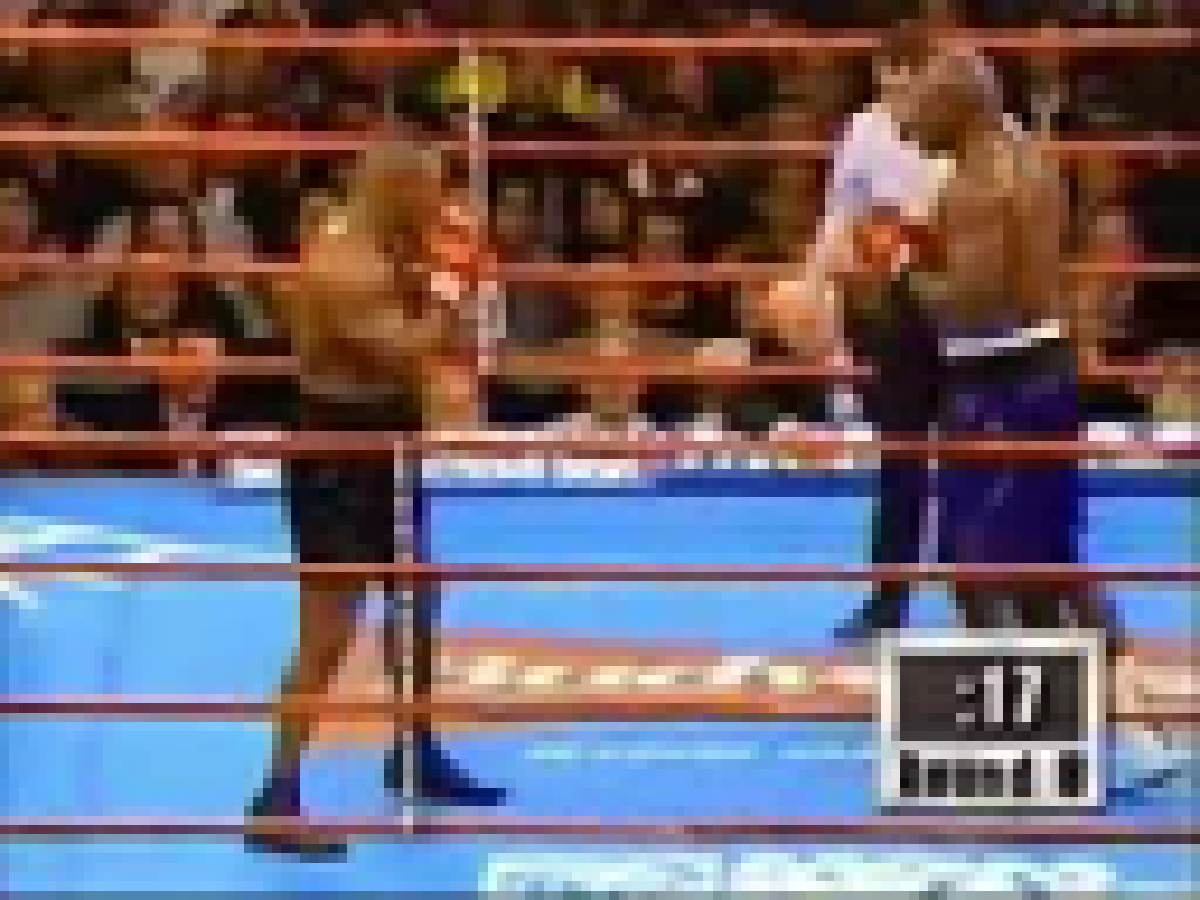}
    \end{minipage}
    &
    \begin{minipage}{.1\textwidth}
      \vspace*{0.05\textwidth}
      \includegraphics[width=\linewidth, height=10mm]{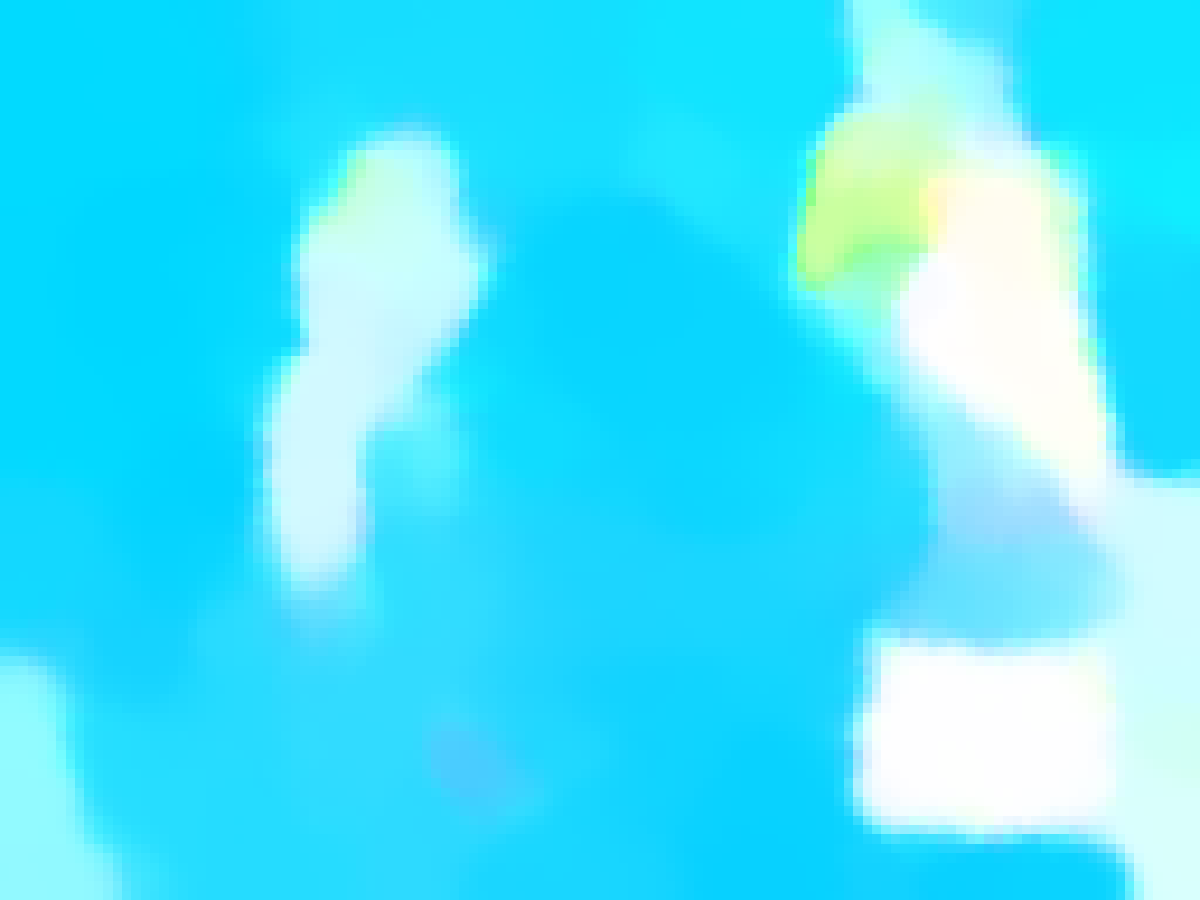}
    \end{minipage}
    & 
    \begin{minipage}{.1\textwidth}
      \vspace*{0.05\textwidth}
      \includegraphics[width=\linewidth, height=10mm]{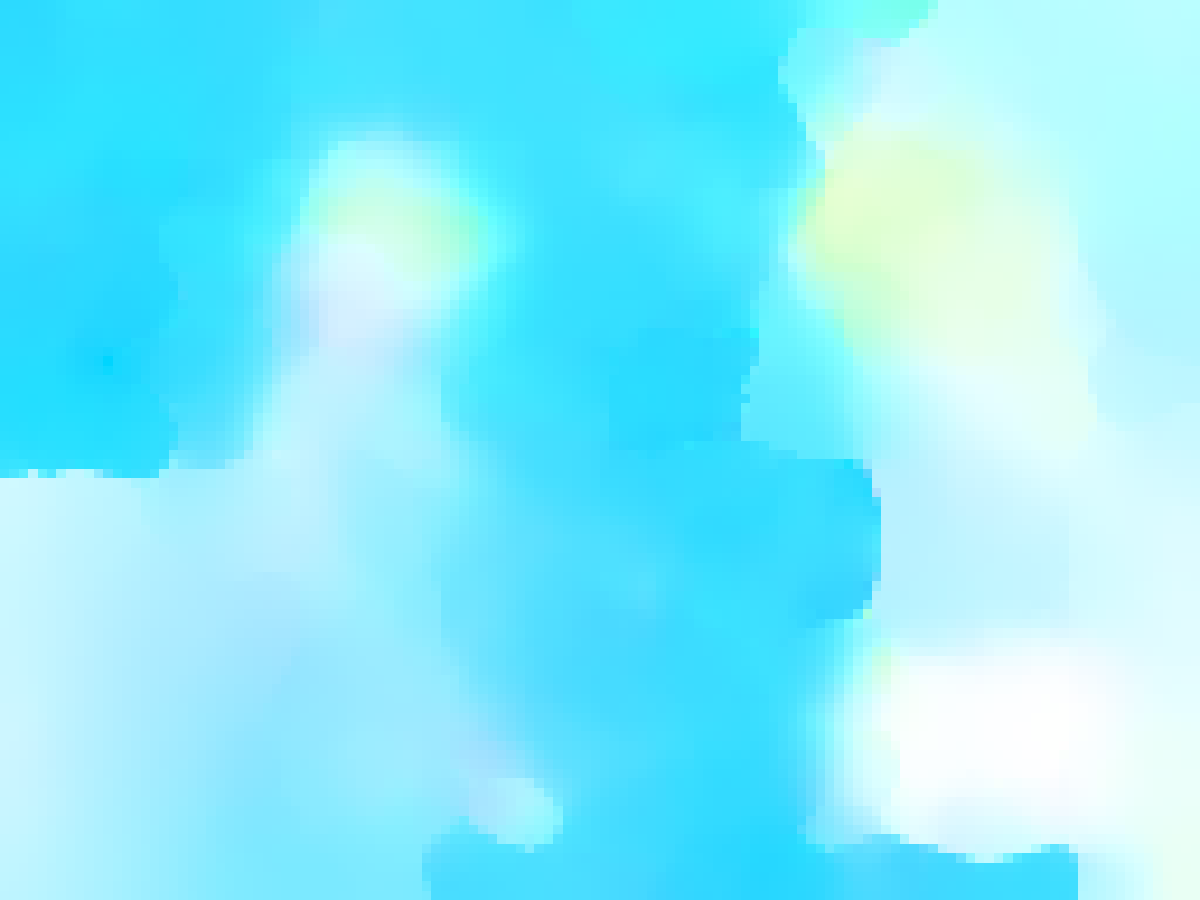}
    \end{minipage}
    &
    \begin{minipage}{.1\textwidth}
      \vspace*{0.05\textwidth}
      \includegraphics[width=\linewidth, height=10mm]{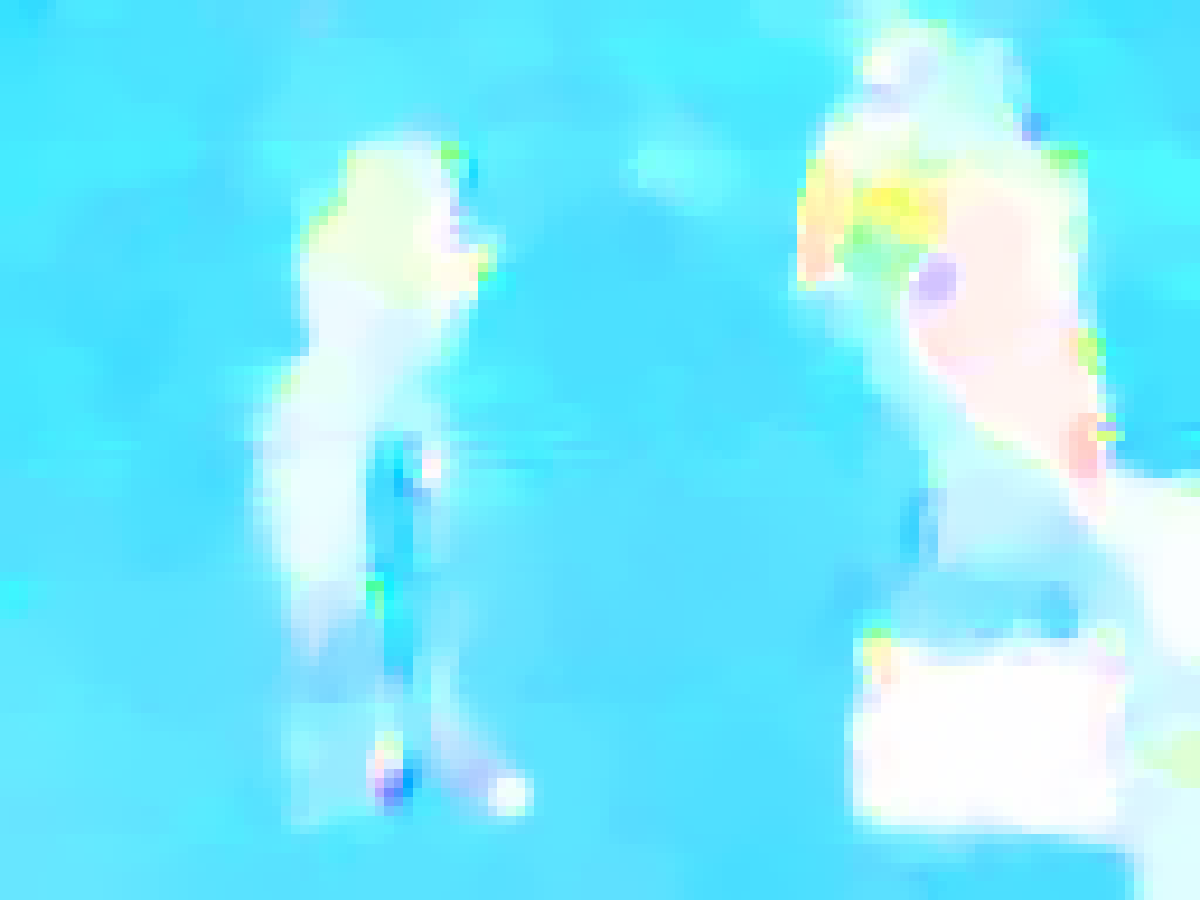}
    \end{minipage}
    &
    \begin{minipage}{.1\textwidth}
      \vspace*{0.05\textwidth}
      \includegraphics[width=\linewidth, height=10mm]{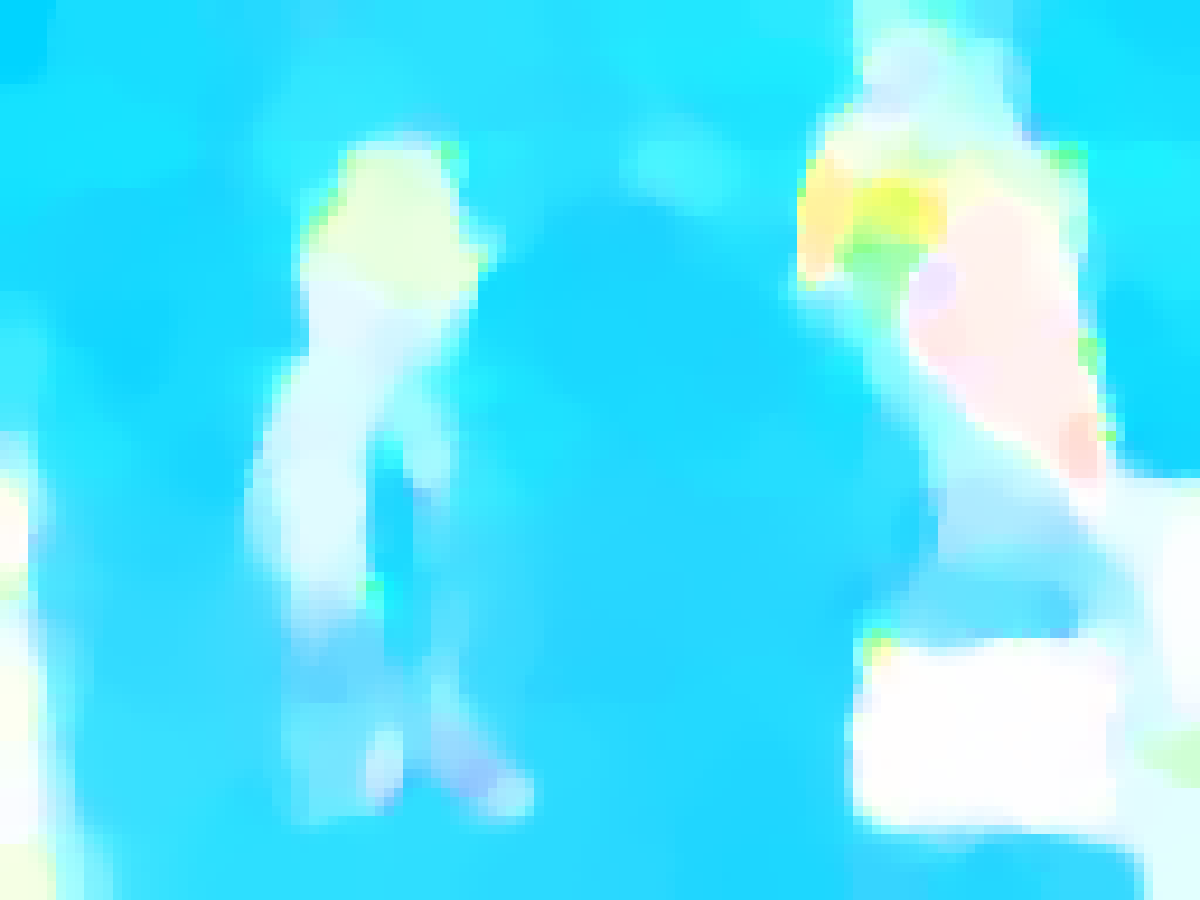}
    \end{minipage}
    &
    \begin{minipage}{.1\textwidth}
      \vspace*{0.05\textwidth}
      \includegraphics[width=\linewidth, height=10mm]{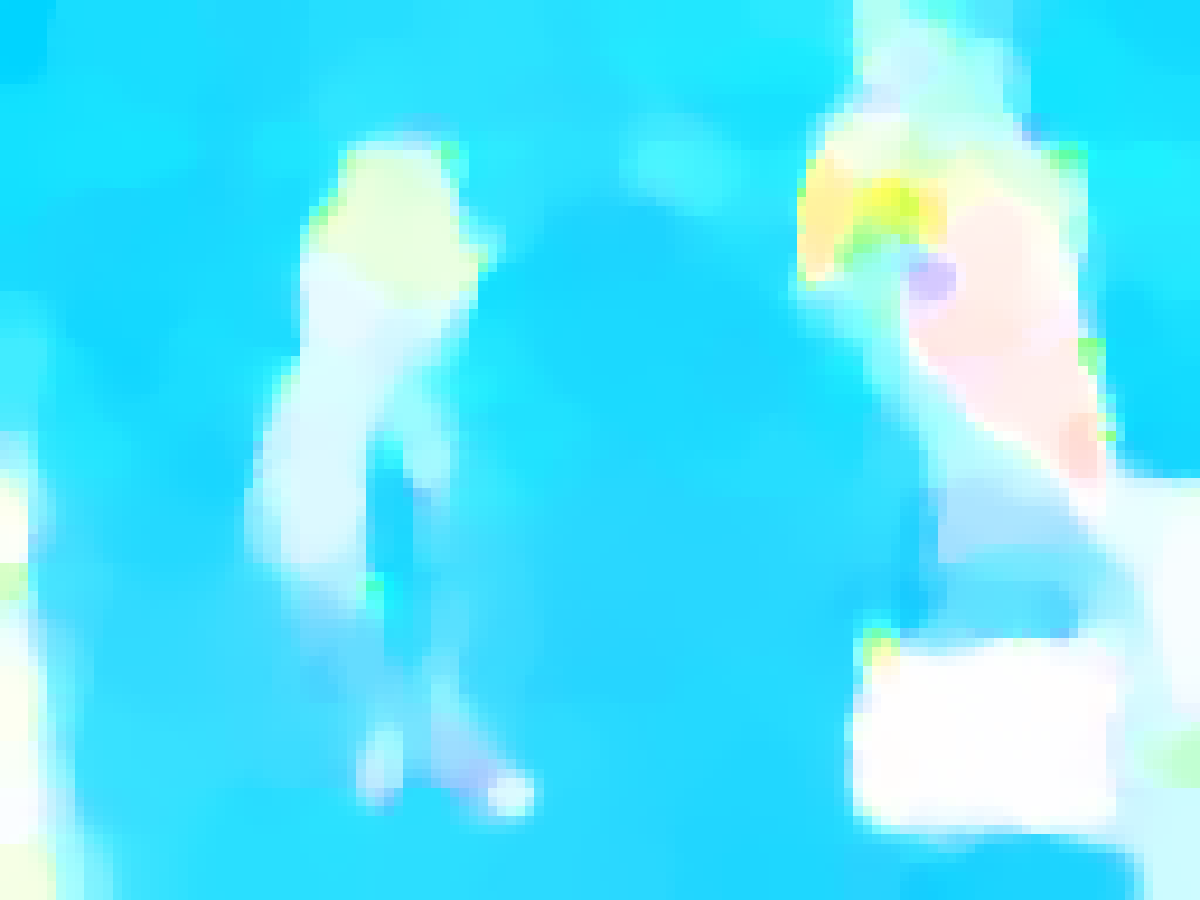}
    \end{minipage}
    \\
    
    \begin{minipage}{.1\textwidth}
      \vspace*{0.05\textwidth}
      \includegraphics[width=\linewidth, height=10mm]{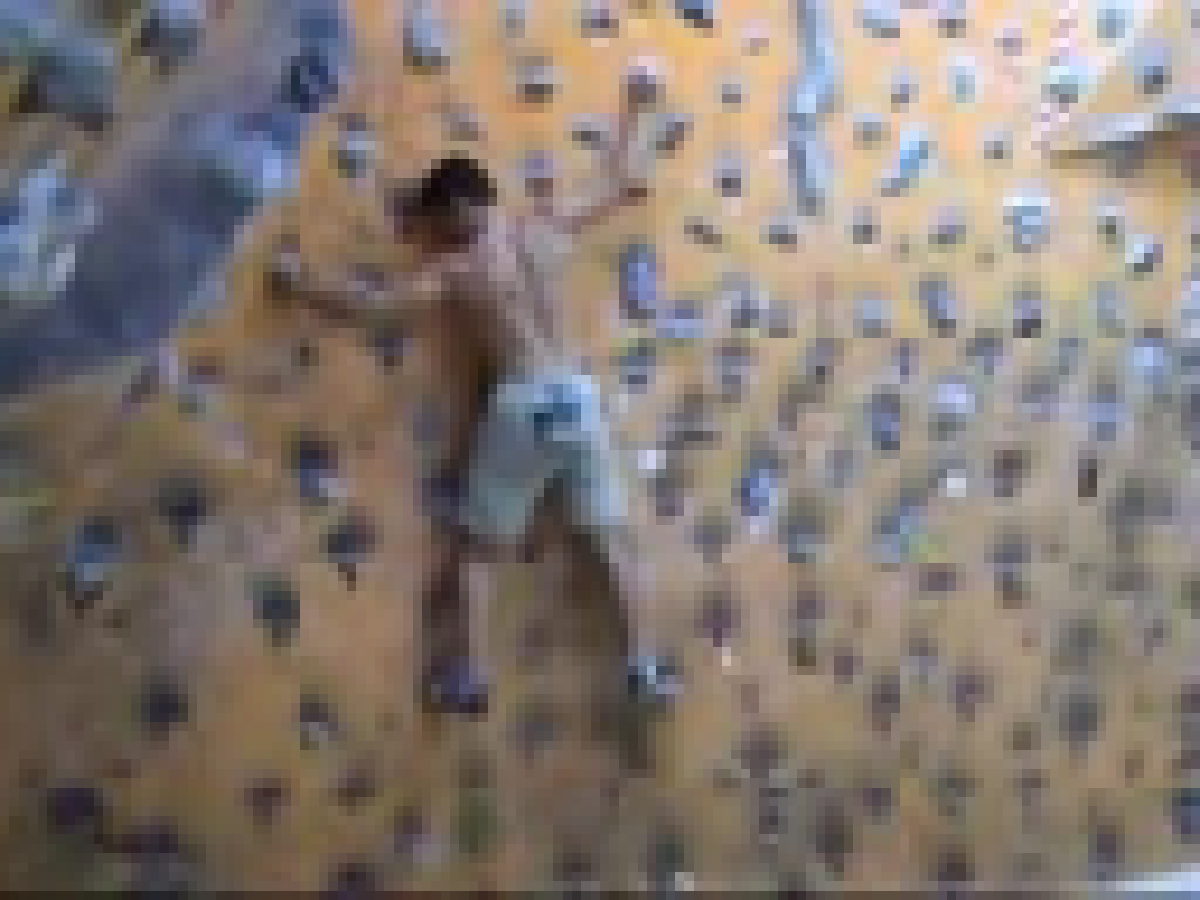}
    \end{minipage}
    &
    \begin{minipage}{.1\textwidth}
      \vspace*{0.05\textwidth}
      \includegraphics[width=\linewidth, height=10mm]{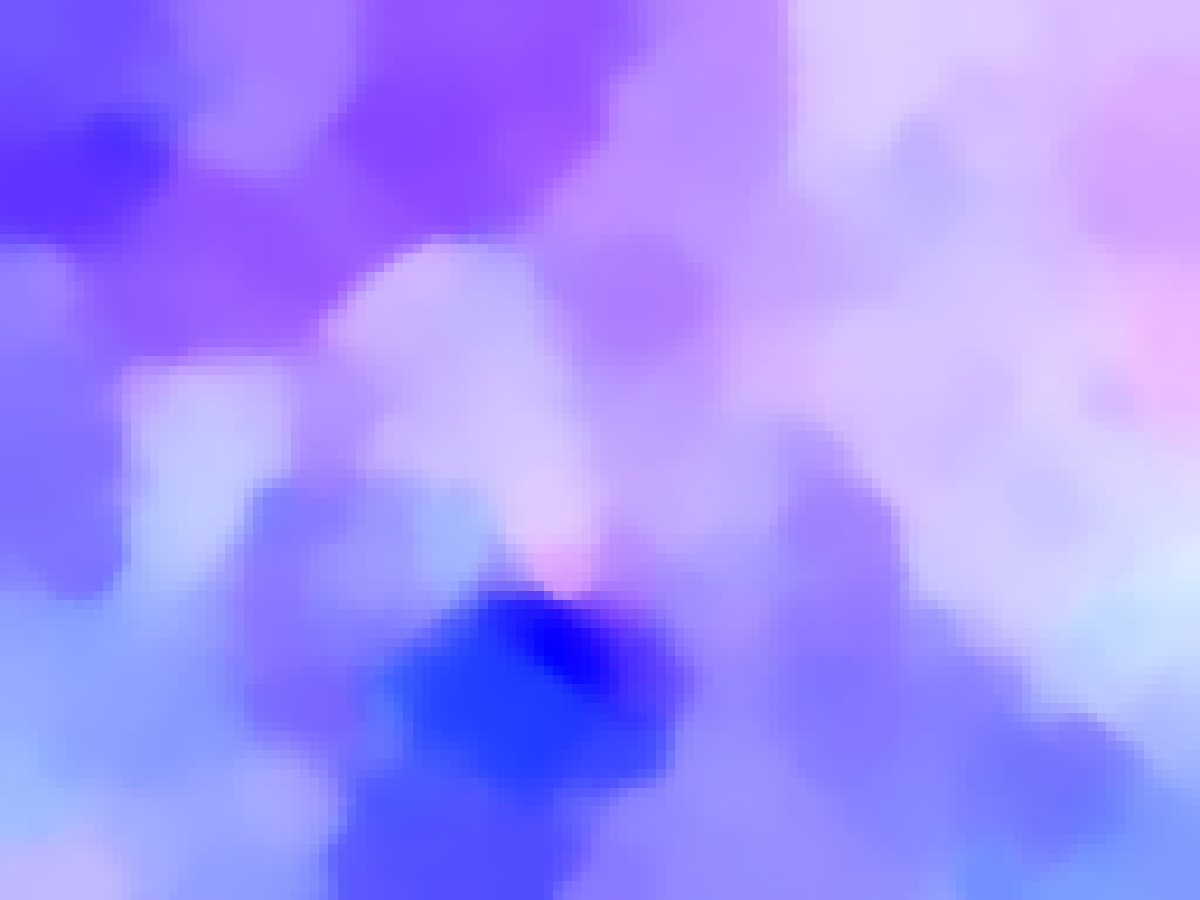}
    \end{minipage}
    & 
    \begin{minipage}{.1\textwidth}
      \vspace*{0.05\textwidth}
      \includegraphics[width=\linewidth, height=10mm]{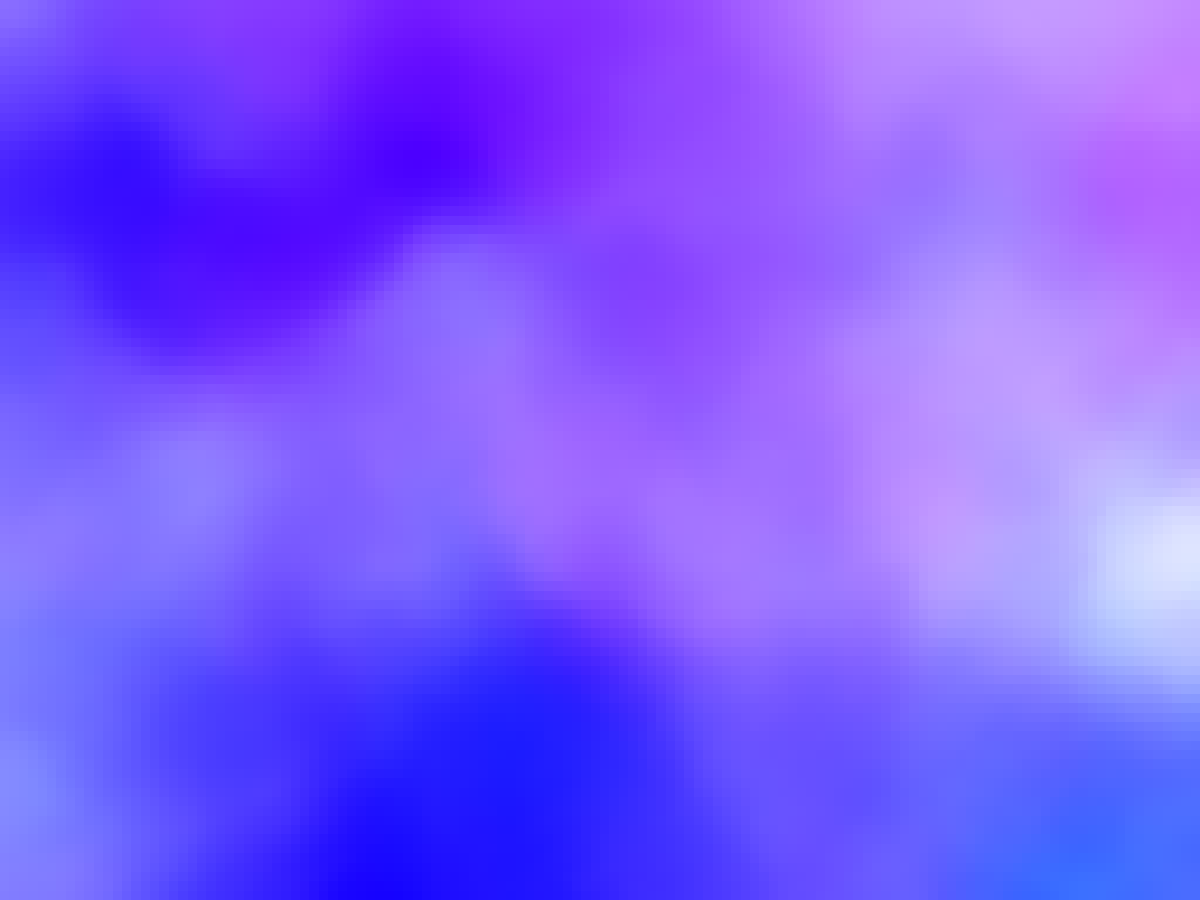}
    \end{minipage}
    &
    \begin{minipage}{.1\textwidth}
      \vspace*{0.05\textwidth}
      \includegraphics[width=\linewidth, height=10mm]{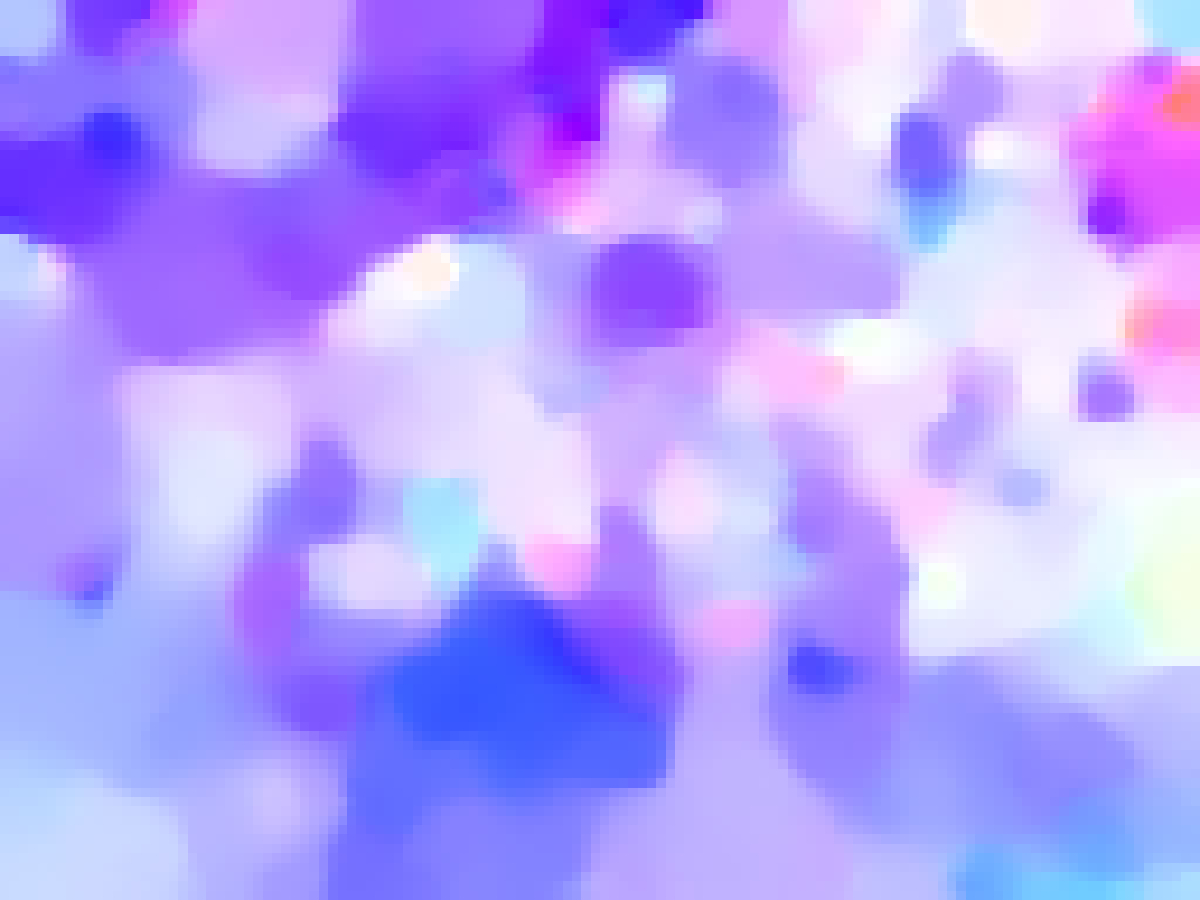}
    \end{minipage}
    &
    \begin{minipage}{.1\textwidth}
      \vspace*{0.05\textwidth}
      \includegraphics[width=\linewidth, height=10mm]{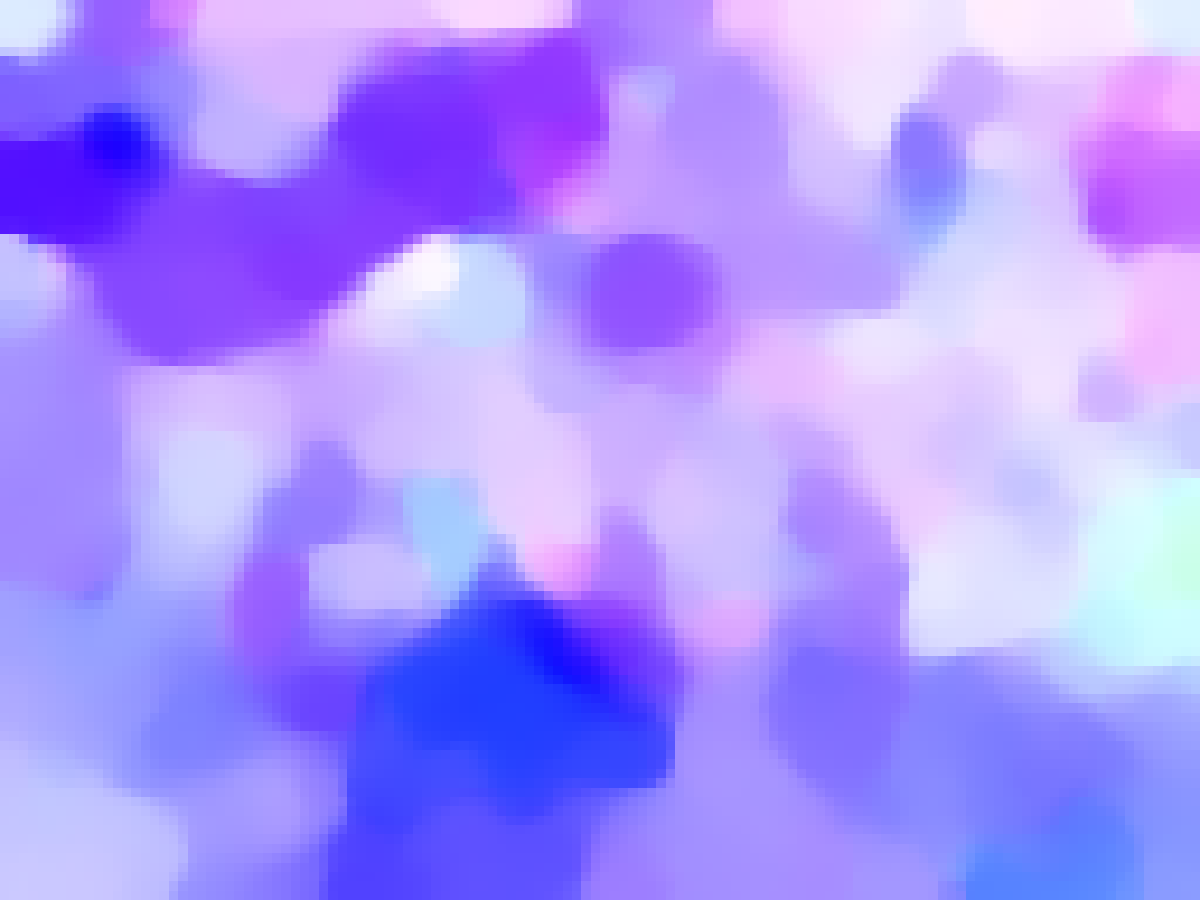}
    \end{minipage}
    &
    \begin{minipage}{.1\textwidth}
      \vspace*{0.05\textwidth}
      \includegraphics[width=\linewidth, height=10mm]{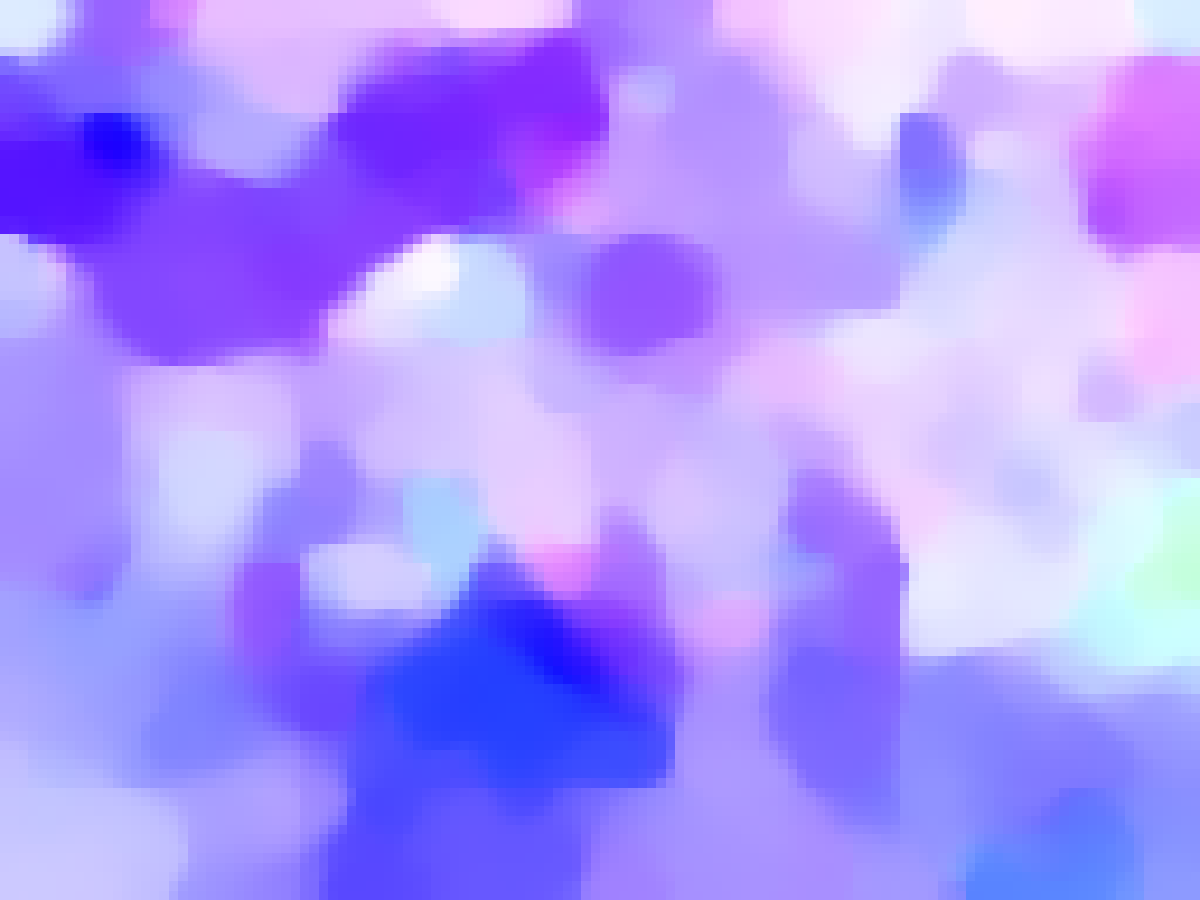}
    \end{minipage}
    \\
  \end{tabular}
  \caption{The motion field predicted by different motion estimation methods on several samples of UCF101.}\label{fig:MF_UCF}
\end{figure*}

 
\begin{figure*}
  
  \centering
  \begin{tabular}{  c  c  c  c  c  c  c }
    Images & Ground Truth & DeepFlow & EpicFlow & HAOF & LDOF & USCNN \\ 
    \begin{minipage}{.1\textwidth}
      \vspace*{0.05\textwidth}
      \vspace*{0.05\textwidth}
      \includegraphics[width=\linewidth, height=10mm]{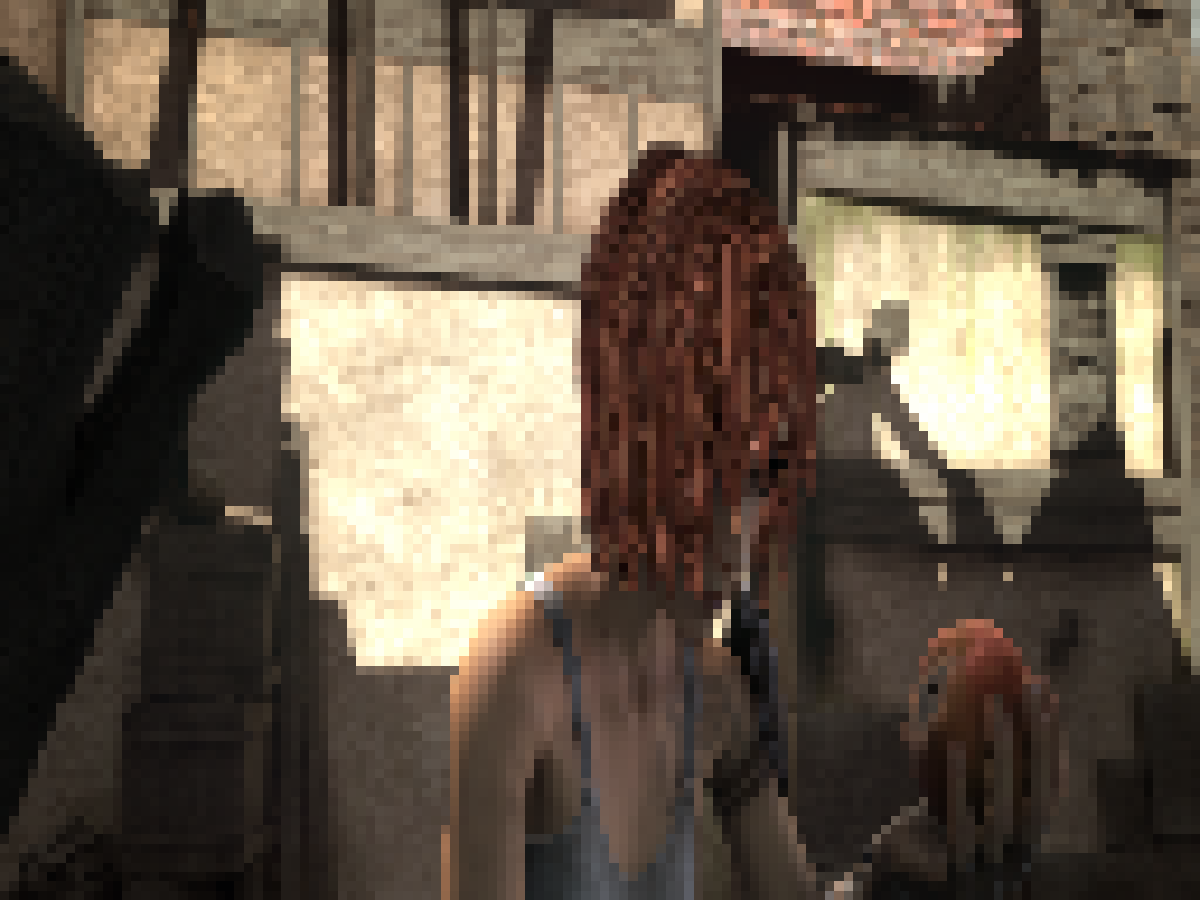}
    \end{minipage}
    &
    \begin{minipage}{.1\textwidth}
      \vspace*{0.05\textwidth}
      \vspace*{0.05\textwidth}
      \includegraphics[width=\linewidth, height=10mm]{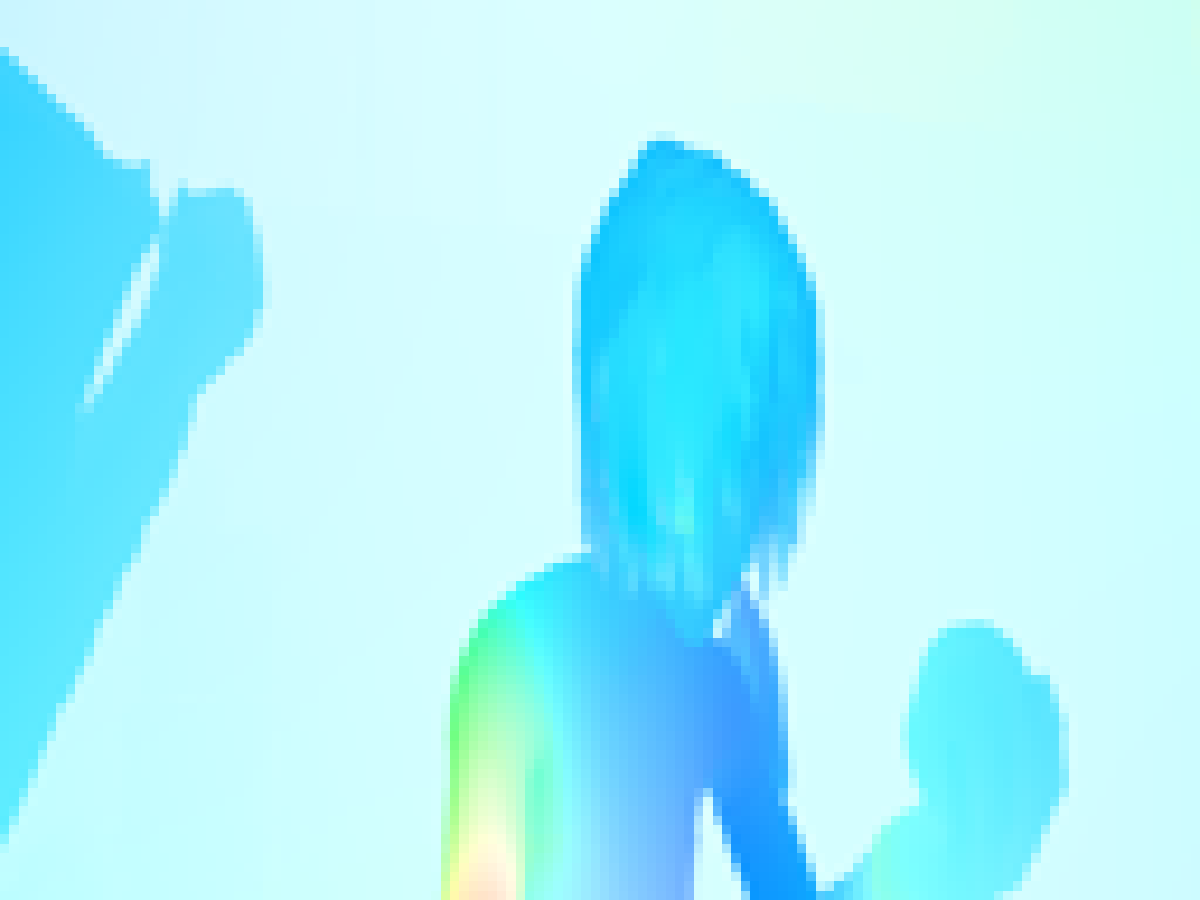}
    \end{minipage}
    &
    \begin{minipage}{.1\textwidth}
      \vspace*{0.05\textwidth}
      \vspace*{0.05\textwidth}
      \includegraphics[width=\linewidth, height=10mm]{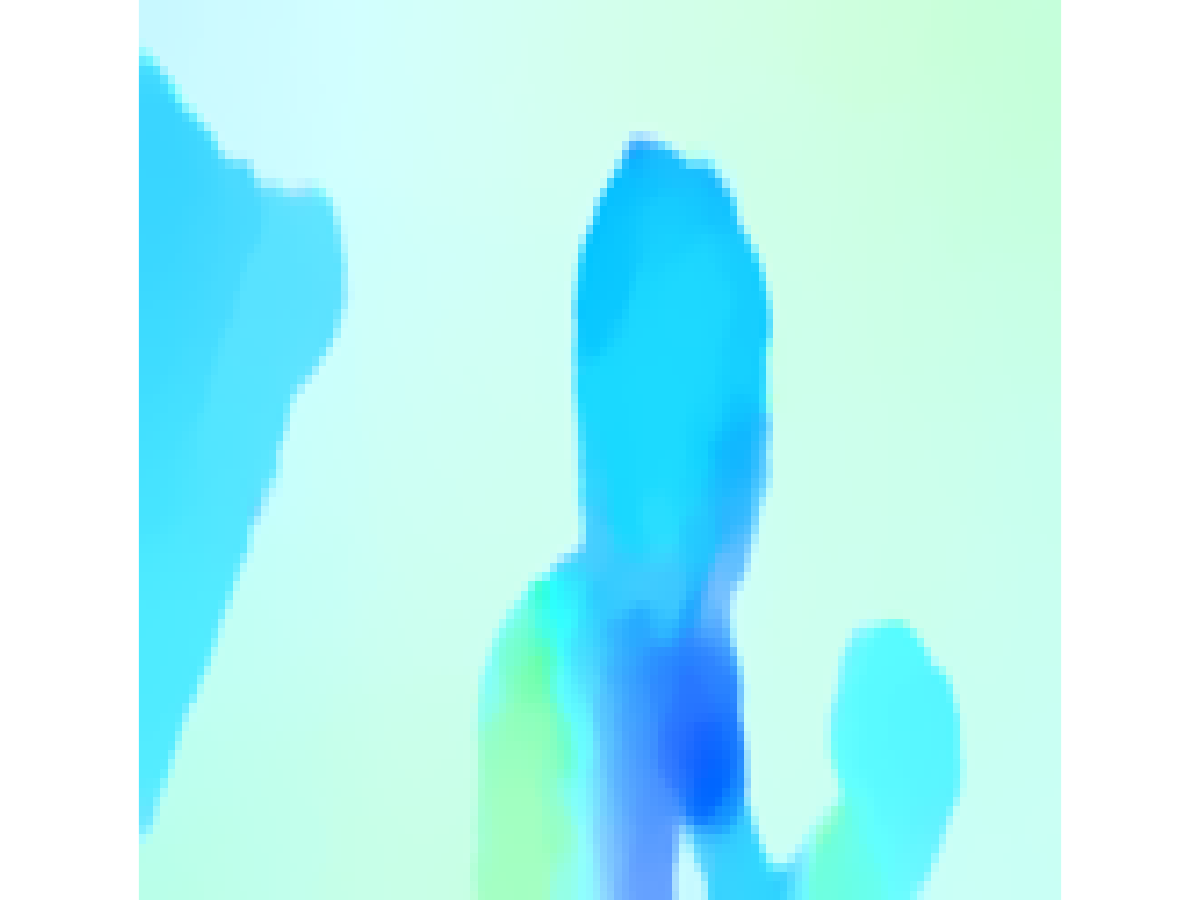}
    \end{minipage}
    & 
    \begin{minipage}{.1\textwidth}
      \vspace*{0.05\textwidth}
      \vspace*{0.05\textwidth}
      \includegraphics[width=\linewidth, height=10mm]{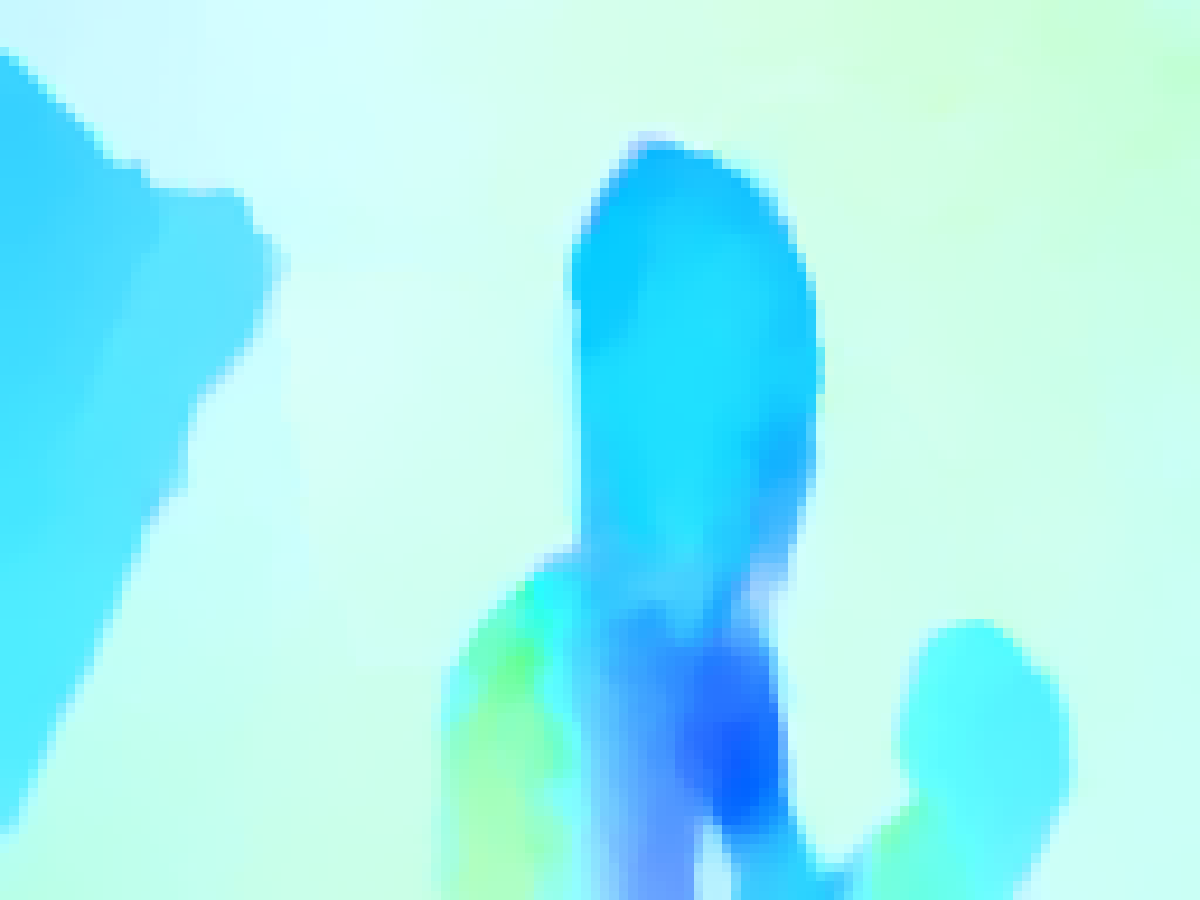}
    \end{minipage}
    &
    \begin{minipage}{.1\textwidth}
      \vspace*{0.05\textwidth}
      \vspace*{0.05\textwidth}
      \includegraphics[width=\linewidth, height=10mm]{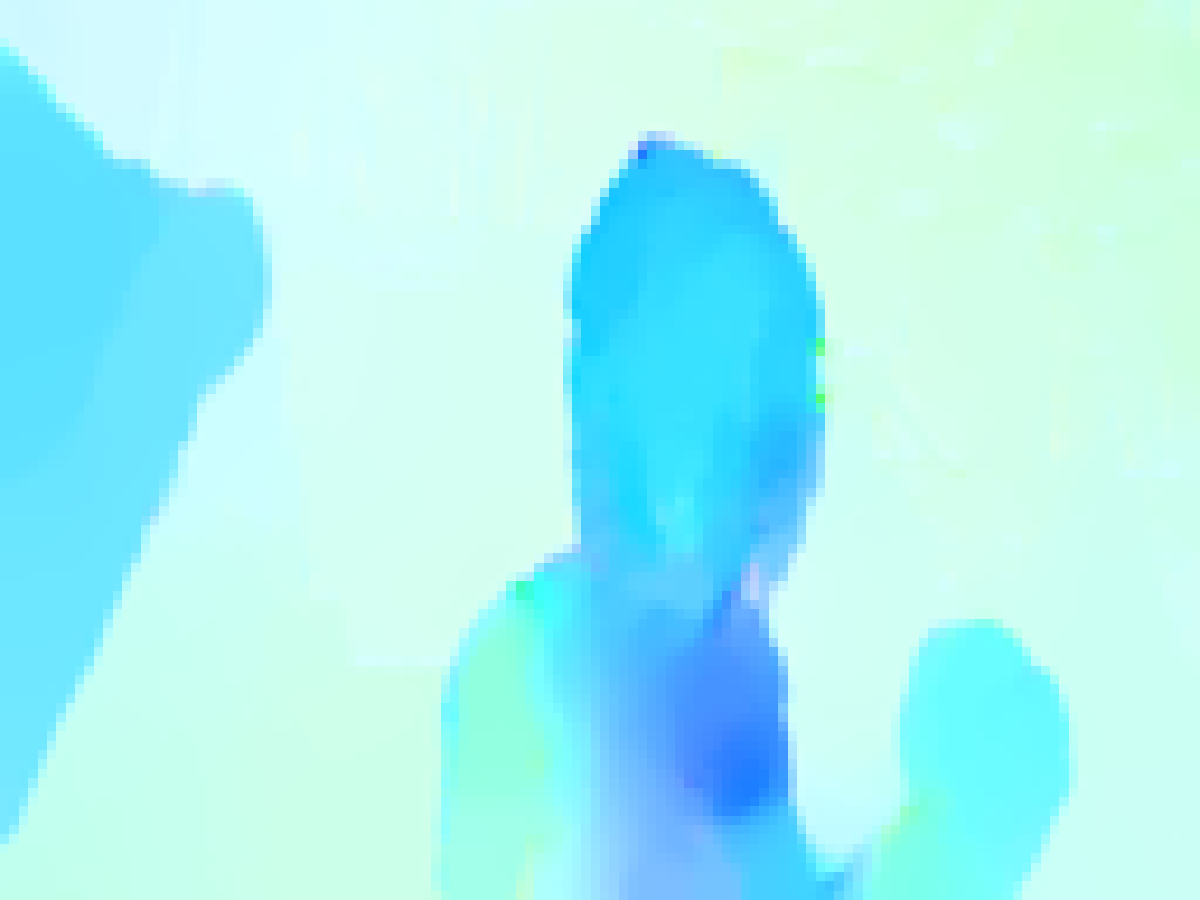}
    \end{minipage}
    &
    \begin{minipage}{.1\textwidth}
      \vspace*{0.05\textwidth}
      \vspace*{0.05\textwidth}
      \includegraphics[width=\linewidth, height=10mm]{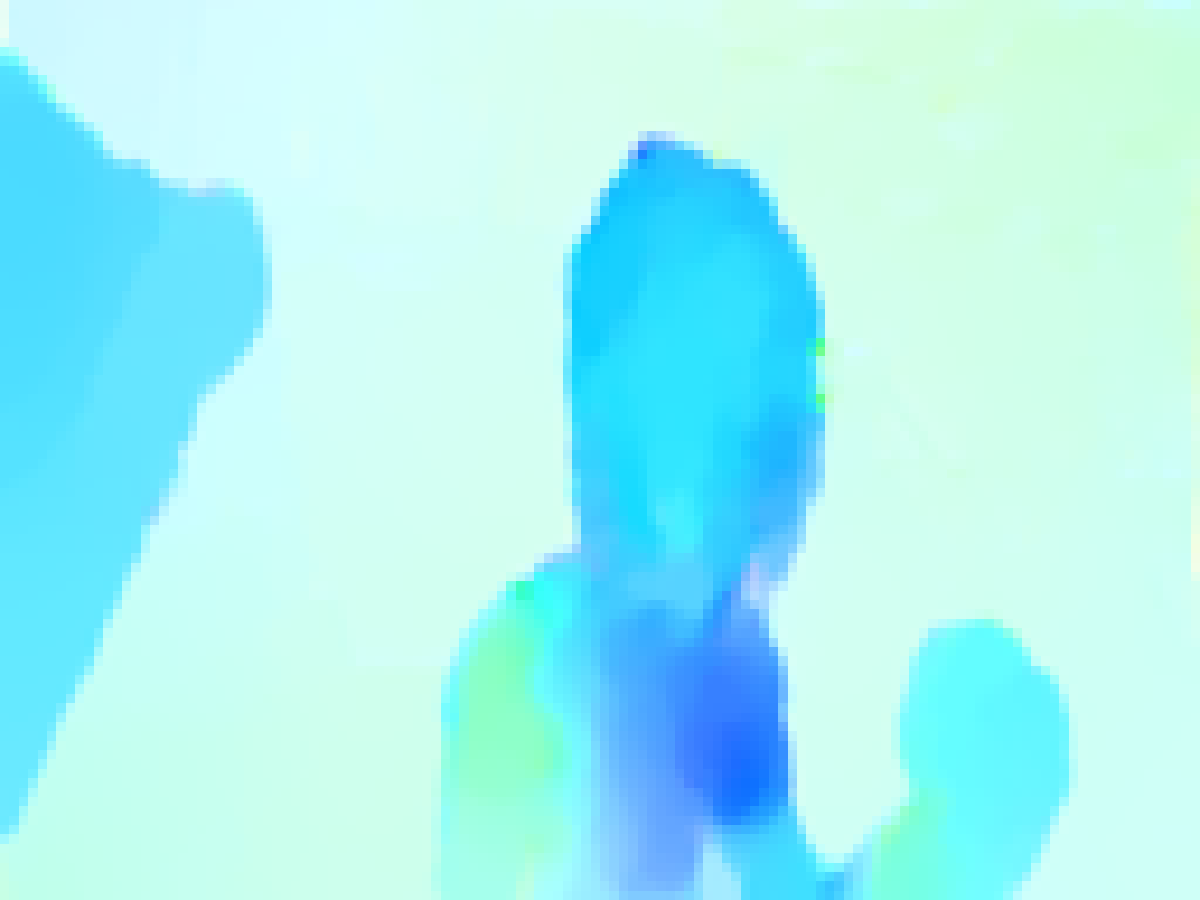}
    \end{minipage}
    &
    \begin{minipage}{.1\textwidth}
      \vspace*{0.05\textwidth}
      \vspace*{0.05\textwidth}
      \includegraphics[width=\linewidth, height=10mm]{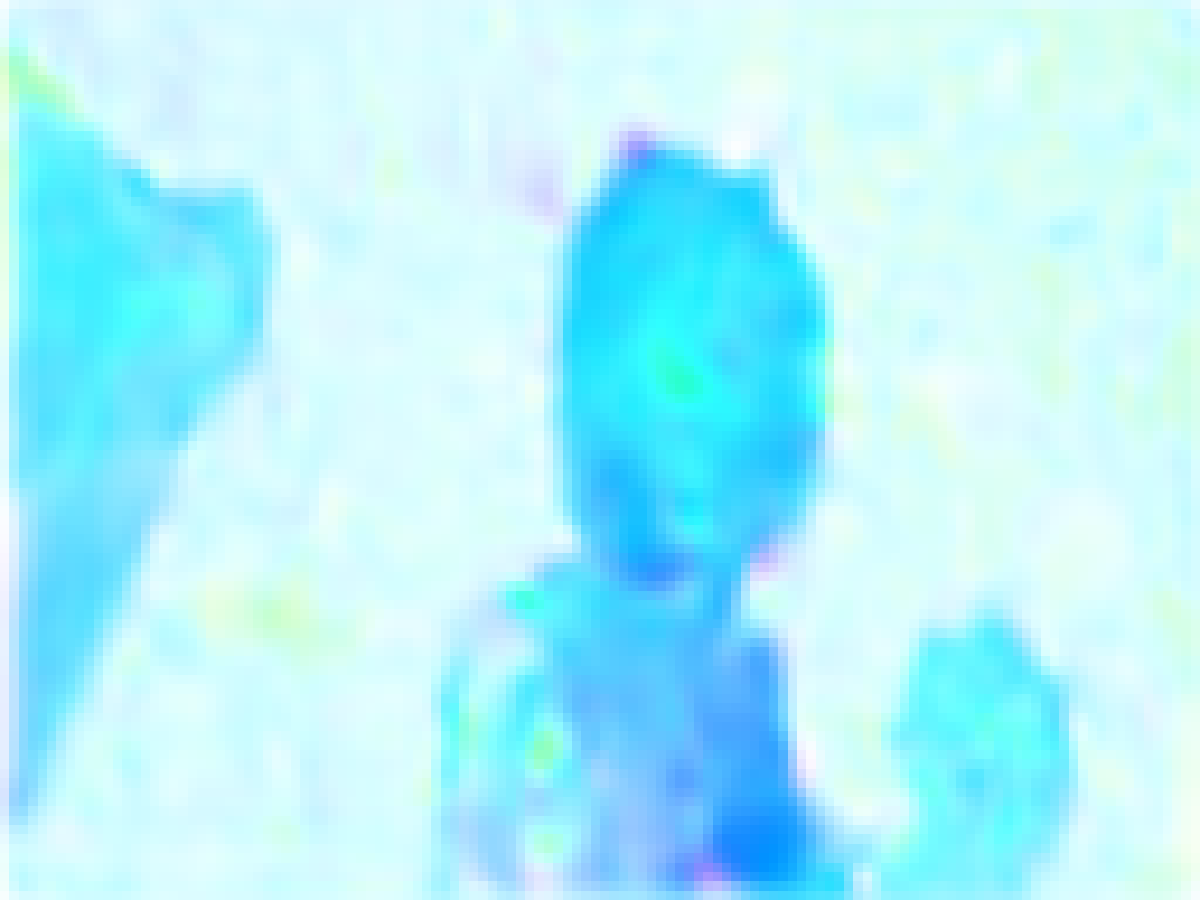}
    \end{minipage}
    
    \\
    
    \begin{minipage}{.1\textwidth}
      \vspace*{0.05\textwidth}
      \vspace*{0.05\textwidth}
      \includegraphics[width=\linewidth, height=10mm]{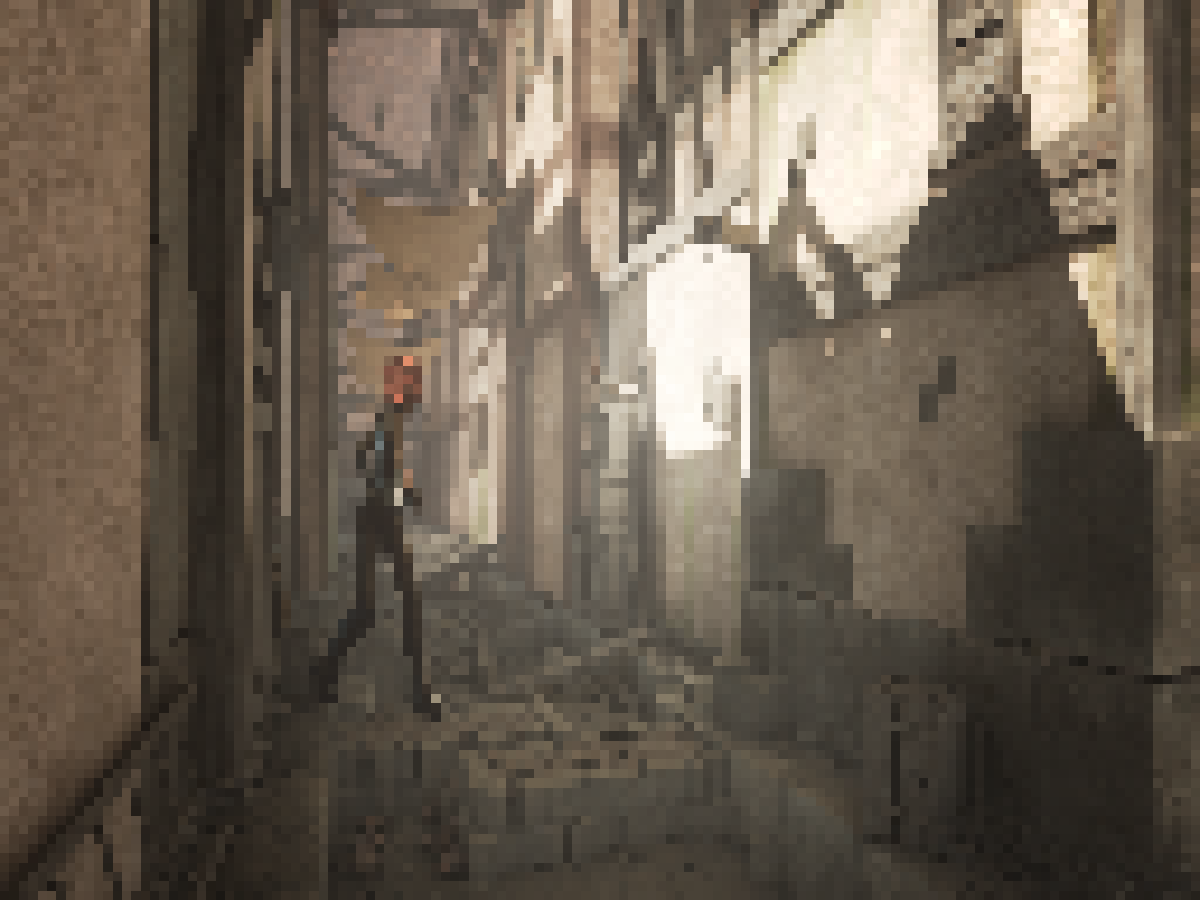}
    \end{minipage}
    &
    \begin{minipage}{.1\textwidth}
      \vspace*{0.05\textwidth}
      \vspace*{0.05\textwidth}
      \includegraphics[width=\linewidth, height=10mm]{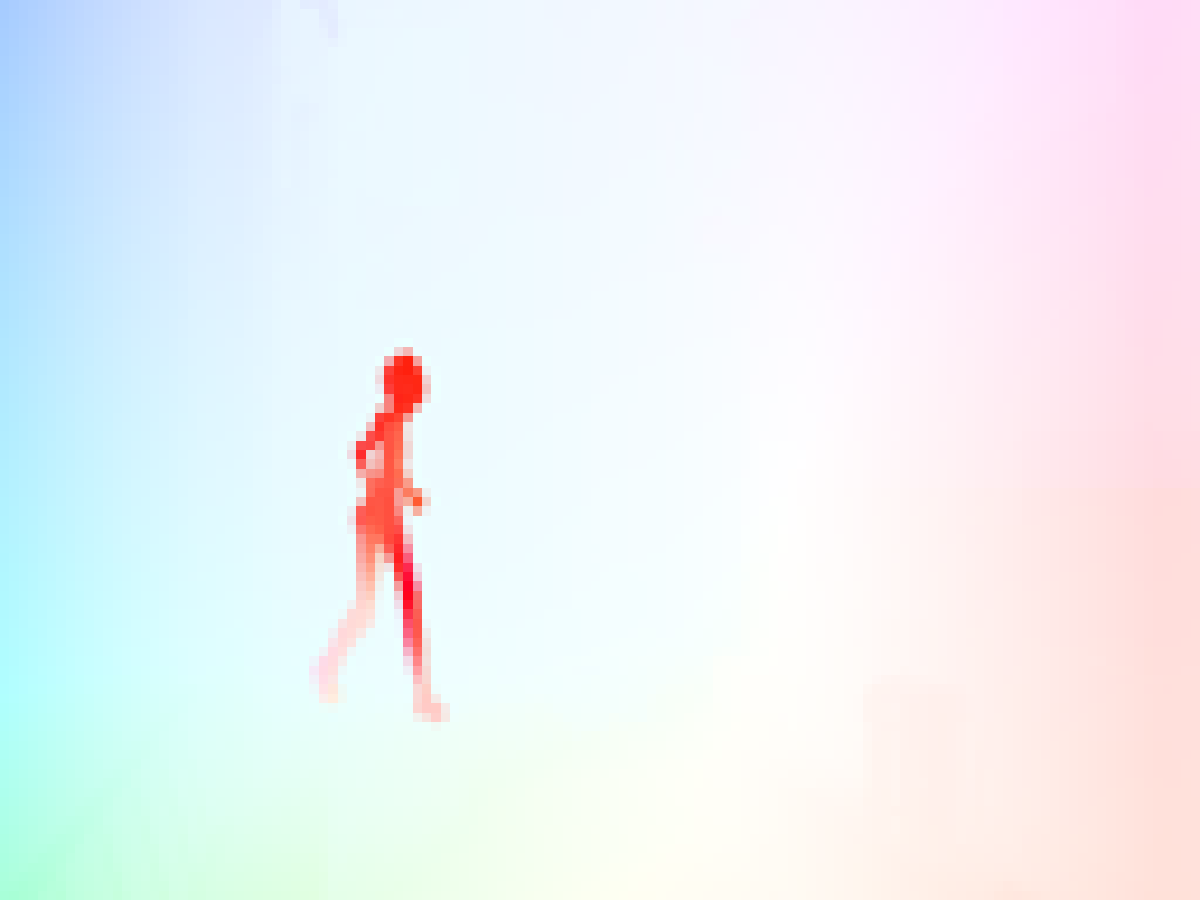}
    \end{minipage}
    &
    \begin{minipage}{.1\textwidth}
      \vspace*{0.05\textwidth}
      \vspace*{0.05\textwidth}
      \includegraphics[width=\linewidth, height=10mm]{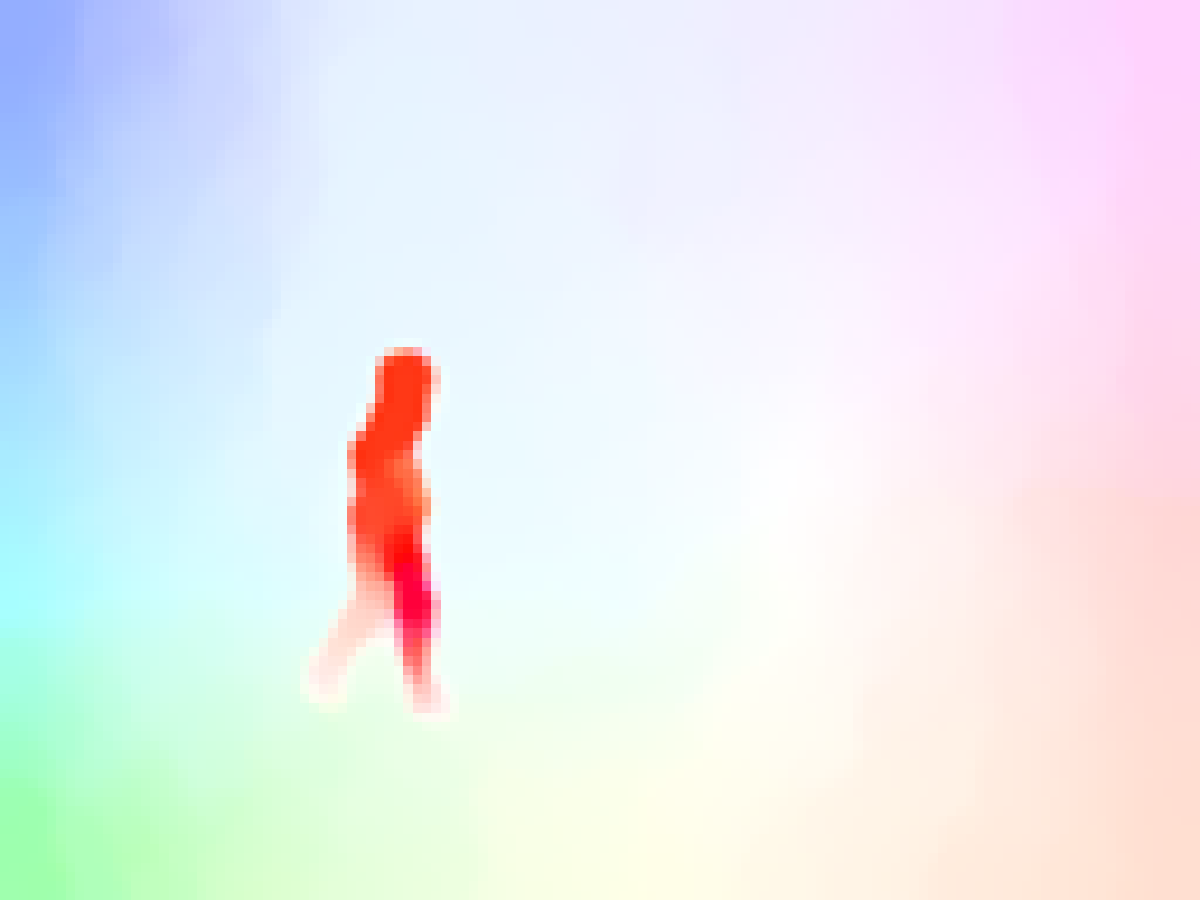}
    \end{minipage}
    & 
    \begin{minipage}{.1\textwidth}
      \vspace*{0.05\textwidth}
      \vspace*{0.05\textwidth}
      \includegraphics[width=\linewidth, height=10mm]{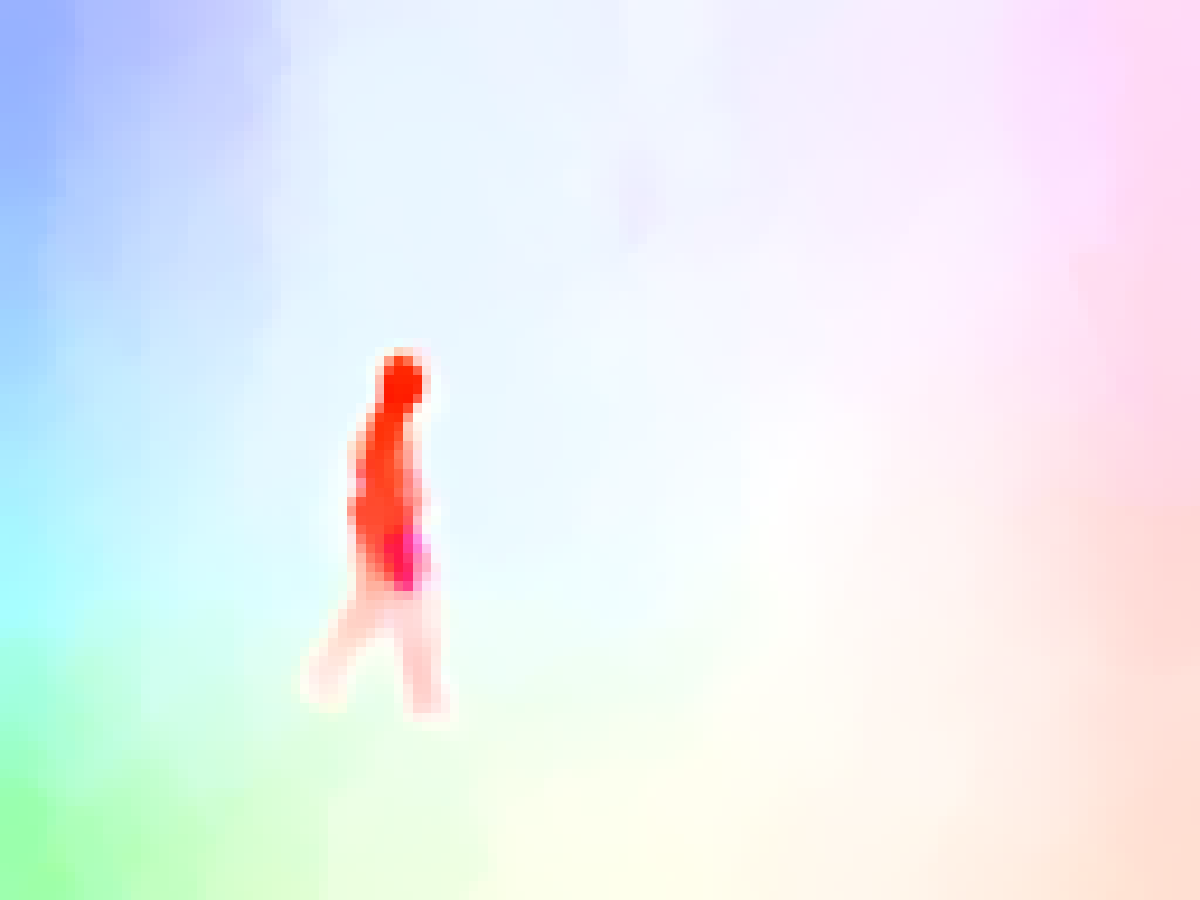}
    \end{minipage}
    &
    \begin{minipage}{.1\textwidth}
      \vspace*{0.05\textwidth}
      \includegraphics[width=\linewidth, height=10mm]{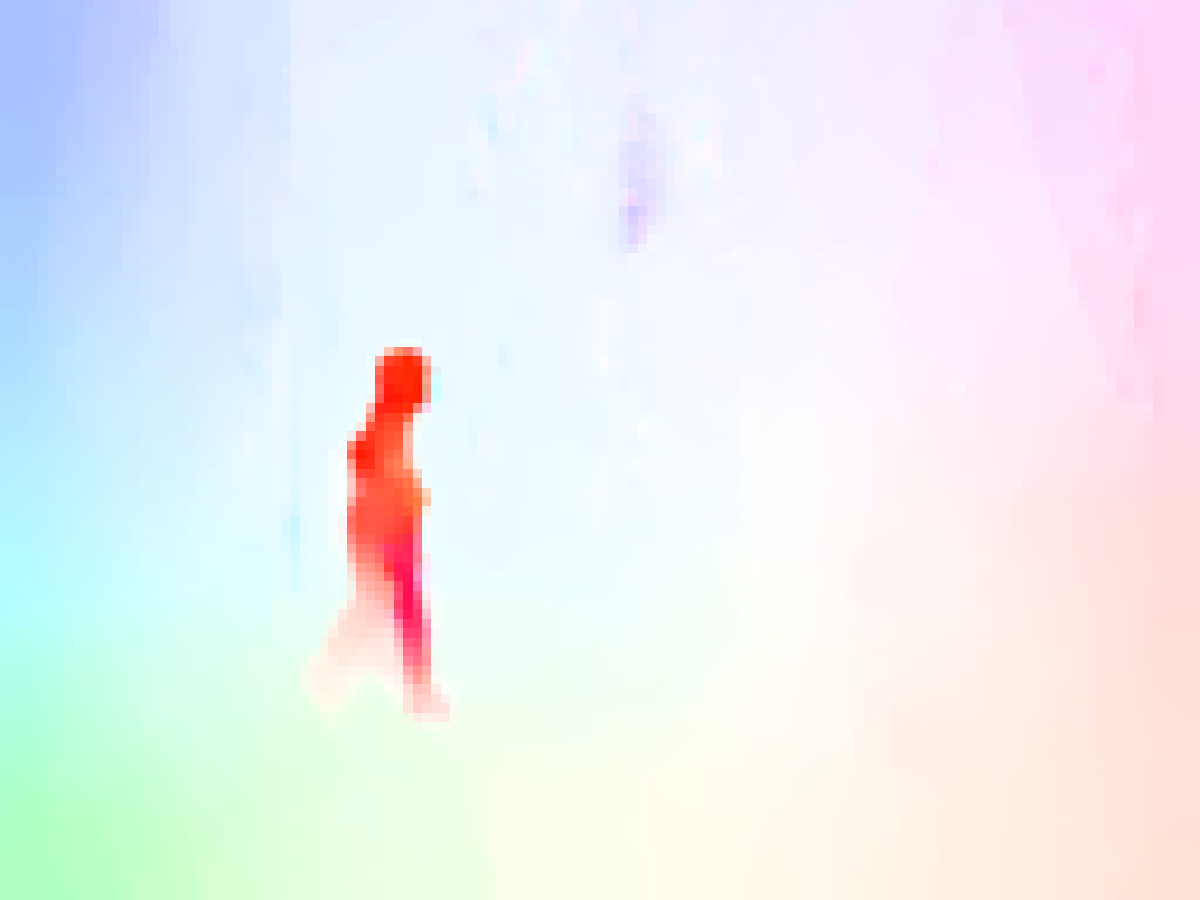}
    \end{minipage}
    &
    \begin{minipage}{.1\textwidth}
      \vspace*{0.05\textwidth}
      \includegraphics[width=\linewidth, height=10mm]{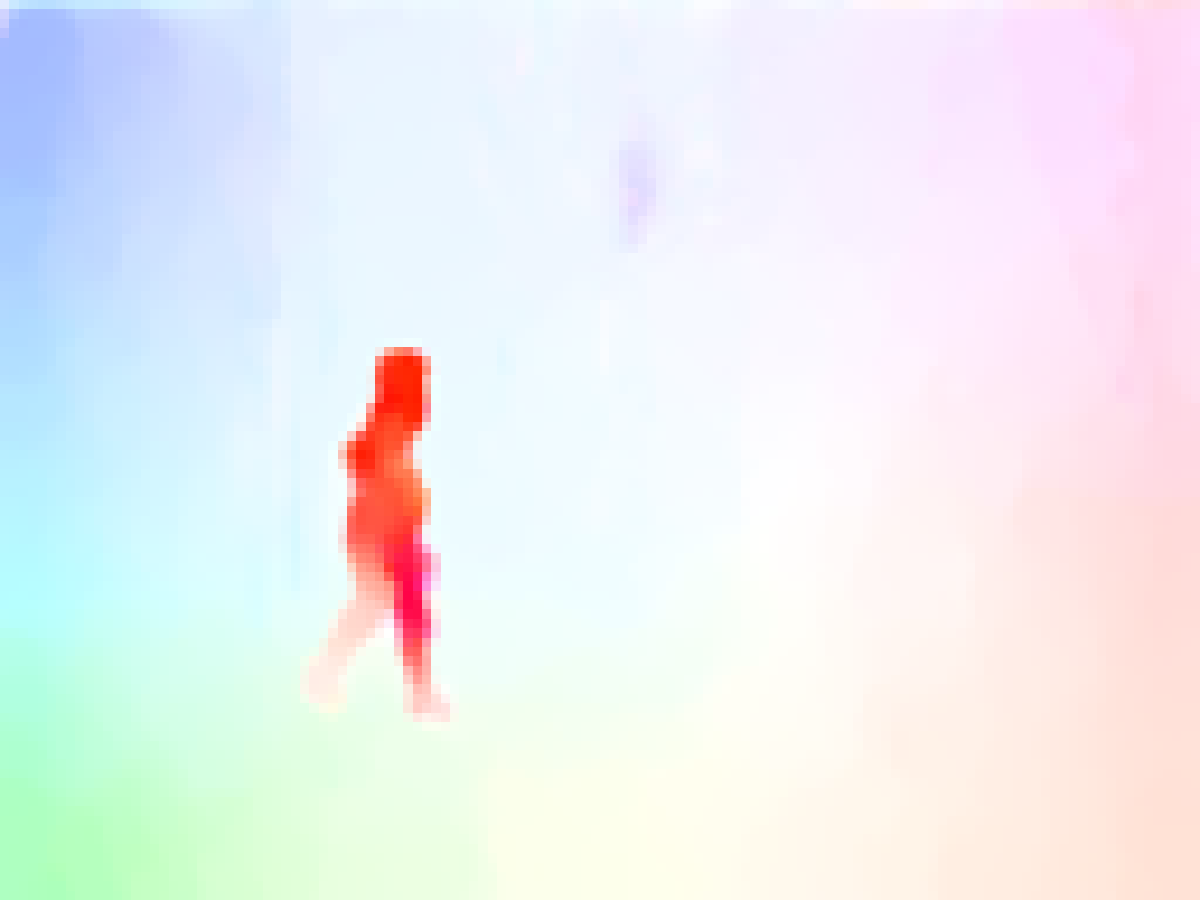}
    \end{minipage}
    &
    \begin{minipage}{.1\textwidth}
      \vspace*{0.05\textwidth}
      \includegraphics[width=\linewidth, height=10mm]{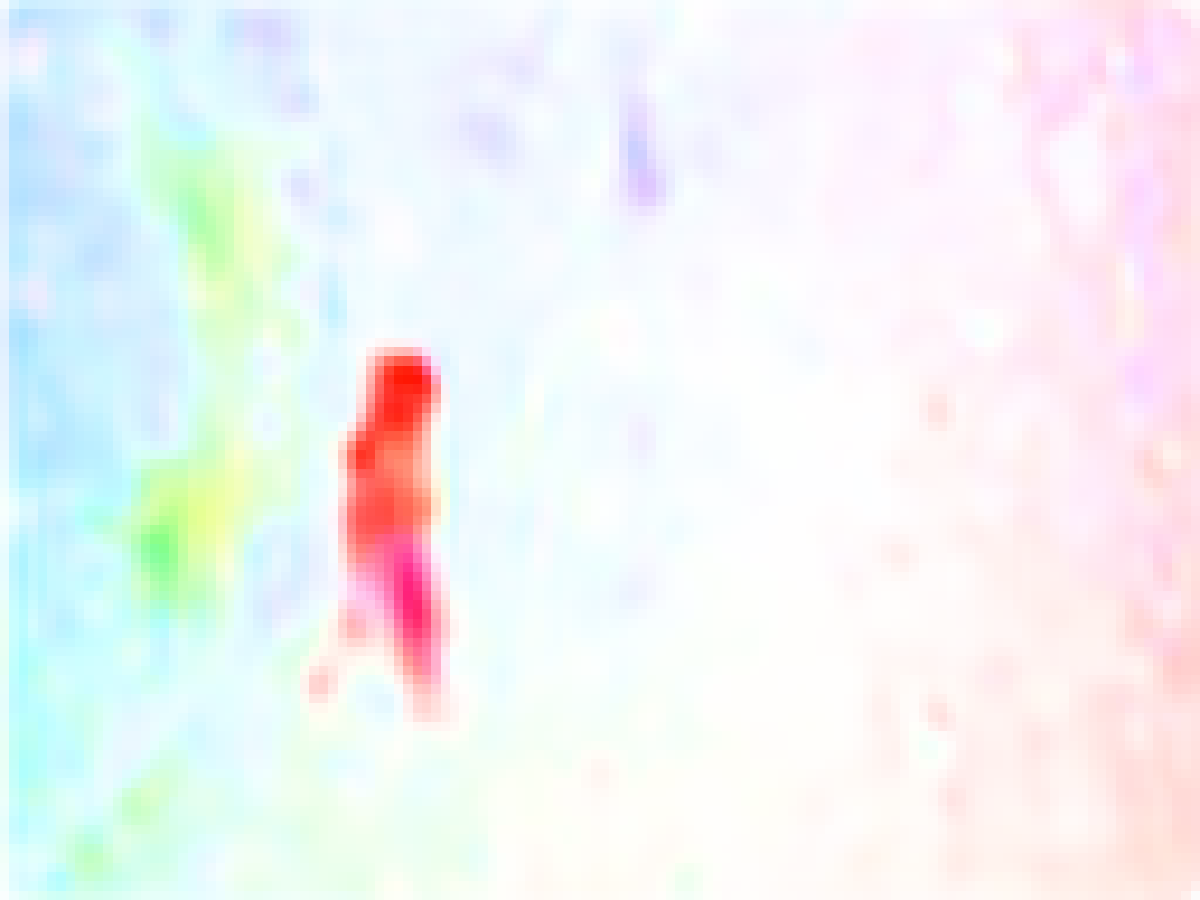}
    \end{minipage}
    
    \\
    
    \begin{minipage}{.1\textwidth}
      \vspace*{0.05\textwidth}
      \includegraphics[width=\linewidth, height=10mm]{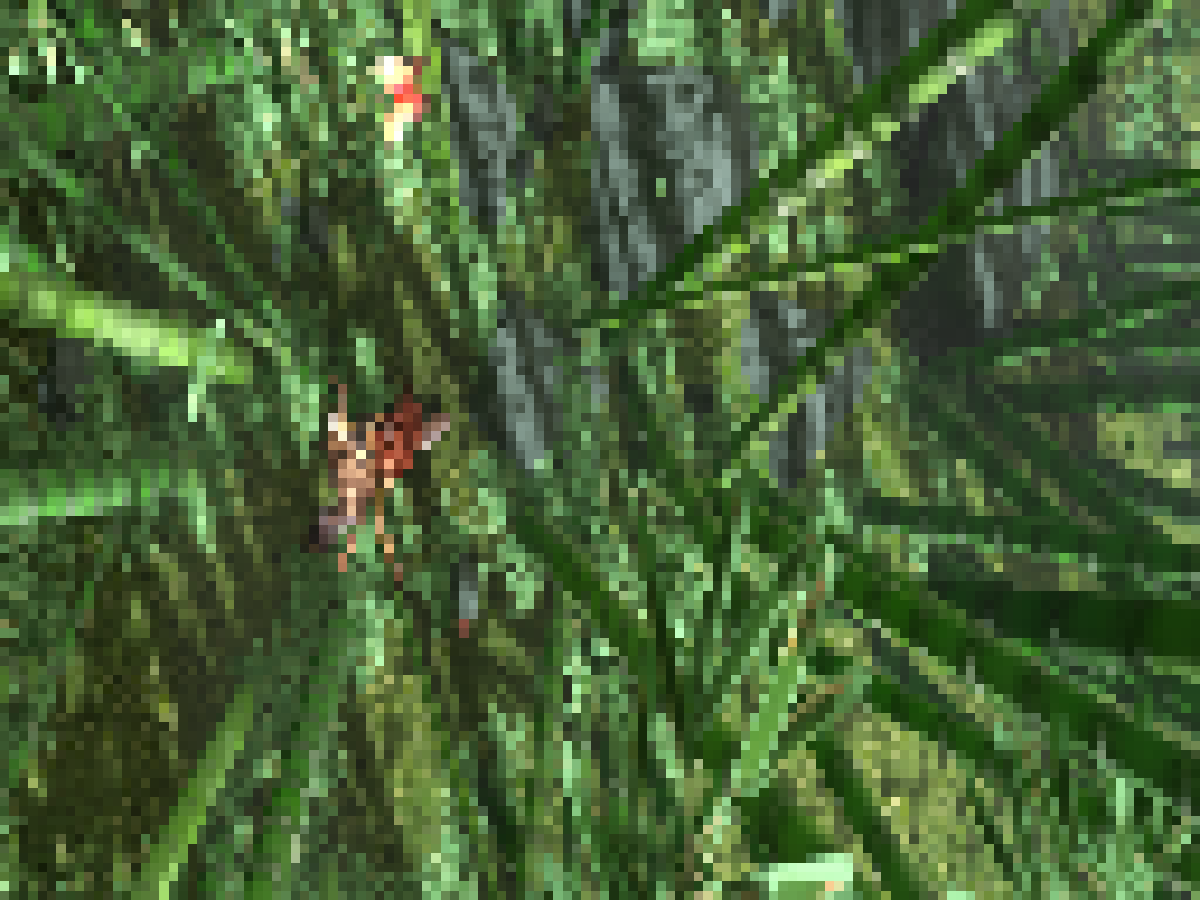}
    \end{minipage}
    &
    \begin{minipage}{.1\textwidth}
      \vspace*{0.05\textwidth}
      \includegraphics[width=\linewidth, height=10mm]{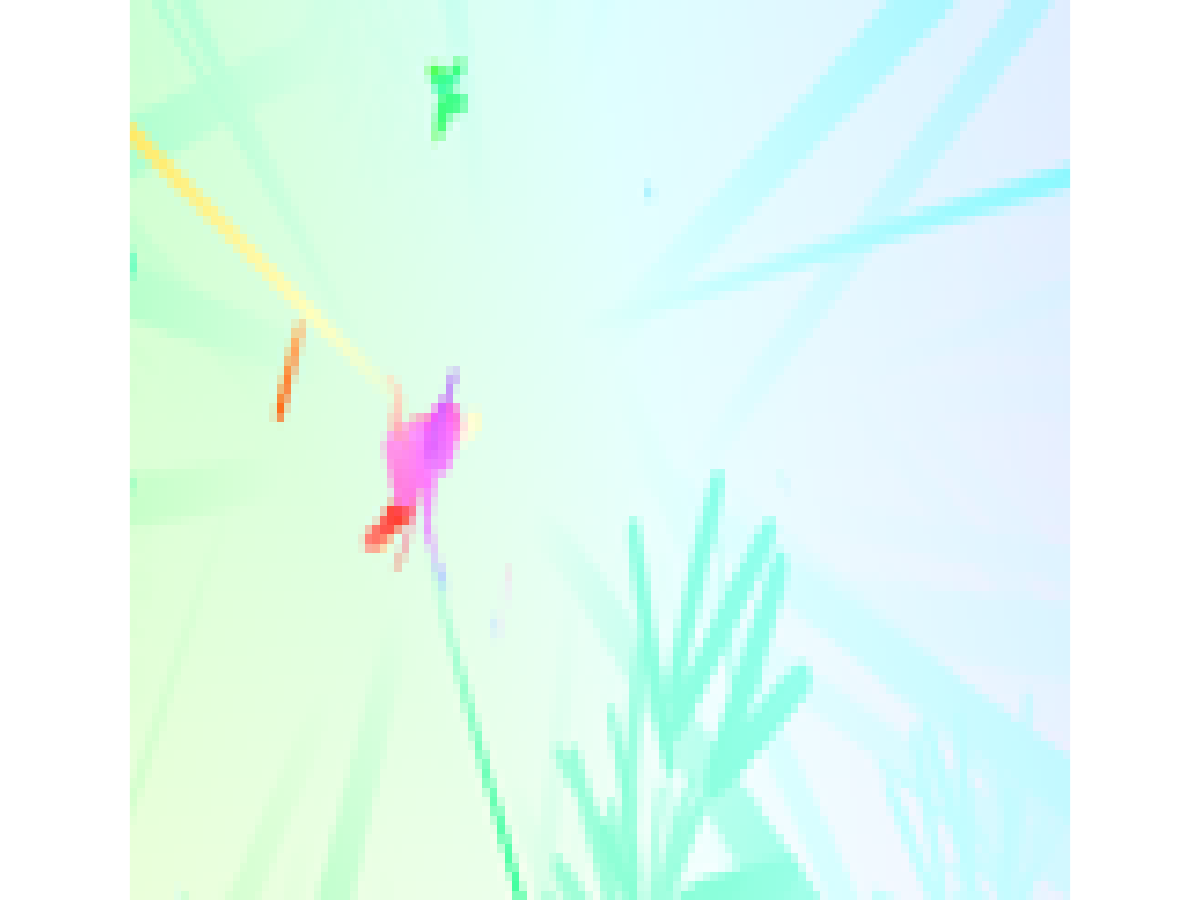}
    \end{minipage}
    &
    \begin{minipage}{.1\textwidth}
      \vspace*{0.05\textwidth}
      \includegraphics[width=\linewidth, height=10mm]{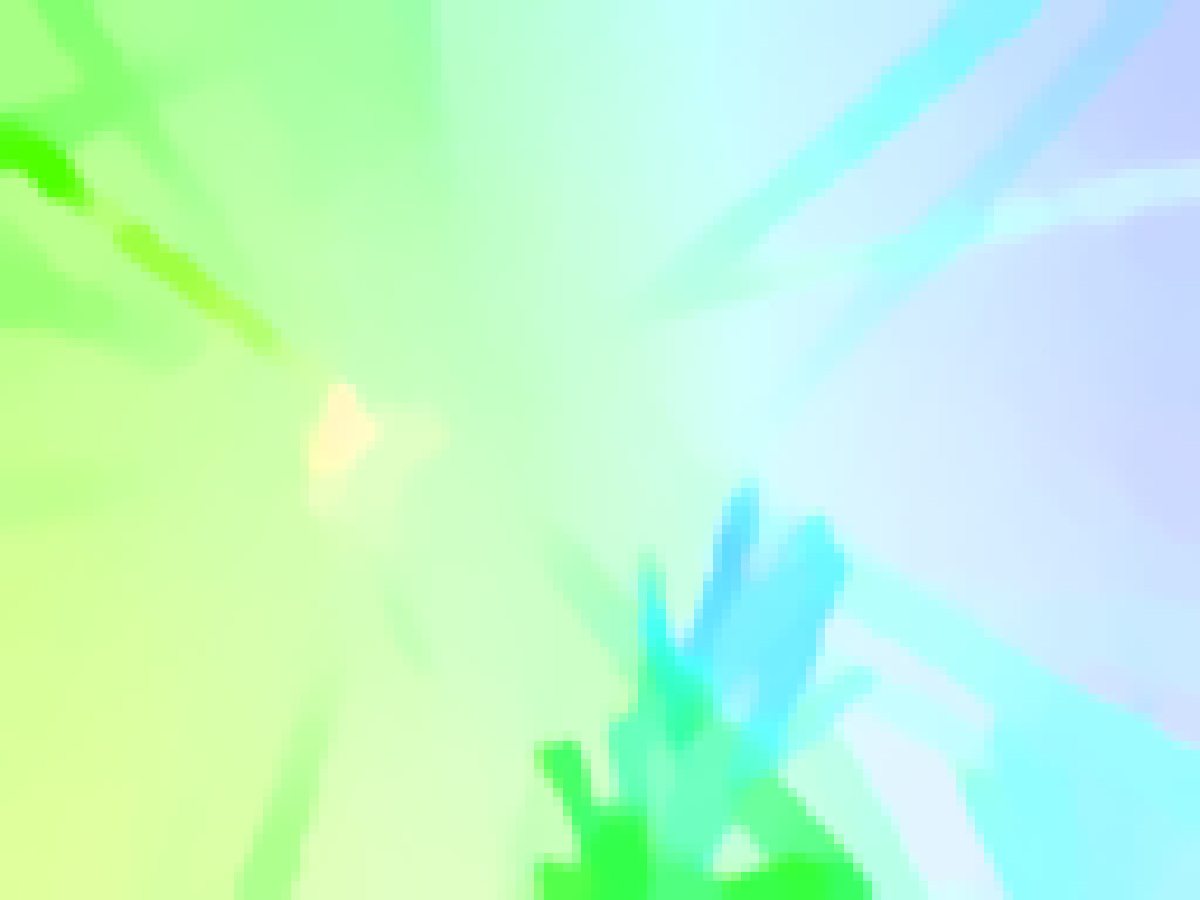}
    \end{minipage}
    & 
    \begin{minipage}{.1\textwidth}
      \vspace*{0.05\textwidth}
      \includegraphics[width=\linewidth, height=10mm]{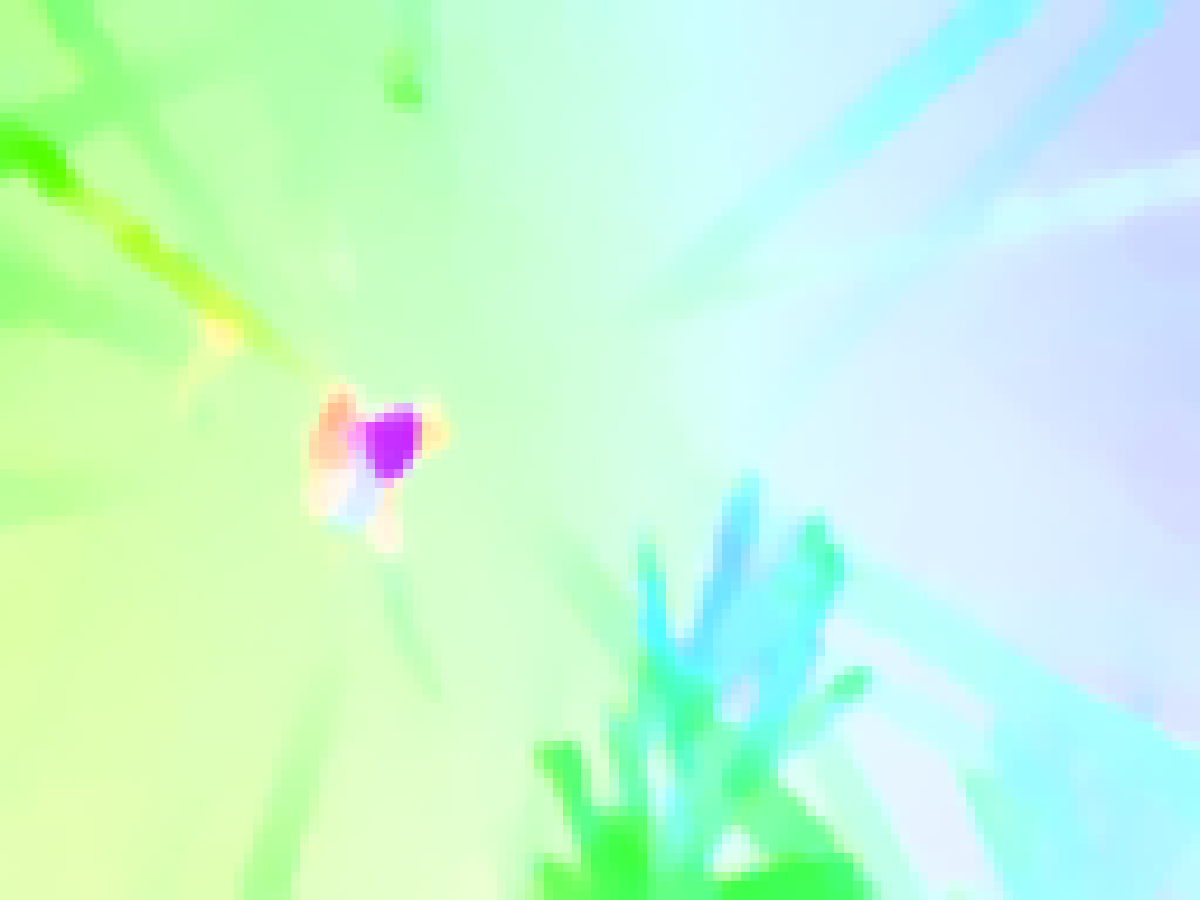}
    \end{minipage}
    &
    \begin{minipage}{.1\textwidth}
      \vspace*{0.05\textwidth}
      \includegraphics[width=\linewidth, height=10mm]{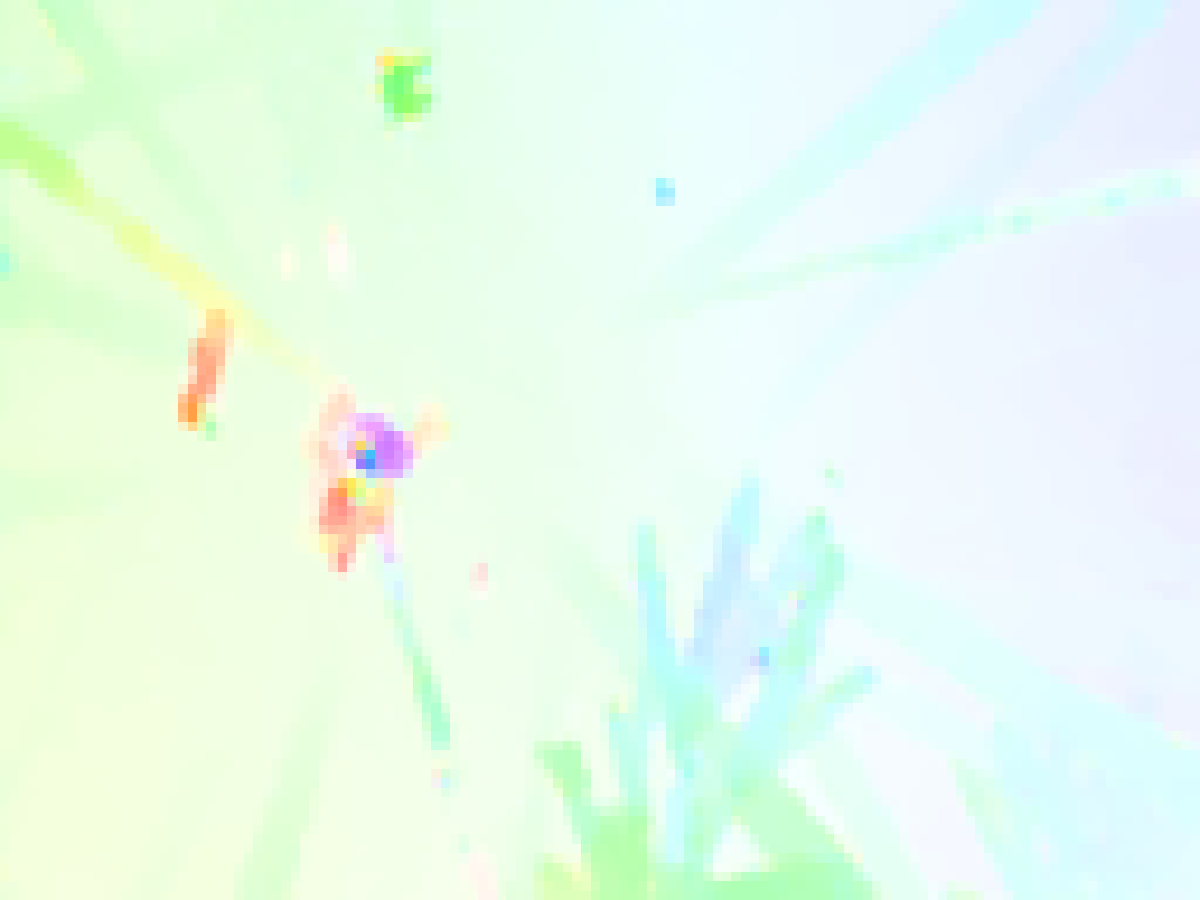}
    \end{minipage}
    &
    \begin{minipage}{.1\textwidth}
      \vspace*{0.05\textwidth}
      \includegraphics[width=\linewidth, height=10mm]{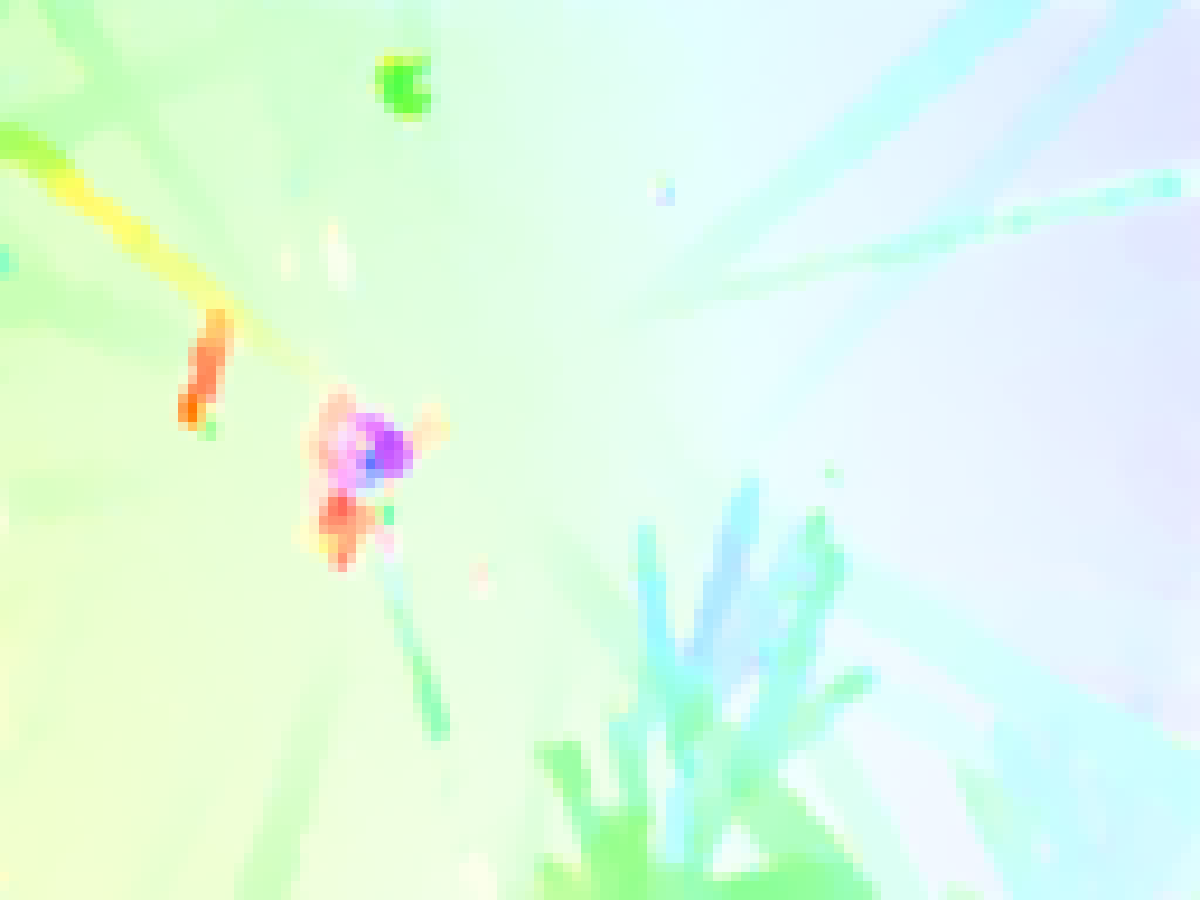}
    \end{minipage}
    &
    \begin{minipage}{.1\textwidth}
      \vspace*{0.05\textwidth}
      \includegraphics[width=\linewidth, height=10mm]{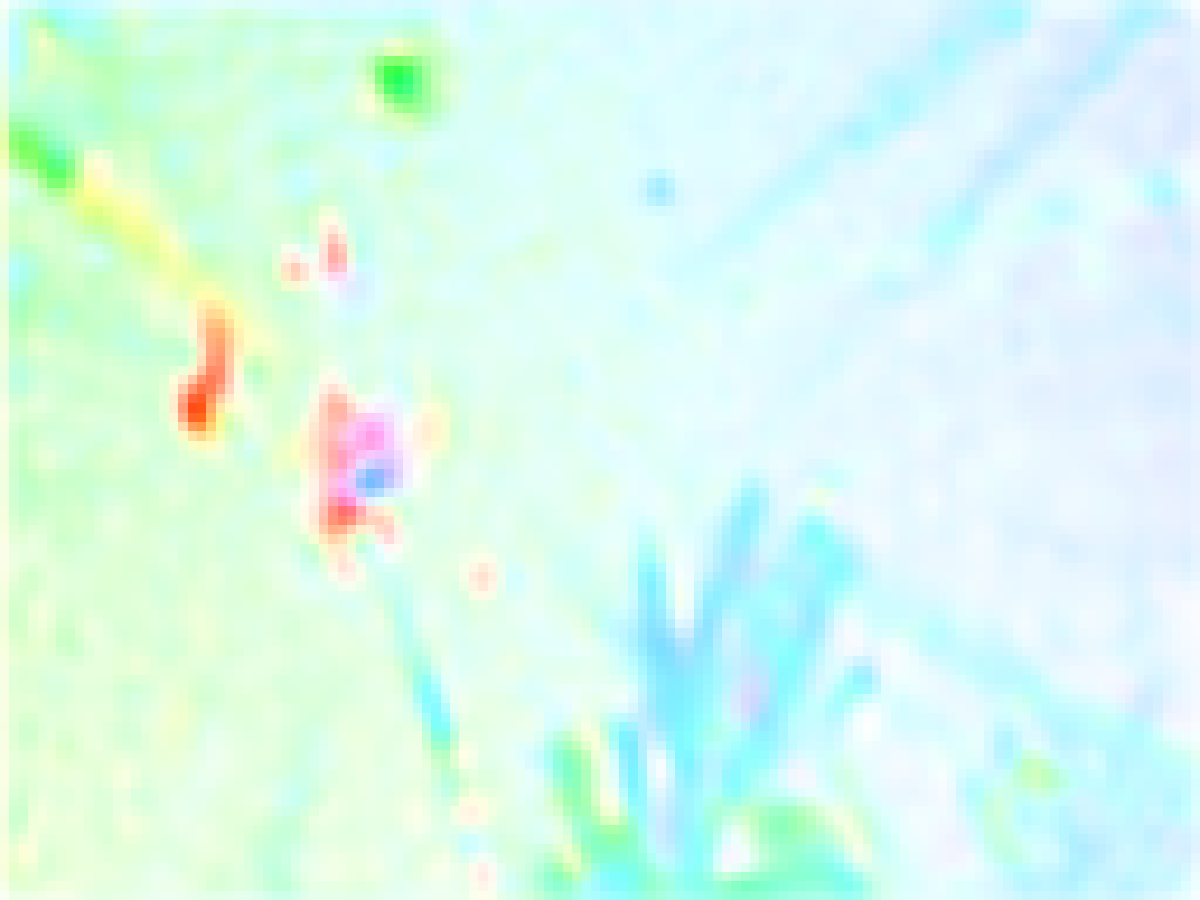}
    \end{minipage}
    
    \\
    
    \begin{minipage}{.1\textwidth}
      \vspace*{0.05\textwidth}
      \includegraphics[width=\linewidth, height=10mm]{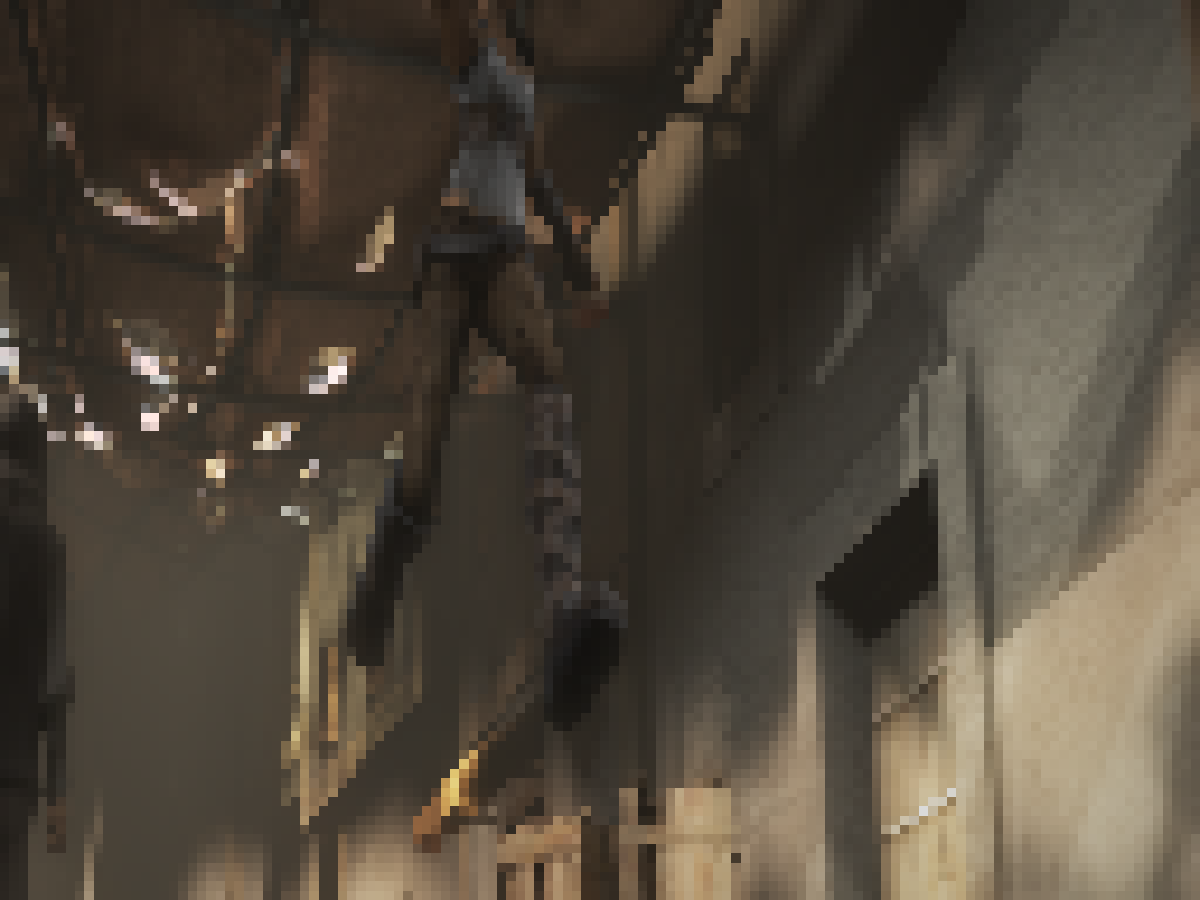}
    \end{minipage}
    &
    \begin{minipage}{.1\textwidth}
      \vspace*{0.05\textwidth}
      \includegraphics[width=\linewidth, height=10mm]{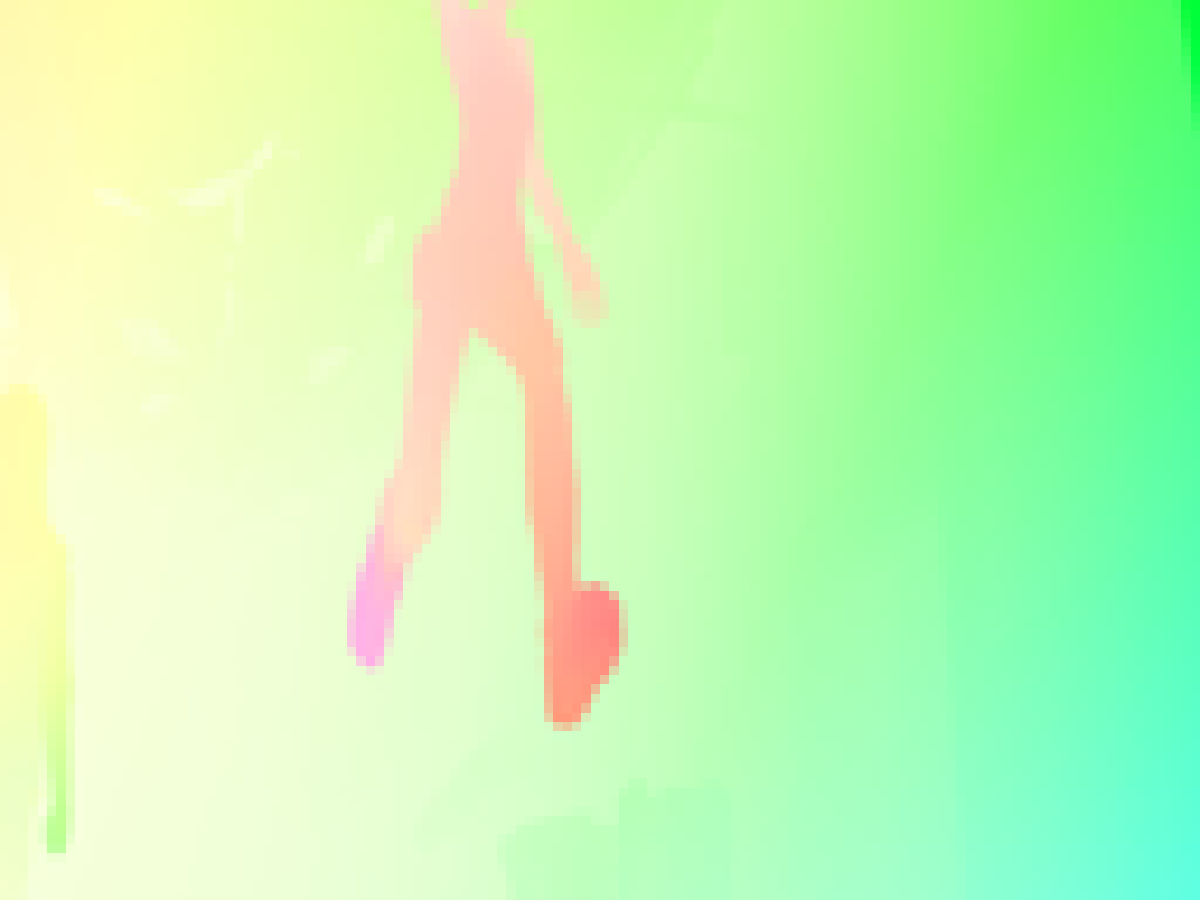}
    \end{minipage}
    &
    \begin{minipage}{.1\textwidth}
      \vspace*{0.05\textwidth}
      \includegraphics[width=\linewidth, height=10mm]{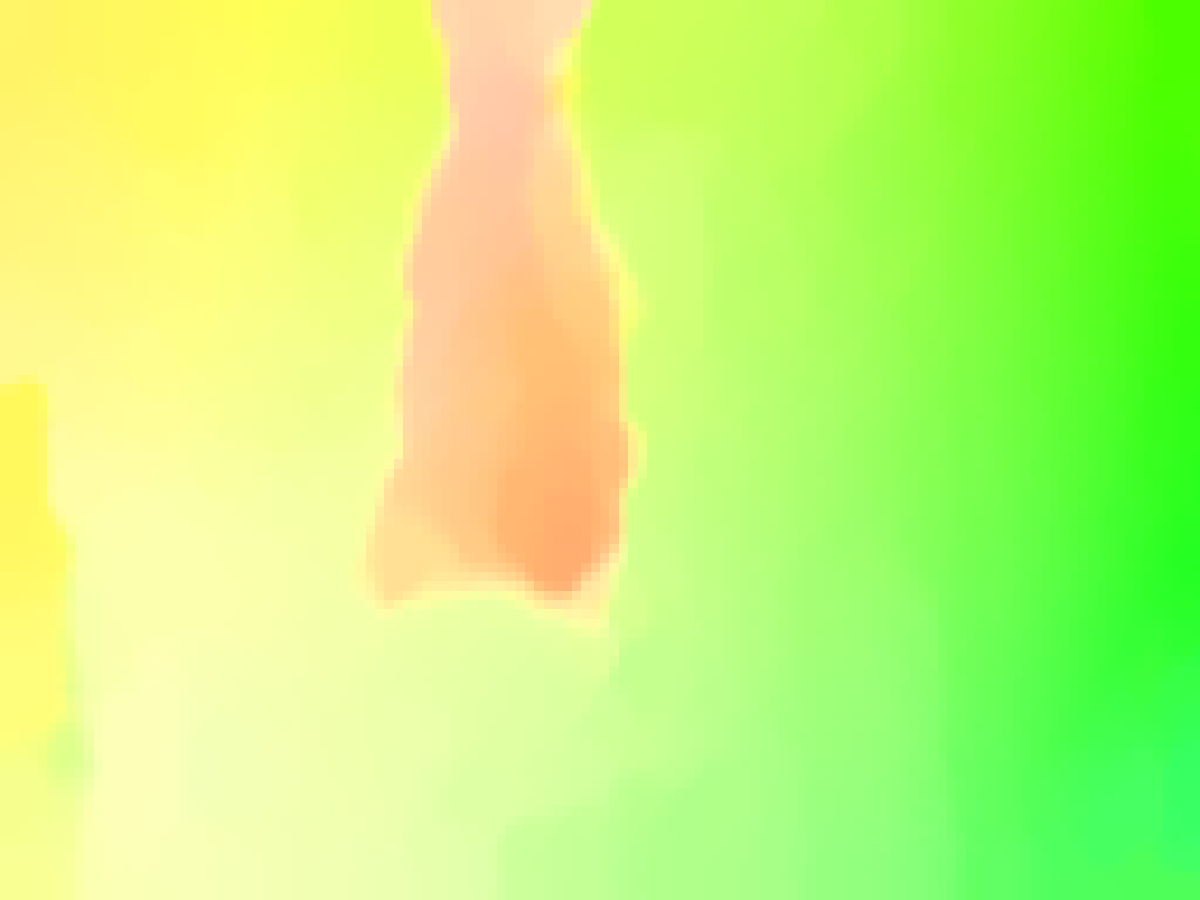}
    \end{minipage}
    & 
    \begin{minipage}{.1\textwidth}
      \vspace*{0.05\textwidth}
      \includegraphics[width=\linewidth, height=10mm]{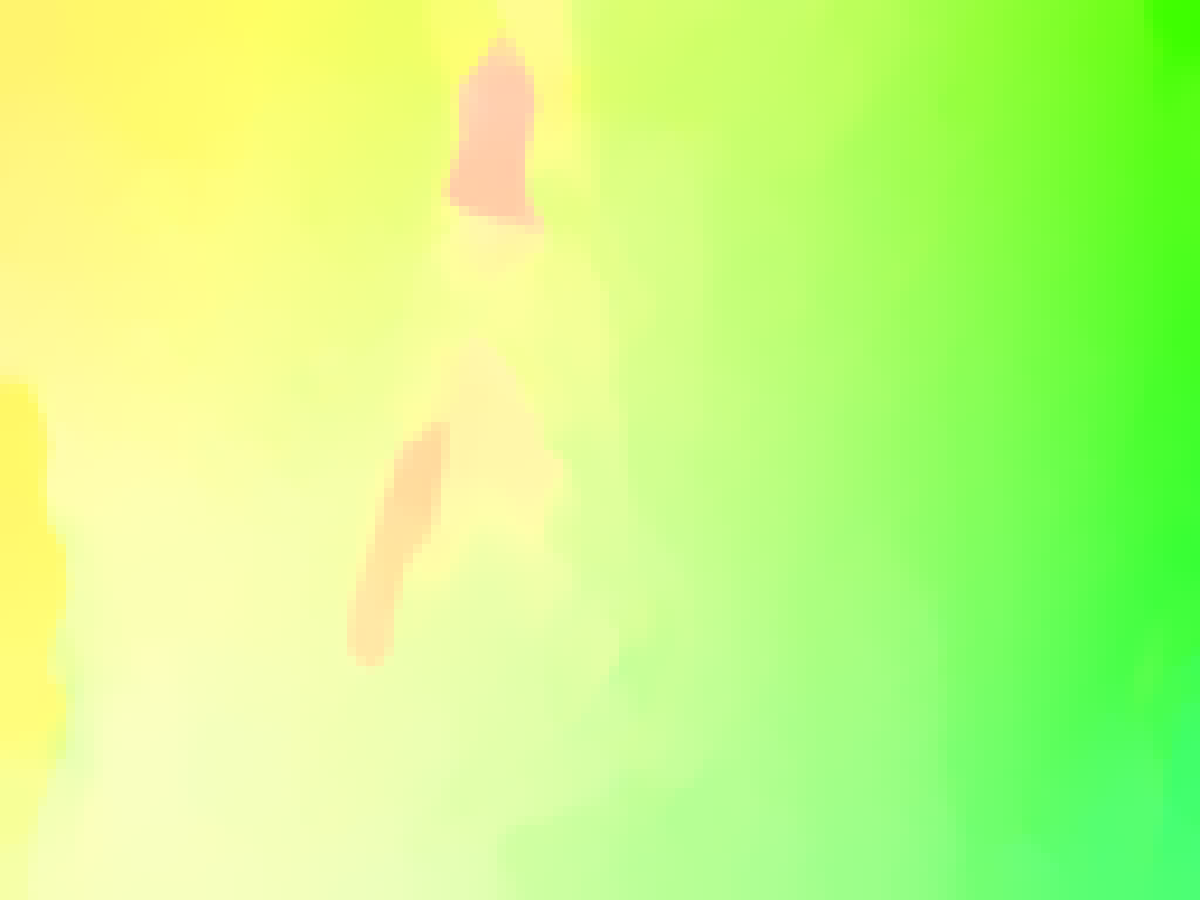}
    \end{minipage}
    &
    \begin{minipage}{.1\textwidth}
      \vspace*{0.05\textwidth}
      \includegraphics[width=\linewidth, height=10mm]{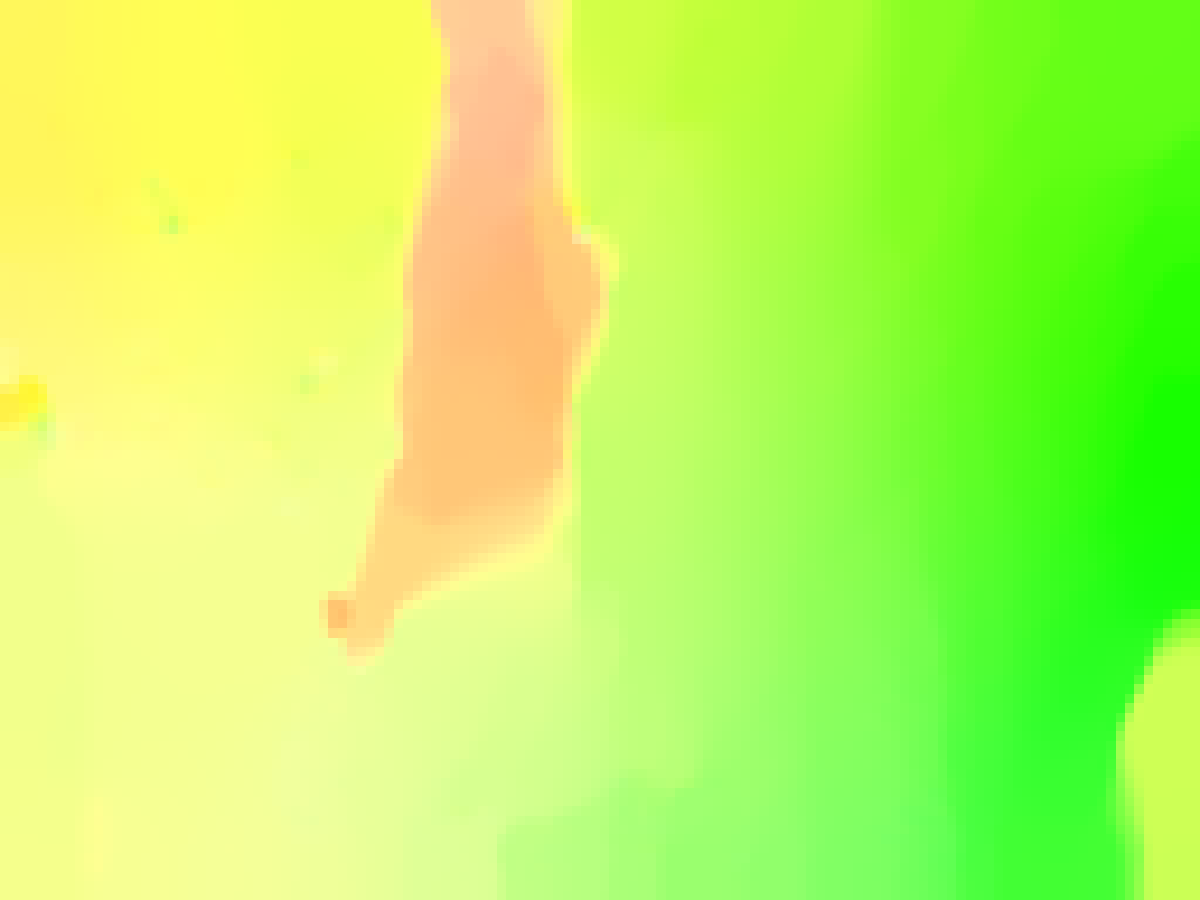}
    \end{minipage}
    &
    \begin{minipage}{.1\textwidth}
      \vspace*{0.05\textwidth}
      \includegraphics[width=\linewidth, height=10mm]{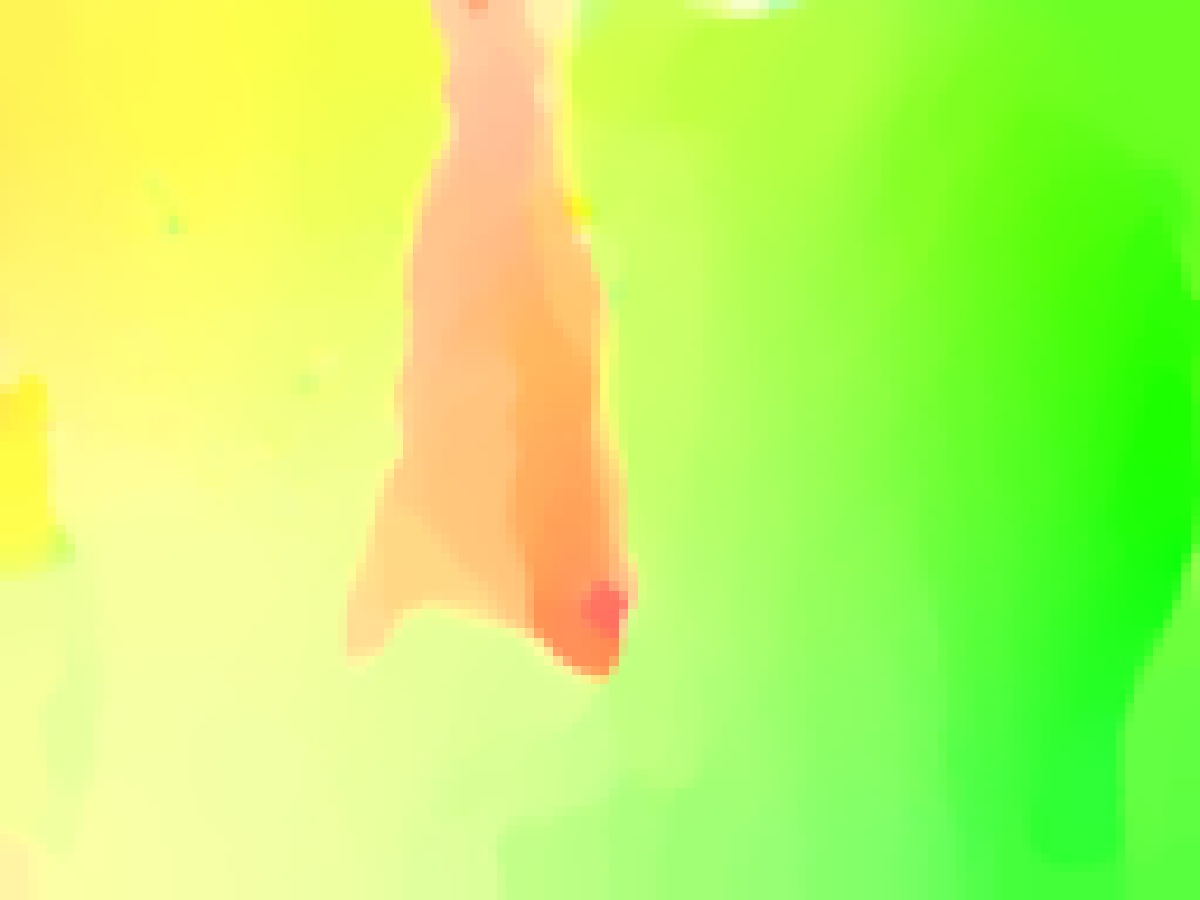}
    \end{minipage}
    &
    \begin{minipage}{.1\textwidth}
      \vspace*{0.05\textwidth}
      \includegraphics[width=\linewidth, height=10mm]{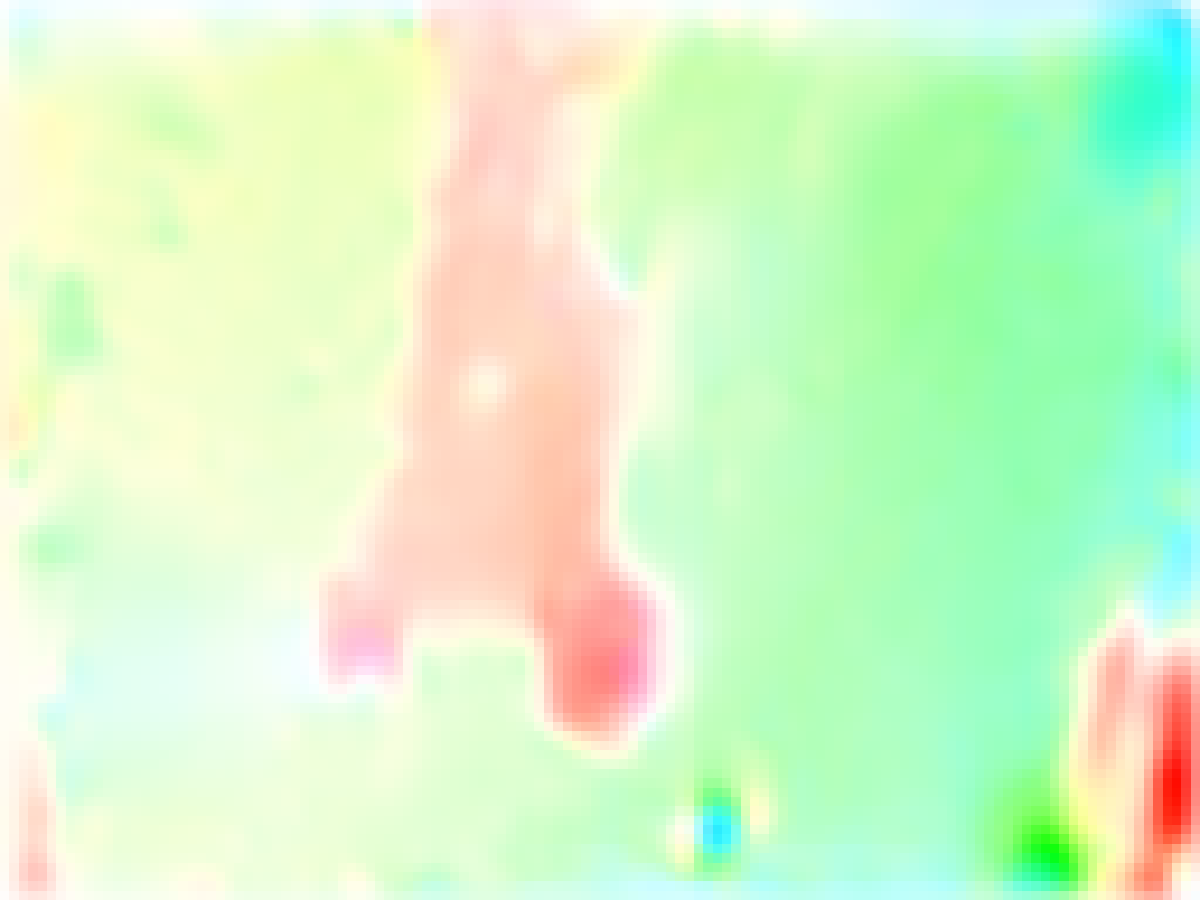}
    \end{minipage}
    
    \\
    
    \begin{minipage}{.1\textwidth}
      \vspace*{0.05\textwidth}
      \includegraphics[width=\linewidth, height=10mm]{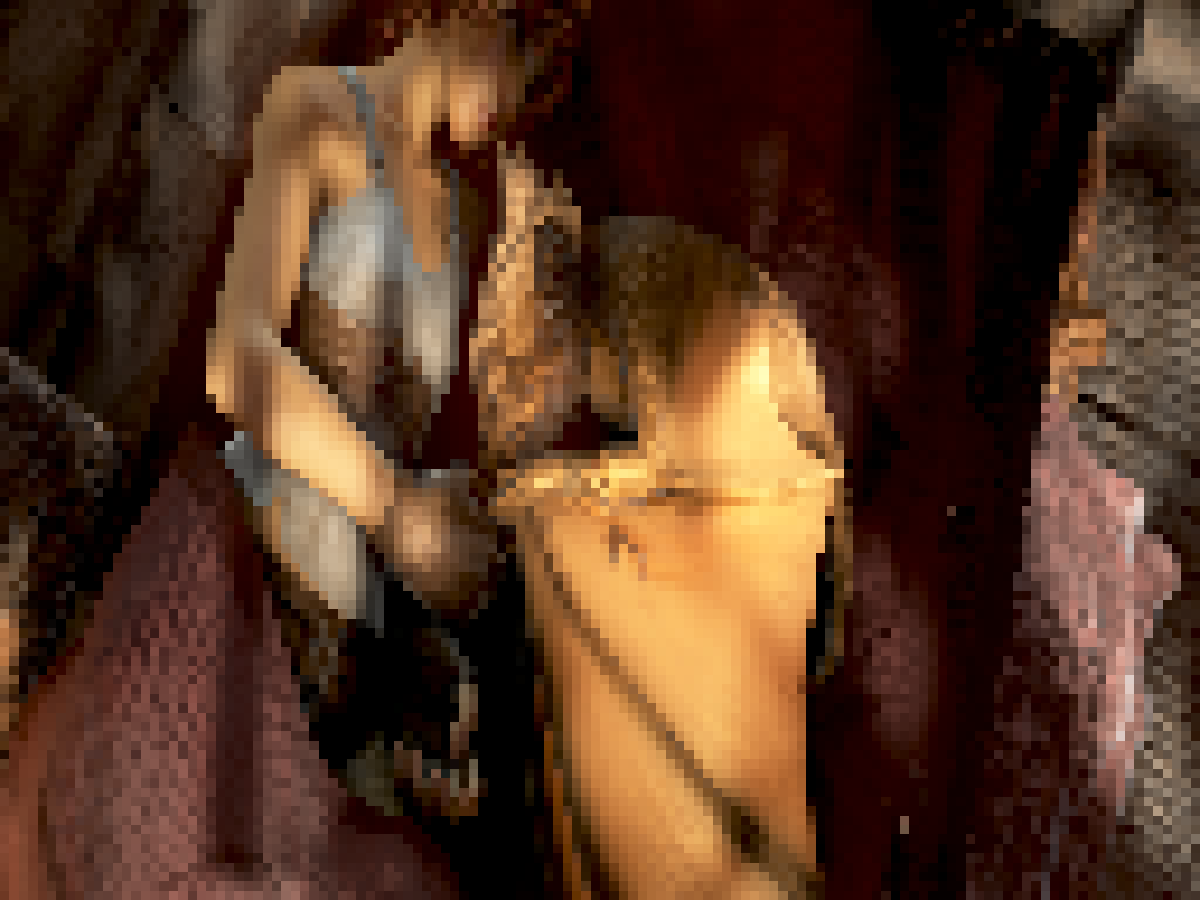}
    \end{minipage}
    &
    \begin{minipage}{.1\textwidth}
      \vspace*{0.05\textwidth}
      \includegraphics[width=\linewidth, height=10mm]{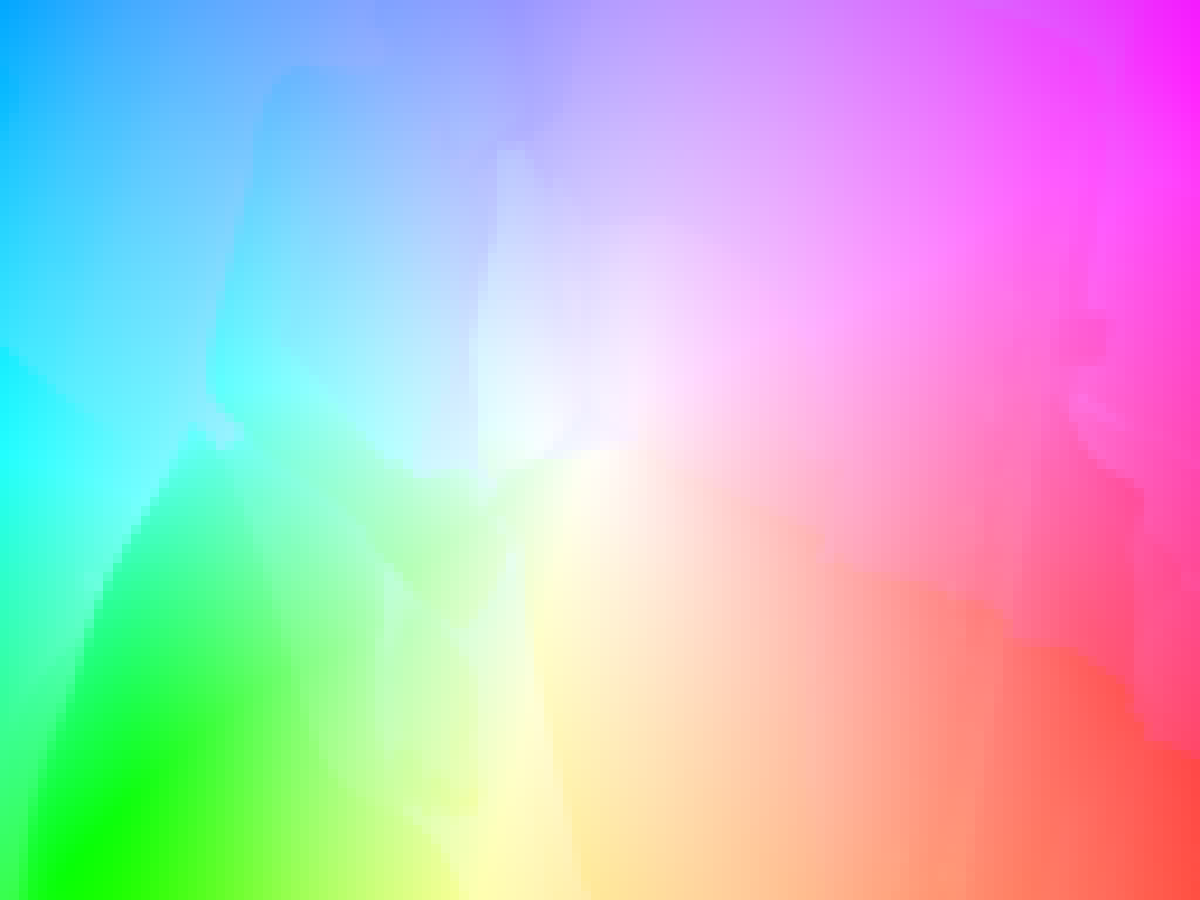}
    \end{minipage}
    &
    \begin{minipage}{.1\textwidth}
      \vspace*{0.05\textwidth}
      \includegraphics[width=\linewidth, height=10mm]{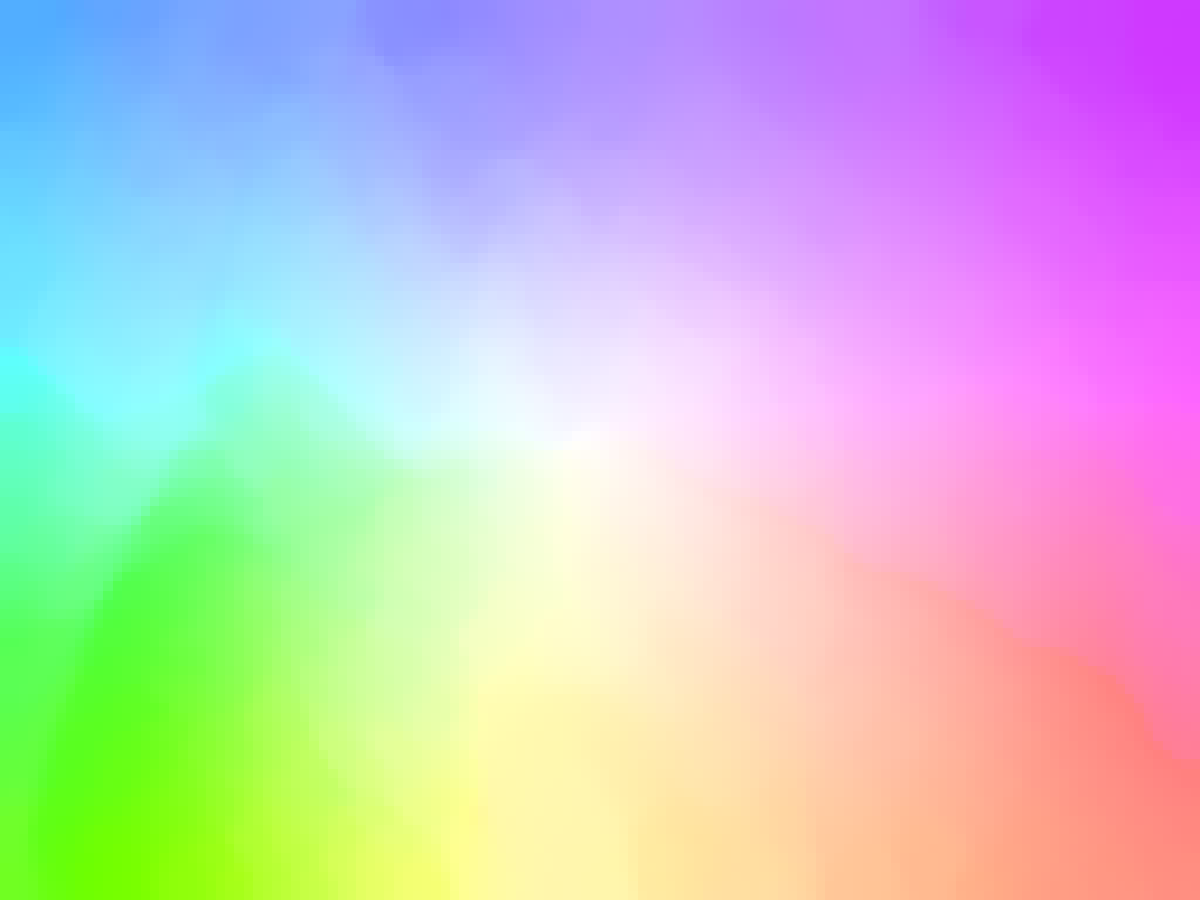}
    \end{minipage}
    & 
    \begin{minipage}{.1\textwidth}
      \vspace*{0.05\textwidth}
      \includegraphics[width=\linewidth, height=10mm]{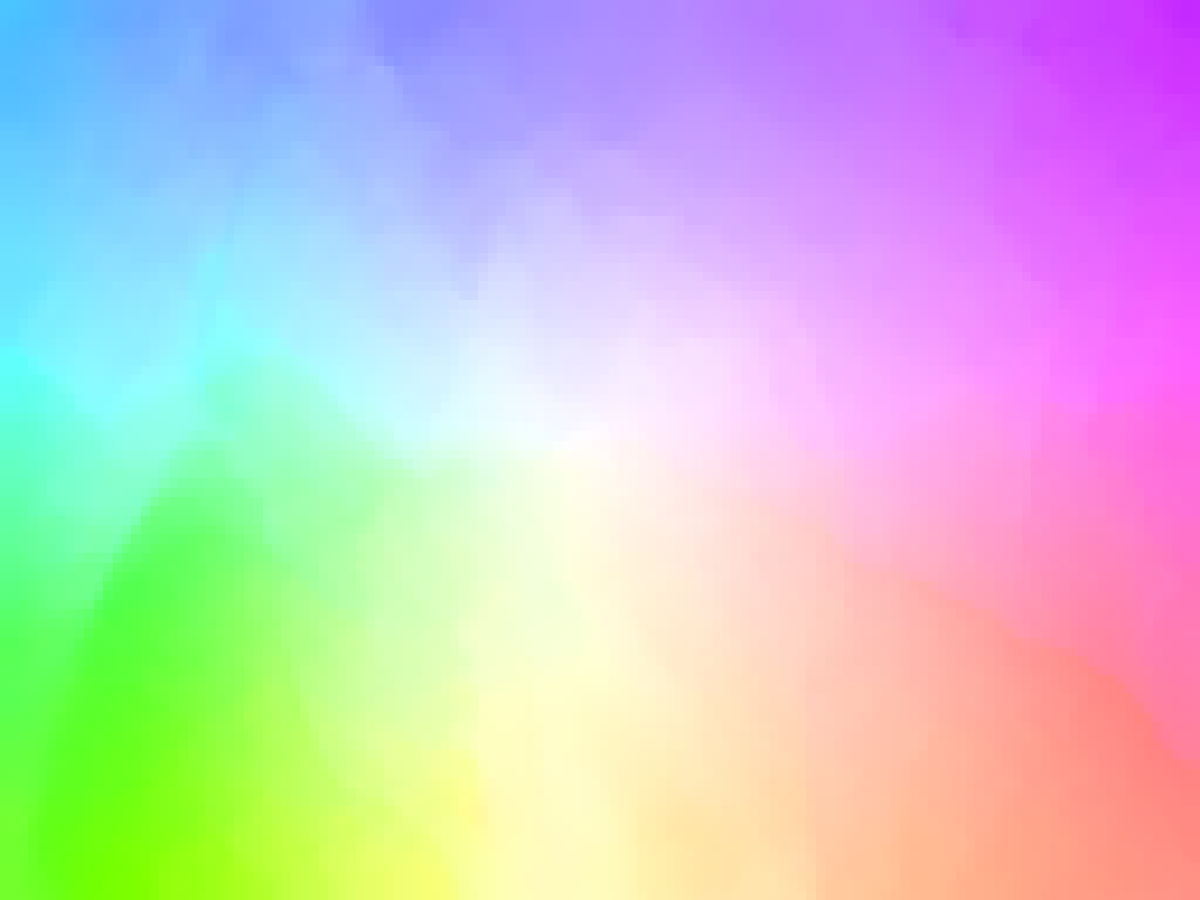}
    \end{minipage}
    &
    \begin{minipage}{.1\textwidth}
      \vspace*{0.05\textwidth}
      \includegraphics[width=\linewidth, height=10mm]{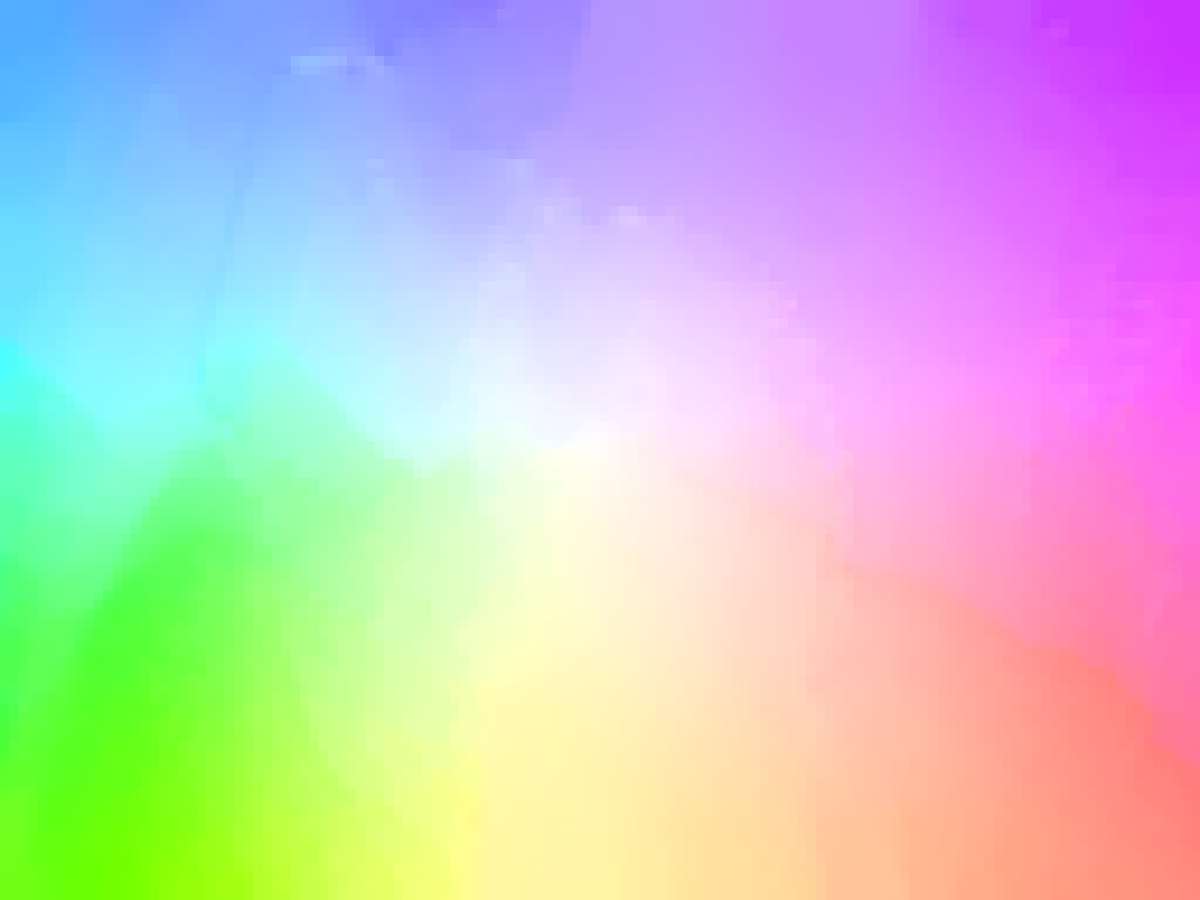}
    \end{minipage}
    &
    \begin{minipage}{.1\textwidth}
      \vspace*{0.05\textwidth}
      \includegraphics[width=\linewidth, height=10mm]{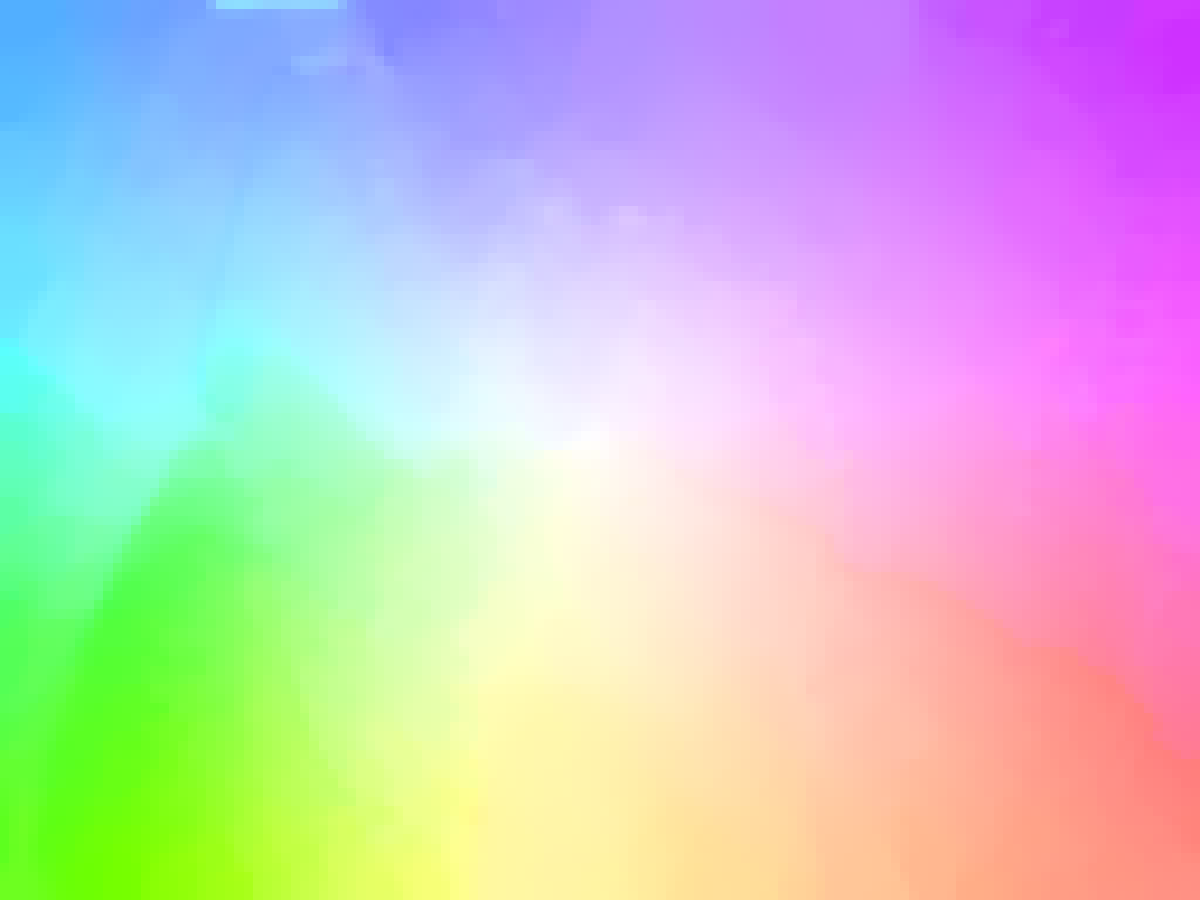}
    \end{minipage}
    &
    \begin{minipage}{.1\textwidth}
      \vspace*{0.05\textwidth}
      \includegraphics[width=\linewidth, height=10mm]{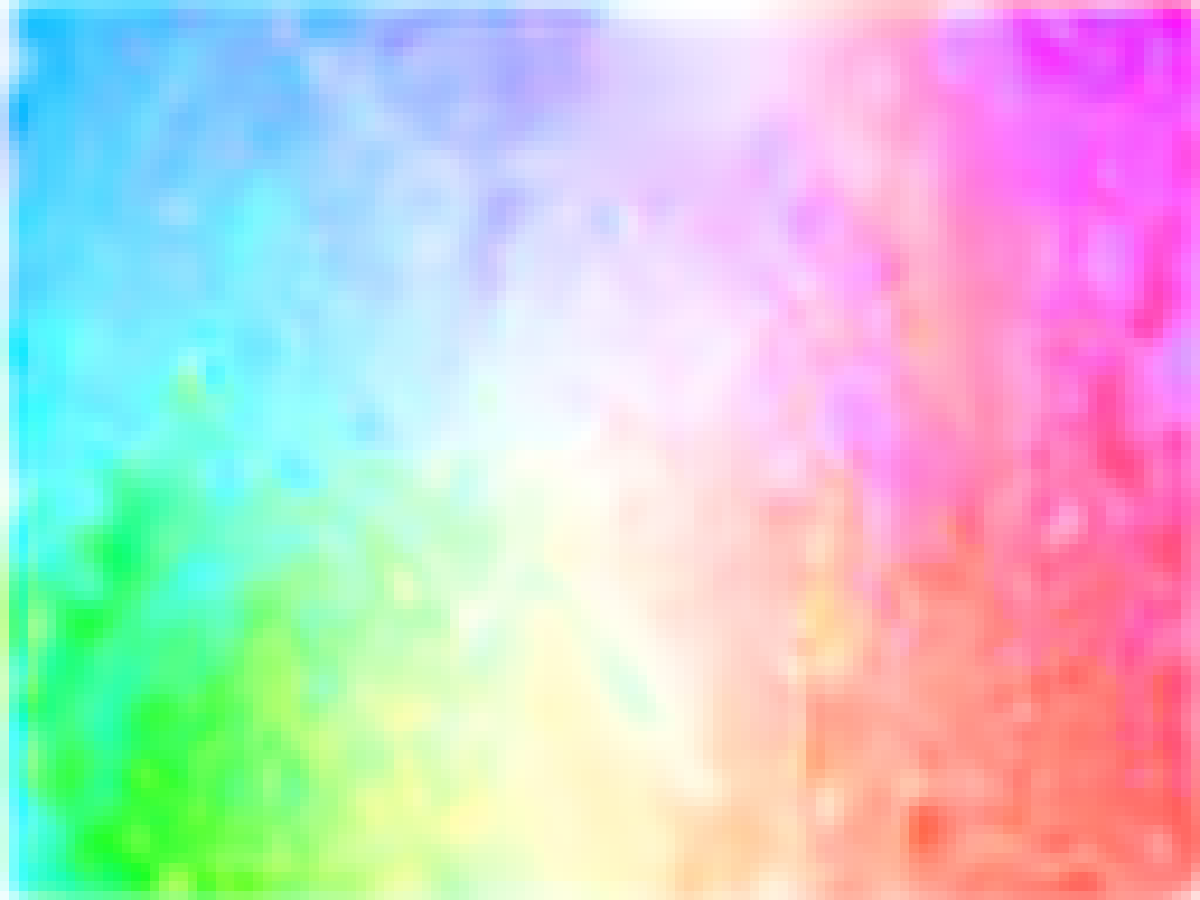}
    \end{minipage}
    
    \\

  \end{tabular}
  \caption{The motion field predicted by different motion estimation methods on several samples of MPISintel. }\label{fig:MF_Sintel}
\end{figure*}



\subsection{Architecture}
\begin{figure*}[!t]
\centering
\includegraphics[width=0.9\textwidth]{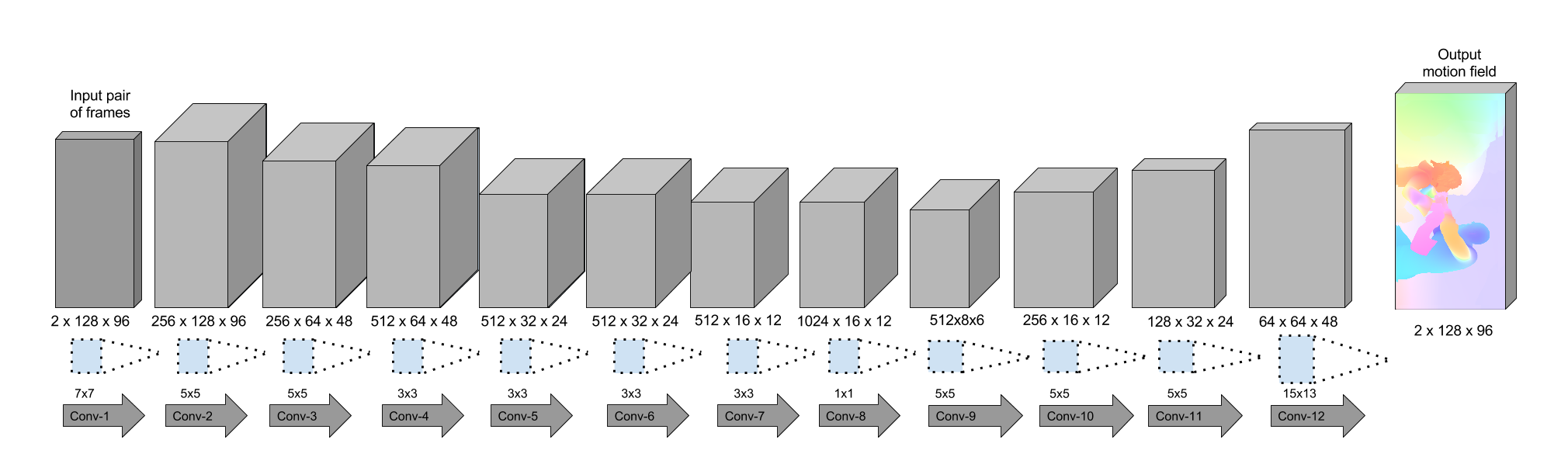}
\caption{The architecture of our proposed CNN. We have assumed the height and width of the input is 128x96. The illustrated motion field is chosen from MPI-Sintel dataset \cite{butler2012naturalistic} }
\label{fig:CNNArchDwSmp}
\end{figure*}

We propose a fully convolutional neural network with 12 convolutional layers. The architecture could be imagined as two parts. The CNN makes a compact representation of motion information in the first part which involves 4 downsamplings. This compact representation is then used to reconstruct the motion field in the second part which involves 4 up-samplings. The up-sampled is performed by simply repeating the rows and columns of the feature maps. Since our proposed DNN is fully convolutional, the input could be of any size. \fig{fig:CNNArchDwSmp} shows the two parts of the proposed CNN. To update the CNN weights during the training phase, we used ADAM \cite{kingma2014adam} and calculate that spatiotemporal intensity derivatives $I_x$, $I_y$ and $I_t$ as proposed in \cite{horn1981determining}.

\begin{table*}[!ht]
\begin{center}

\caption{Evaluation of different methods on UCF101 dataset. AEE-05 stands for Average End-point Error for the motions smaller than 5 pixels. AEE-5so refers to the motions bigger than 5 pixels, and AEE-tot refers to the total error value. AAE stands for Average Angular Error.}
\begin{tabular}{|c|c|c|c|c|c|c|c|}
\hline
Method &  AEE-05 & AEE-5so & AEE-tot &  AAE-05 & AAE-5so & AAE-tot \\
\hline\hline
DeepFlow & 0.30 & 3.99 & 0.47 & 9.35& 14.65& 9.59\\ 
\hline
HAOF  & 0.37& 4.59& 0.56& 10.95& 19.40& 11.33\\ 
\hline
LDOF  & 0.35& 2.85& 0.46& 9.91& 9.72& 9.9\\ 
\hline
USCNN  & 0.46& 8.7& 0.81& 12.74& 59.50& 14.70\\ 
\hline
\end{tabular}
\label{CompareOnUCF101}
\end{center}
\end{table*}

\section{Experiments}
\label{sec:experiment}

In order to evaluate the performance of our method, we report its results on 2 datasets, namely the real UCF101 and the synthetically generated MPI-Sintel, and compare them with the results of other state-of-the-art methods. We train the proposed Unsupervised trained CNN (USCNN) on 20K pairs of consecutive grayscale frames randomly selected from the about 1 million frames of the UCF101 dataset \cite{UCF101}. We first test the method on the same dataset, using for the evaluation 10K pairs of frames that were randomly selected from the UCF101 dataset making sure that there is no overlap with the training set. Since there is no ground truth motion field for UCF101 dataset, we use as ground truth the output of the EpicFlow - to the best of our knowledge this is the state-of-the-art method for motion estimation. \tbl{CompareOnUCF101} reports the results of the proposed method and three other state-of-the-art methods in the field, with the notable exception of FlowNet for which, neither the training network nor the training dataset are available. As it can be seen, the proposed method has comparable performance for motions less than 5 pixels - for larger motions as it is largely expected from methods that rely on coarse-to-fine schemes that involve downsampling it has lower accuracy.




To evaluate how well our method can generalize to an unknown dataset, we have applied it on the MPI-Sintel Final \cite{butler2012naturalistic} which is one of the most realistic synthetic datasets for which ground truth is available. As reported in \tbl{MPISintelComparison}, the USCNN has a better performance than LDOF and has a comparable performance in comparison with the other state-of-the-art methods. 
In \tbl{MPISintelComparison}, the performance measures for EpicFlow, DeepFlow, LDOF, and FlowNet are from \cite{DFIB15} and the performance measure for HAOF is calculated using their publicly available code.

To have a fair comparison, we report the performance measure for the two architectures introduced by \cite{DFIB15} without finetuning. 'FlowNetS' has a rather generic architecture which receives as input the two images stacked together. In 'FlowNetC', first meaningful representations are made in two separate but identical streams each receiving one of the input images and then their combination is fed into another stream for motion estimation. Our network architecture is closest to 'FlowNetS'. The motion field of several samples from UCF101 and MPI-Sintel datasets, estimated by various methods are depicted in \fig{fig:MF_UCF} and \fig{fig:MF_Sintel} respectively.

\begin{table}[!h]
\begin{center}
\caption{The performance comparison between different methods on MPI-Sintel.}
\begin{tabular}{|l|c|}
\hline
Method &  MPI-Sintel Final\\
\hline\hline
EpicFlow & 6.29 \\
DeepFlow & 7.212 \\
HAOF & 7.56 \\
LDOF & 9.12 \\
\hline
FlowNetS & 8.43 \\
FlowNetC & 8.81 \\
\hline
USCNN & 8.88 \\
\hline
\end{tabular}
\label{MPISintelComparison}
\end{center}
\end{table}


\section{Conclusions}
\label{sec:conclusion}
In this work, we propose estimating dense motion fields with CNNs. We show that, surprisingly perhaps, a simple cost function that relies on the optical flow equation can be used successfully for training a deep convolutional network in a completely unsupervised manner and without the need of any regularization or other constraints. Our CNN is trained on the UCF101 dataset. We show that it has a performance that is comparable to other state-of-the-art methods, especially for motions that are not large in magnitude, and that it can generalize very well to an unknown dataset, MPI-Sintel, without the need for refinement. The proposed method in this paper is among the very few studies conducted on the application of DNNs for motion estimation.


\section{Acknowledgment}
We gratefully acknowledge the support of NVIDIA Corporation with the donation of the Tesla K40 GPU used for this research.


\bibliographystyle{IEEEtran}
\bibliography{icip15}

\begin{thebibliography}{10}
\providecommand{\url}[1]{#1}
\csname url@samestyle\endcsname
\providecommand{\newblock}{\relax}
\providecommand{\bibinfo}[2]{#2}
\providecommand{\BIBentrySTDinterwordspacing}{\spaceskip=0pt\relax}
\providecommand{\BIBentryALTinterwordstretchfactor}{4}
\providecommand{\BIBentryALTinterwordspacing}{\spaceskip=\fontdimen2\font plus
\BIBentryALTinterwordstretchfactor\fontdimen3\font minus
  \fontdimen4\font\relax}
\providecommand{\BIBforeignlanguage}[2]{{%
\expandafter\ifx\csname l@#1\endcsname\relax
\typeout{** WARNING: IEEEtran.bst: No hyphenation pattern has been}%
\typeout{** loaded for the language `#1'. Using the pattern for}%
\typeout{** the default language instead.}%
\else
\language=\csname l@#1\endcsname
\fi
#2}}
\providecommand{\BIBdecl}{\relax}
\BIBdecl

\bibitem{dufaux2000efficient}
F.~Dufaux and J.~Konrad, ``Efficient, robust, and fast global motion estimation
  for video coding,'' \emph{Image Processing, IEEE Transactions on}, vol.~9,
  no.~3, pp. 497--501, 2000.

\bibitem{li2001mpeg}
W.~Li, ``Mpeg-4 video verification model version 18.0,'' \emph{ISO/IEC
  JTC1/SC29/WG11, N3908}, 2001.

\bibitem{keeling2006medical}
S.~L. Keeling, ``Medical image registration and interpolation by optical flow
  with maximal rigidity,'' in \emph{Mathematical Models for Registration and
  Applications to Medical imaging}.\hskip 1em plus 0.5em minus 0.4em\relax
  Springer, 2006, pp. 27--61.

\bibitem{long_shelhamer_fcn}
J.~Long, E.~Shelhamer, and T.~Darrell, ``Fully convolutional networks for
  semantic segmentation,'' \emph{CVPR (to appear)}, Nov. 2015.

\bibitem{jain2013better}
M.~Jain, H.~J{\'e}gou, and P.~Bouthemy, ``Better exploiting motion for better
  action recognition,'' in \emph{Computer Vision and Pattern Recognition
  (CVPR), 2013 IEEE Conference on}.\hskip 1em plus 0.5em minus 0.4em\relax
  IEEE, 2013, pp. 2555--2562.

\bibitem{wang2013action}
H.~Wang and C.~Schmid, ``Action recognition with improved trajectories,'' in
  \emph{Computer Vision (ICCV), 2013 IEEE International Conference on}.\hskip
  1em plus 0.5em minus 0.4em\relax IEEE, 2013, pp. 3551--3558.

\bibitem{horn1981determining}
B.~K. Horn and B.~G. Schunck, ``Determining optical flow,'' in \emph{1981
  Technical symposium east}.\hskip 1em plus 0.5em minus 0.4em\relax
  International Society for Optics and Photonics, 1981, pp. 319--331.

\bibitem{brox2004high}
T.~Brox, A.~Bruhn, N.~Papenberg, and J.~Weickert, ``High accuracy optical flow
  estimation based on a theory for warping,'' in \emph{Computer Vision-ECCV
  2004}.\hskip 1em plus 0.5em minus 0.4em\relax Springer, 2004, pp. 25--36.

\bibitem{brox2011large}
T.~Brox and J.~Malik, ``Large displacement optical flow: descriptor matching in
  variational motion estimation,'' \emph{Pattern Analysis and Machine
  Intelligence, IEEE Transactions on}, vol.~33, no.~3, pp. 500--513, 2011.

\bibitem{krizhevsky2012imagenet}
A.~Krizhevsky, I.~Sutskever, and G.~E. Hinton, ``Imagenet classification with
  deep convolutional neural networks,'' in \emph{Advances in neural information
  processing systems}, 2012, pp. 1097--1105.

\bibitem{DFIB15}
\BIBentryALTinterwordspacing
A.~Dosovitskiy, P.~Fischer, E.~Ilg, P.~H{\"o}usser, C.~Haz{\i}rba{\c{s}},
  V.~Golkov, P.~v.~d. Smagt, D.~Cremers, and T.~Brox, ``Flownet: Learning
  optical flow with convolutional networks,'' in \emph{IEEE International
  Conference on Computer Vision (ICCV)}, Dec 2015. [Online]. Available:
  \url{http://lmb.informatik.uni-freiburg.de//Publications/2015/DFIB15}
\BIBentrySTDinterwordspacing

\bibitem{goldstein2012global}
T.~Goldstein, X.~Bresson, and S.~Osher, ``Global minimization of markov random
  fields with applications to optical flow,'' \emph{Inverse Problems and
  Imaging}, vol.~6, no.~4, pp. 623--644, 2012.

\bibitem{bruhn2005lucas}
A.~Bruhn, J.~Weickert, and C.~Schn{\"o}rr, ``Lucas/kanade meets horn/schunck:
  Combining local and global optic flow methods,'' \emph{International Journal
  of Computer Vision}, vol.~61, no.~3, pp. 211--231, 2005.

\bibitem{sun2014quantitative}
D.~Sun, S.~Roth, and M.~J. Black, ``A quantitative analysis of current
  practices in optical flow estimation and the principles behind them,''
  \emph{International Journal of Computer Vision}, vol. 106, no.~2, pp.
  115--137, 2014.

\bibitem{kingma2014adam}
D.~P. Kingma and J.~Ba, ``Adam: A method for stochastic optimization,''
  \emph{Proceedings of the 3rd International Conference on Learning
  Representations (ICLR)}, 2014.

\bibitem{butler2012naturalistic}
D.~J. Butler, J.~Wulff, G.~B. Stanley, and M.~J. Black, ``A naturalistic open
  source movie for optical flow evaluation,'' in \emph{Computer Vision--ECCV
  2012}.\hskip 1em plus 0.5em minus 0.4em\relax Springer, 2012, pp. 611--625.

\bibitem{UCF101}
K.~Soomro, A.~Roshan~Zamir, and M.~Shah, ``{UCF101}: A dataset of 101 human
  actions classes from videos in the wild,'' in \emph{CRCV-TR-12-01}, 2012.

\end{thebibliography}

\end{document}